\documentclass[a4paper,fleqn]{cas-dc}

\usepackage{natbib}
\usepackage{hyperref}
\usepackage{algorithm}
\usepackage{algorithmic}
\usepackage[normalem]{ulem}
\usepackage{multirow}
\def\tsc#1{\csdef{#1}{\textsc{\lowercase{#1}}\xspace}}
\tsc{WGM}
\tsc{QE}
\tsc{EP}
\tsc{PMS}
\tsc{BEC}
\tsc{DE}

\begin{document}

\renewcommand{\arraystretch}{0.6}

\let\WriteBookmarks\relax
\def\floatpagepagefraction{1}
\def\textpagefraction{.001}
\shorttitle{Benchmark Evaluation of Federated  Learning on Multi-organ Images}
\shortauthors{Mao et~al.}

\title [mode = title]{Benchmark Evaluation of Federated Learning on Multi-organ Images}                      

\author[1]{Junbin Mao}
\author[1]{Xu Tian}
\author[1]{Jianchun Zhu}
\author[2]{Ludi Li}
\cortext[cor1]{Corresponding author.}
\author[1,2]{Jin Liu \corref{cor1}}[orcid = 0000-0002-4961-7074]
\ead{liujin06@csu.edu.cn}

\address[1]{Hunan Provincial Key Laboratory on Bioinformatics, School of Computer Science and Engineering, Central South University, Changsha 410083, China}
\address[2]{Xinjiang Engineering Research Center of Big Data  and Intelligent Software, School of software, Xinjiang University, Urumqi, 830008, China}

\begin{abstract}
The privacy requirements of medical data and its substantial variations across organs and modalities hinder the clinical implementation of medical AI. Federated learning (FL) is a feasible approach to overcome these challenges. Due to the continuous emergence of FL algorithms and the highly heterogeneous nature of medical data, objectively evaluating their performance in real-world clinical settings remains difficult. Therefore, a comprehensive federated medical imaging benchmark, serving as a unified evaluation standard, is crucial for advancing the technology toward reliable clinical application. Existing federated medical imaging benchmarks have not yet adequately incorporated state-of-the-art algorithms, are limited to data from single organs or modalities, and overly emphasize model accuracy, making it difficult to comprehensively assess the overall efficacy of FL in real-world medical environments. To address these challenges, we developed the MobenFL benchmark. This benchmark integrates 20 cutting-edge FL algorithms and 22 medical imaging datasets, covering 12 critical organs across the human body, surpassing existing benchmark in breadth. In terms of evaluation dimensions, MobenFL not only assesses performance but also systematically incorporates key metrics such as algorithmic efficiency and privacy protection capabilities. Additionally, it conducts specialized evaluations for complex real-world clinical scenarios involving different diseases, devices, and imaging modalities, thereby providing a comprehensive and in-depth evaluation framework for the clinical application of FL in the medical field. Our code is published at \url{https://github.com/yutian0315/MobenFL}.
\end{abstract}




\maketitle

\section{Introduction}

Currently, AI-based medical image analysis has matured in competitive research scenarios, but a significant gap remains between its technical maturity and large-scale clinical implementation. This gap stems primarily from the inherent characteristics of medical data: data accumulated by hospitals in different regions and specialties vary greatly in organ types and imaging modalities. For instance, one hospital may possess abundant lung CT data, while another specializes in brain MRI data. As a result, diagnostic models trained on single-center data suffer a sharp decline in generalization performance when applied across institutions~\citep{dayan2021federated}. Additionally, medical imaging data involve sensitive patient information and are strictly regulated by privacy protection laws such as HIPAA and GDPR~\citep{pati2022federated}, making the traditional approach of consolidating multi-center data to enhance model generalization nearly unfeasible both legally and compliantly. 

\begin{figure}[h]
    \centering
        \centering
        \includegraphics[width=\linewidth]{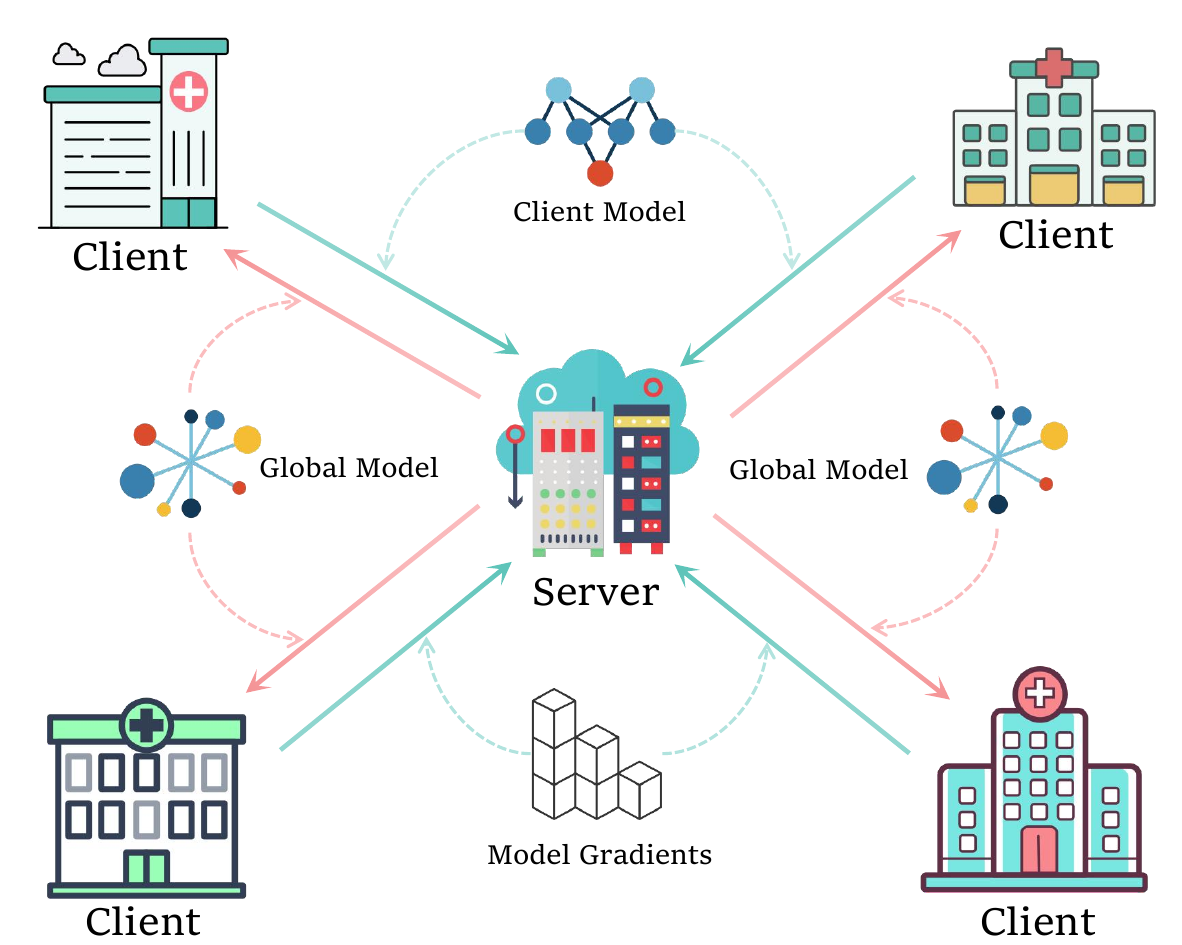} 
        \caption{Overview of the basic principles of federated learning. Clients send model parameters or model gradients to the server, and the server distributes the aggregated global model to the clients.} 
        \label{FL} 
\end{figure}

It is for these reasons that federated learning~\citep{mcmahan2017communication}, with its distributed collaborative paradigm of "moving models, not data" (shown in Fig.~\ref{FL}) demonstrates critical value. FL can effectively coordinate data resources from various centers without the need to aggregate raw data~\citep{karargyris2023federated}. By encrypting and aggregating locally trained model parameters, it enables the collaborative training of a global model with broader knowledge and stronger adaptability. 

FL is developing rapidly, with new algorithms emerging continuously. Meanwhile, the medical data environment is inherently highly heterogeneous, with significant variations in imaging devices, population distributions, and organ imaging across different institutions. Against this backdrop, the lack of unified evaluation standards makes it difficult to objectively and fairly measure the performance and robustness of different federated algorithms in real-world clinical scenarios, which greatly hinders the translation of high-performing algorithms into clinical applications. Therefore, establishing a comprehensive federated medical imaging benchmark is of critical importance. Such a benchmark would provide a public and fair evaluation standard, systematically assessing the performance of different algorithms in key clinical metrics including cross-center generalization capability, privacy protection strength, and communication efficiency. This will not only promote the evolution of federated learning toward greater reliability and practicality but also lay a crucial foundation for the safe, compliant, and widespread adoption of medical imaging AI.

\begin{figure}[htbp]
        \centering
        \includegraphics[width=\linewidth]{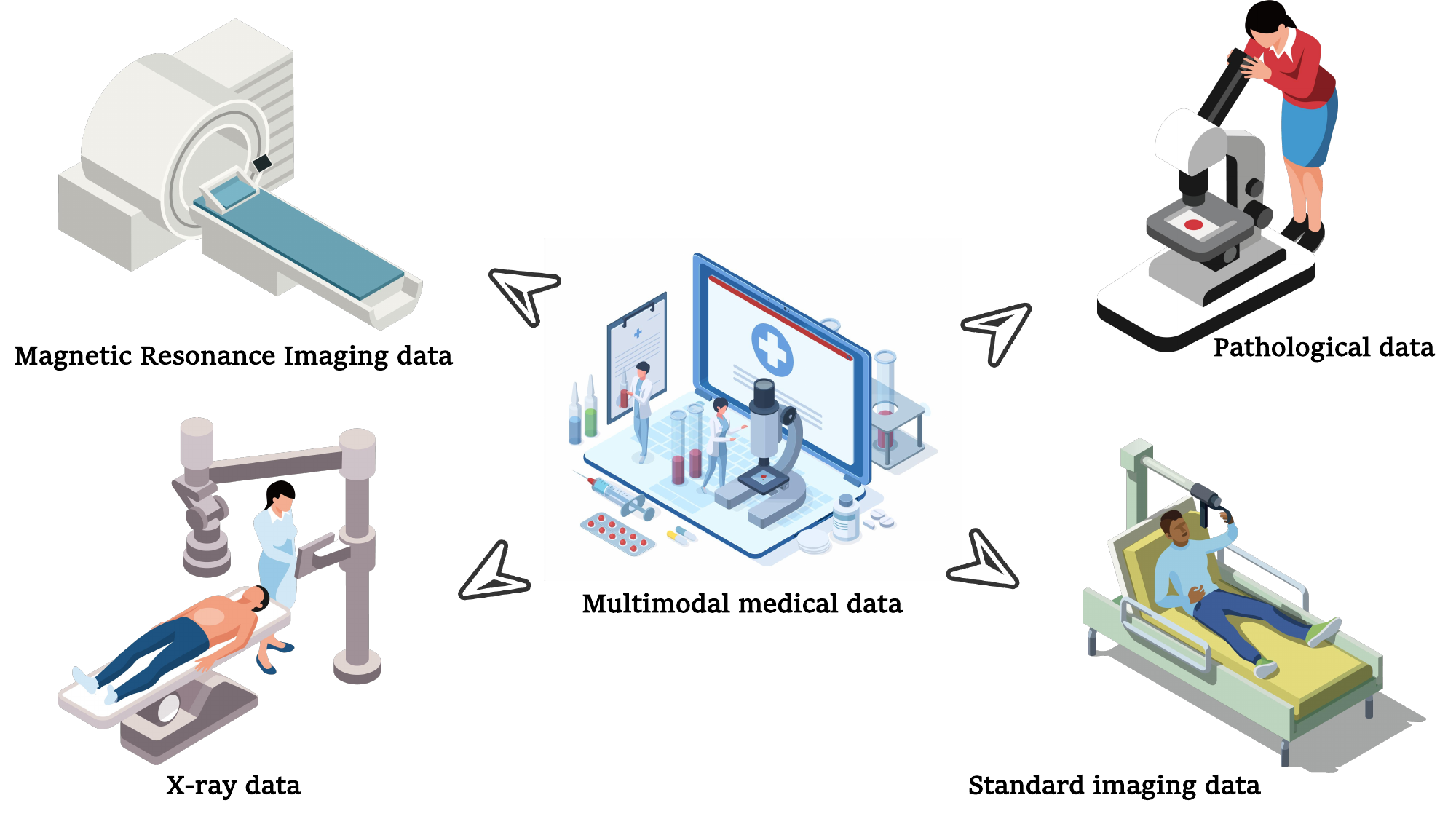} 
        \caption{The figure illustrates the multimodal medical data involved in this study. Among these, standard imaging data refers to dermatological and autism facial images captured via conventional cameras.
} 
        \label{MMD}
\end{figure}

In the field of general computing, widely recognized benchmark frameworks such as FedML\footnote{https://fedml.ai/home} and FATE\footnote{https://github.com/FederatedAI/FATE} have emerged. However, the professional evaluation system tailored for the specific domain of medical imaging remains relatively underdeveloped. Currently, only a limited number of benchmarks, such as FedCBD~\citep{lee2020federated} and Flamby~\citep{ogier2022flamby}, provide foundational evaluation support for federated medical imaging research, shown in Table~\ref{BenchCompare}.
Existing federated medical imaging benchmarks still exhibit significant shortcomings across multiple dimensions. At the algorithmic level, current benchmarks incorporate outdated methods and fail to promptly integrate cutting-edge algorithms with superior performance and updated architectures. There is a particular lack of systematic evaluation for emerging directions such as personalized federated learning~\citep{9743558}, split learning~\citep{GUPTA20181}, and decentralized learning~\citep{beltran2023decentralized}, which prevents these benchmarks from accurately reflecting the true state of current technological advancements.
In terms of data diversity, most benchmarks remain confined to single organs (such as the lungs or retina) or single imaging modalities (such as MRI or CT), lacking a comprehensive data system that covers multiple organs, multiple centers, and multiple disease types. This limitation not only undermines the clinical representativeness and generalization capability of the benchmarks but also restricts their evaluative value in environments with heterogeneous data distributions.
Furthermore, regarding evaluation metrics, existing efforts often overemphasize traditional performance indicators like model accuracy, while neglecting critical dimensions such as algorithmic efficiency and privacy protection strength, which are essential trade-offs in practical medical applications. This one-sided evaluation perspective makes it difficult to comprehensively assess the overall effectiveness of federated learning in real-world clinical scenarios.

\begin{table*}[!h]
\centering
\renewcommand\arraystretch{1.3}
\caption{Comparison between MobenFL and existing FL benchmarks in Medical Image.}
\begin{tabular}{lllll}
\hline
\textbf{Benchmark} & \textbf{Algorithms} & \textbf{Dataset} & \textbf{Data Simulation} & \textbf{Evaluations} \\ \hline
FedCBD~\citep{lee2020federated}             &        5             &         3         &             IID                      &          Performance            \\
Flamby~\citep{ogier2022flamby}             &          7           &         7         &              IID  and No-IID                    &          Performance          \\
MobenFL(ours)    &        20             &         22         &               IID and No-IID                     &         Performance, Privacy, Efficiency             \\ \hline
\label{BenchCompare}
\end{tabular}
\end{table*}

\begin{figure*}
    \centering
        \includegraphics[width=.9\linewidth]{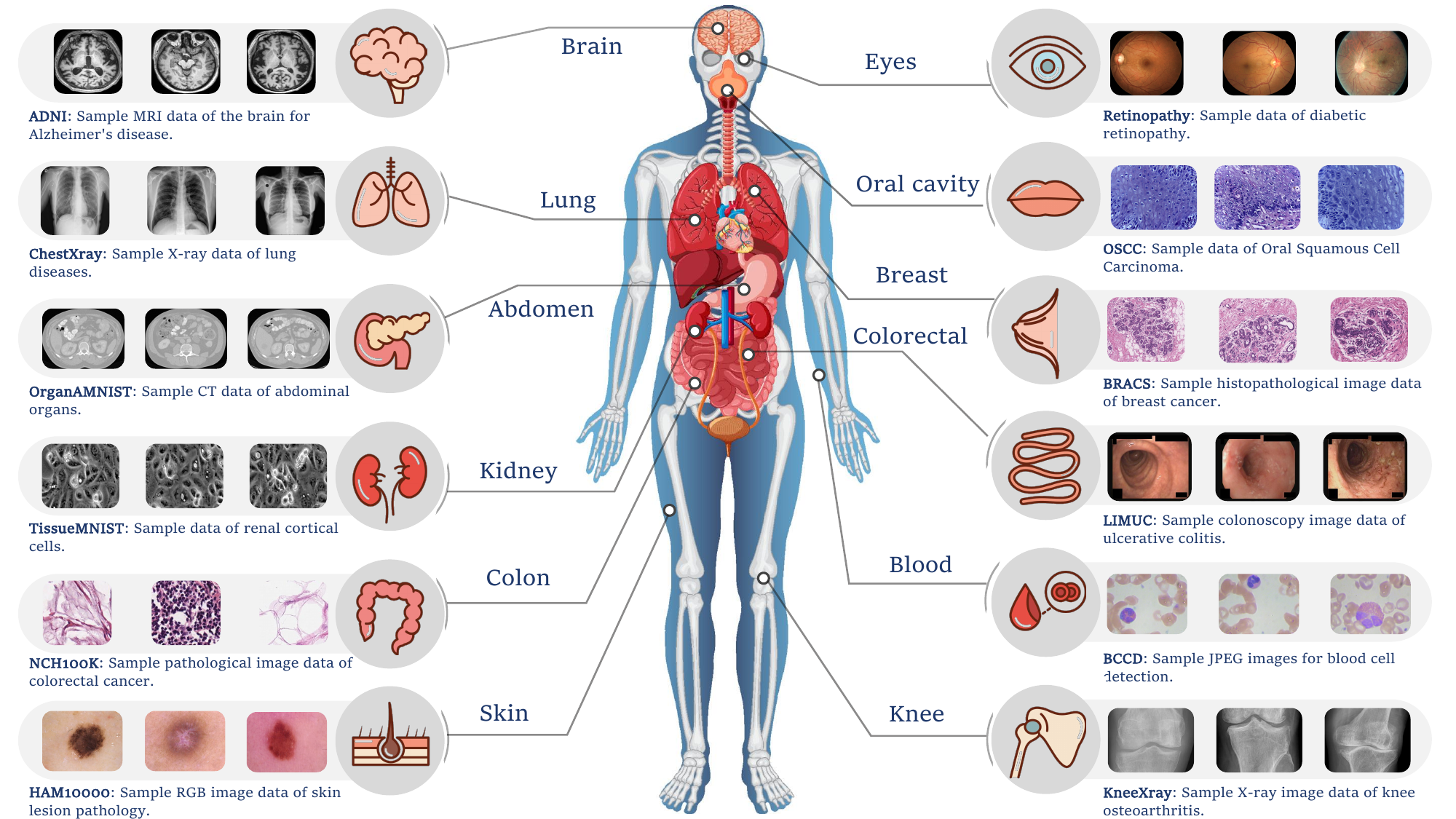}
        \caption{A multi-organ medical imaging dataset, covering various body parts such as the brain, abdomen, kidneys, colon, skin, eyes, oral cavity, breast, blood, and knee, with specific annotations showcasing publicly available sample data for diseases including Alzheimer's disease, diabetic retinopathy, oral squamous cell carcinoma, breast cancer, ulcerative colitis, and knee.}
        \label{organ}
\end{figure*}

In response to the aforementioned challenges, we have developed MobenFL—a comprehensive evaluation benchmark encompassing diverse federated learning algorithm types and multi-organ medical imaging data. Its purpose is to comprehensively evaluate federated learning algorithms through a rich medical data foundation, thereby promoting the clinical application of federated learning in the field of medical image analysis. MobenFL integrates 20 cutting-edge federated learning algorithms from the past three years (shown in Table~\ref{Benchmark_Methods}), systematically categorized by their types.
At the dataset level, MobenFL comprises 22 medical imaging datasets of different modalities, including magnetic resonance imaging, pathological images, X-ray images, and natural standard image data (shown in Fig.~\ref{MMD}). These datasets cover 12 critical human organs such as the brain, eyes, oral cavity, breasts, lung, abdomen, kidneys, and colon (shown in Fig.~\ref{organ}).
Regarding algorithm evaluation, MobenFL not only assesses the performance across all federated learning algorithm and dataset combinations, but also systematically evaluates the operational efficiency and privacy protection capabilities of the algorithms. Furthermore, the benchmark conducts specialized evaluations and analyses for three complex scenarios: different diseases, different modalities, and different imaging devices. The main contributions of our work are as follows:

\begin{enumerate}
    \item To the best of our knowledge, this is the most comprehensive benchmark in the field of federated medical image classification to date, covering 22 datasets involving 12 human organs and 20 different types of FL algorithms.

    \item We comprehensively evaluated the performance of FL algorithms from the aspects of effectiveness, efficiency, privacy, and different clinical scenarios.

    \item We have integrated all FL algorithms within a unified framework and standardized input interfaces for diverse data types, while also open-sourcing our framework to enable subsequent researchers to directly conduct further research and development.
    
\end{enumerate}

\section{Related works}

\subsection{Federated Learning}

FedAvg~\citep{mcmahan2017communication} is a foundational work in federated learning. Its core lies in simple weighted averaging aggregation: after multiple rounds of local stochastic gradient descent updates on each client, the server aggregates the received model parameters proportionally to their data volumes. However, its performance degradation and convergence instability under non-IID data have spurred extensive follow-up research. SCAFFOLD~\citep{karimireddy2020scaffold} addresses this by introducing control variates to estimate and correct the bias in client update directions, theoretically ensuring convergence speed under heterogeneous data and significantly improving communication efficiency. FedNova~\citep{wang2020tackling} approaches the issue from the perspective of aggregation normalization, correcting inconsistencies in update magnitudes caused by varying numbers of local iterations across clients, thereby aligning the aggregated global update direction more closely with that of ideal centralized training. FedDyn~\citep{durmus2021federated} sets regularization terms for each device in each round, ensuring that the global and local solutions converge in the limit. These methods have advanced the fundamental framework of federated learning from perspectives such as constrained optimization, variance reduction, and update correction. Based on the algorithmic foundations such as FedAvg and its variants, researchers have delved into the inherent challenges of federated learning from more granular perspectives.

A portion of research focus has shifted from simple weighted averaging toward more intelligent aggregation strategies. FedBN is among the earliest studies in this direction~\citep{li2021fedbn}, which preserves local feature distributions by excluding the sharing of BN layer parameters and only aggregating non-BN layers. FedCD~\citep{kopparapu2020fedcd} mitigates the negative impact of concept drift through local drift detection and weighted aggregation based on drift feedback. FedGH~\citep{yi2023fedgh} supports model heterogeneity by sharing feature representations and a global prediction head, thereby reducing communication and computational costs. FCCL~\citep{huang2023generalizable} achieves feature-level knowledge communication between heterogeneous models by aligning cross-correlation matrices and instance similarity distributions. FedAU~\citep{wang2024a} dynamically adjusts aggregation weights based on client participation history to improve performance.

While intelligent aggregation strategies improve the server-side synthesis of model updates, they largely still treat local client training as a black box. To more directly guide and stabilize the client-side optimization process, an alternative approach introduces explicit regularization terms into the local objective function. FedProx~\citep{li2020federated}, as an early regularization method in federated learning, addresses statistical heterogeneity by introducing a proximal term to constrain the deviation between local models and the global model. MOON~\citep{li2021model} mitigates client training drift caused by non-IID data using model contrastive learning as a regularization constraint. HarmoFL~\citep{jiang2022harmofl} reduces feature distribution heterogeneity through an amplitude normalization strategy and guides the model toward a flat optimum via weight perturbation. PGFed~\citep{luo2023pgfed} achieves efficient personalized learning by constructing a local objective function that incorporates approximations of the empirical risks from other clients. FedDecorr~\citep{10336535} prevents dimensional collapse by minimizing the Frobenius norm of the representation covariance matrix.

Regularization methods provide implicit guidance for local training at the optimization level, while another class of strategies shifts toward redesigning the model architecture itself. Instead of being confined to modifying the loss function, this approach explicitly adapts to data heterogeneity by introducing additional network components, thereby enabling more flexible and personalized federated learning. FedNP~\citep{wu2023fednp} employs a probabilistic neural network based on Expectation Propagation to estimate the global data distribution and prevents local overfitting through auxiliary task regularization. FedALA~\citep{zhang2023fedala} integrates global and local models via an Adaptive Local Aggregation module using element-wise weighting. GPFL~\citep{zhang2023gpfl} combines Global Category Embeddings and a Conditional Valve module to jointly learn global and personalized feature representations. TurboSVM~\citep{wang2024turbosvm} utilizes Support Vector Machines for selective aggregation and maximum margin extension regularization, adopting a "model-as-sample" strategy to distinguish beneficial updates.

The pursuit of personalized and efficient federated learning has driven architectural innovations that go beyond parametric adaptation. Split learning~\citep{GUPTA20181} represents such a paradigm shift, aiming to reduce the computational burden on clients and enhance input privacy protection. This approach retains the initial layers and raw data on the client side while offloading computationally intensive deeper layers to the server. This not only enables lightweight participation but also establishes a natural input firewall through intermediate feature obfuscation. FLOP~\citep{yang2021flop} freezes pre-trained model parameters and learns only client-side prompts, balancing shared knowledge and personalized features by aggregating global and local prompts. SplitFed~\citep{thapa2022splitfed} first integrates federated learning with split learning, significantly reducing client resource requirements through vertical model partitioning. SplitAVG~\citep{zhang2022splitavg} divides the network into client and server sub-networks, with unified server-side computation after feature map concatenation, effectively reducing communication overhead and mitigating data distribution discrepancies.

\subsection{Federated Learning in Medical Image Classification}

In recent years, numerous studies have explored the application of federated learning in the medical field and achieved encouraging results~\citep{kaissis2021end}. One of the primary medical applications is the development of predictive models for disease diagnosis and treatment~\citep{dayan2021federated}. For example, Bai et al. proposed an open-source medical artificial intelligence framework and implemented COVID-19 diagnosis through federated learning methods~\citep{bai2021advancing}. Dayan et al. developed a federated learning approach for predicting clinical outcomes in COVID-19 patients. Furthermore, federated learning has shown significant potential in the field of medical imaging analysis~\citep{kaissis2020secure}. For instance, Kaissis et al. reviewed emerging privacy-preserving methods in medical imaging analysis and discussed the limitations and drawbacks of existing approaches~\citep{kaissis2020secure}. Additionally, a general federated learning framework called PriMIA~\citep{kaissis2021end} was developed, and its advantages in privacy protection, secure aggregation, and encrypted inference were evaluated through experiments on pediatric chest X-ray image classification. Similarly, a federated learning method was recently proposed for predicting histological responses to neoadjuvant chemotherapy in triple-negative breast cancer~\citep{ogier2023federated}. For the rare disease glioblastoma, researchers leveraged federated learning to propose an automatic tumor boundary detector~\citep{pati2022federated}, which demonstrated excellent performance.

\section{Method}
\subsection{Federated learning}
Federated learning is a distributed machine learning framework that enables collaborative modeling among multiple parties without the need to share raw data. Suppose there are $N$ clients, where each client $i$ possesses a local dataset $\mathcal{D}_{i}=\{(x_{j}, y_{j})\}_{j=1}^{n_{i}}$, with $n_{i}$ being the number of samples. The global objective is to minimize the weighted loss across all clients:

\begin{equation}
\min_{w} \mathcal{L}(w) = \sum_{i=1}^{N} \frac{n_i}{n} \mathcal{L}_i(w), \quad n = \sum_{i=1}^{N} n_i,
\label{fl1}
\end{equation}
where $\mathcal{L}_i(w)$ is the local loss function (e.g., cross-entropy, mean squared error) of client $i$, $w$ represents the model parameters (e.g., weights of a neural network). Each client trains its local model using its own local dataset and then sends the updated model parameters or gradients to the server. The process of locally performing gradient descent to update parameters on each client can be expressed by the following formula:

\begin{equation}
w_i^{(t+1)} = w_i^{(t)} - \eta \nabla \mathcal{L}_i(w_i^{(t)}),
\label{fl2}
\end{equation}
where $w_i^{(t)}$ represents the model parameters of client $i$ at iteration $t$, $\eta$ is the learning rate, $\nabla \mathcal{L}_i(w_i^{(t)})$ is the gradient of the local loss function $\mathcal{L}_i$ with respect to $w_i^{(t)}$. Subsequently the server aggregates model updates from participating clients through weighted averaging, as follows:

\begin{equation}
w^{(t+1)} = \sum_{i=1}^{N} \frac{n_i}{n} w_i^{(t+1)}.
\label{fl3}
\end{equation}

The purpose of $\frac{n_i}{n}$ is to weight updates according to each client's sample proportion, while omitting $\frac{n_i}{n}$ allows equal weighting across all clients. Notably, having clients send gradients instead of model updates can achieve equivalent results, as demonstrated in the derivation of FedAvg~\citep{mcmahan2017communication}. The server then distributes the updated global model parameters to all clients for the next round of training.

\begin{table*}
\centering
\renewcommand\arraystretch{1.2}
\caption{Benchmark Algorithms in MobenFL.}
\label{Benchmark_Methods}
\begin{tabular}{llcc} 
\hline
\multicolumn{2}{c}{\textbf{Method}}                                                                      & \textbf{Year} & \textbf{	Venue}  \\ 
\hline
                                               & FedAvg~\citep{mcmahan2017communication} & 2017          & AIS               \\
\hline
\multirow{4}{*}{\textbf{Additional Network Structures}} & FedNP~\citep{wu2023fednp}               & 2023          & AAAI               \\
                                               & FedALA~\citep{zhang2023fedala}          & 2023          & AAAI             \\
                                               & GPFL~\citep{zhang2023gpfl}              & 2023          & ICCV              \\
                                               & TurboSVM~\citep{wang2024turbosvm}       & 2024          & AAAI               \\ 
\hline
\multirow{5}{*}{\textbf{Regularization-Driven}} & FedProx~\citep{li2020federated}         & 2021          & MLsys              \\
                                               & MOON~\citep{li2021model}                & 2021          & CVPR              \\
                                               & HarmoFL~\citep{jiang2022harmofl}                & 2022          & AAAI              \\
                                               & PGFed~\citep{luo2023pgfed}              & 2023          & ICCV              \\
                                               & FedDecorr~\citep{10336535}              & 2024          & TPAMI              \\ 
\hline                         
\multirow{7}{*}{\textbf{Aggregation Strategies}} & FedBN~\citep{li2021fedbn}               & 2021          & ICLR             \\
                                               & FedCD~\citep{kopparapu2020fedcd}        & 2021          & KDD              \\
                                               & FedGH~\citep{yi2023fedgh}               & 2023          & ACMM                \\
                                               & ProxyFL~\citep{kalra2023decentralized}  & 2023          & NC              \\
                                               & FCCL~\citep{huang2023generalizable}     & 2024          & TPAMI              \\
                                               & FedAu~\citep{wang2024a}                 & 2024          & ICLR              \\ 
                                               & PraVFed~\citep{10843771}                 & 2025          & TIFS              \\ 
\hline
\multirow{3}{*}{\textbf{Split Learning}}                & FLOP~\citep{yang2021flop}               & 2021          & KDD             \\
                                               & SplitFed~\citep{thapa2022splitfed}      & 2022          & AAAI              \\
                                               & SplitAvg~\citep{zhang2022splitavg}      & 2022          & JBHI             \\
\hline
\end{tabular}
\end{table*}

\subsection{Benchmark Algorithms}

The standard FedAvg serves merely as a foundational framework, and its practical application is often constrained by issues such as Non-IID client data and difficulties in model convergence~\citep{tan2022towards}. To enhance the performance, efficiency, and stability of FL in complex real-world environments, researchers have proposed effective improvements from multiple dimensions. In our study, to systematically evaluate these algorithms, we categorize them into four major classes: Enhancing model expressiveness by introducing \textit{Additional Network Structures}; Employing \textit{Regularization-Driven} methods to constrain local training and mitigate client drift; Designing more advanced \textit{Aggregation Strategies} to optimize the efficiency and fairness of server-side aggregation; Leveraging the \textit{Split Learning} architecture to fundamentally alter the paradigm of computation and communication. These benchmark methods are summarized in Table \ref{Benchmark_Methods}. Below, we will elaborate on these four categories of methods.

\subsubsection{Additional Network Structures}
Incorporating additional network structures (e.g., attention mechanisms or auxiliary classifiers) into FL can enhance model performance but increases computational and communication overhead. The advantage lies in better handling non-IID data and mitigating client drift issues, while the drawback involves higher computational costs that may affect deployment efficiency on resource-constrained devices. A trade-off between performance gains and additional costs must be considered. FedNP, FedALA, GPFL, and TurboSVM all fall into this category.

\begin{itemize}
    \item \textbf{FedNP}: FedNP employs a probabilistic neural network inspired by Expectation Propagation (EP), which enables efficient estimation of the global data distribution under given non-IID data partitions. This estimation does not rely on direct access to the global data but is inferred through local data and models on each device. The FedNP algorithm is sampling-free, and the entire framework is end-to-end differentiable, allowing it to be integrated with any traditional federated learning framework to learn richer global data representations. By introducing auxiliary tasks to regularize local models, FedNP prevents local models from overfitting to local data distributions, thereby enhancing the generalization capability of the global model.

    \item \textbf{FedALA}: The core of FedALA is an Adaptive Local Aggregation (ALA) module. Before each training iteration, this module adaptively aggregates the downloaded global model and the local model based on local objectives to initialize the local model. Unlike previous methods, FedALA does not simply overwrite the local model with the global model but achieves aggregation through element-wise weighting, enabling more precise capture of the desired information from the global model.

    \item \textbf{GPFL}: The GPFL framework introduces Global Category Embeddings (GCE) and a Conditional Valve (CoV) module to simultaneously learn global and personalized feature representations. GCE stores global embedding vectors for each category, guiding feature extraction by computing similarity with client data while remaining frozen during training. The CoV module dynamically weights feature vectors to separate global and personalized representations. During local training, clients optimize both global and personalized loss functions, enabling joint learning of shared and user-specific features. This approach effectively overcomes the limitations of traditional and personalized federated learning by balancing global knowledge aggregation and local adaptation.

    \item \textbf{TurboSVM}: TurboSVM employs Support Vector Machines (SVM) to achieve selective aggregation and maximum margin extension regularization. Specifically, during global updates, instead of simply averaging client model parameters, TurboSVM first performs selective aggregation on client class embeddings using SVM, retaining only those embeddings that contribute most significantly to the global model. It then applies maximum margin extension regularization to the aggregated embeddings to accelerate convergence and enhance model generalization. Additionally, the algorithm adopts a ``model-as-sample" strategy by treating each client model as a data sample and leveraging SVM's decision boundary to distinguish beneficial from detrimental updates.

\end{itemize}

\subsubsection{Regularization-Driven}
Employing regularization-driven approaches to enhance the generalizability of local learning and thereby calibrate client drift is an effective strategy. Typically, such methods utilize self-distillation paradigms or modified cross-entropy terms for implementation. Although these methods avoid the additional cost of sharing signals, they tend to be sensitive to hyperparameters and can be fragile in scenarios with extremely heterogeneous data, making it difficult to maintain stable performance. FedProx, MOON, HarmoFL, PGFed and FedDecorr all fall into this category.

\begin{itemize}
    \item \textbf{FedProx}: FedProx introduces a proximal term into the local optimization objective to constrain the deviation between local updates and the global model. This proximal term not only helps address statistical heterogeneity but also enables safe aggregation of partial updates from diverse devices. By adjusting the weight parameter of the proximal term, FedProx achieves more stable and accurate convergence in highly heterogeneous scenarios.

    \item \textbf{MOON}: The MOON framework innovatively combines model contrastive learning with regularization methods by introducing a contrastive loss term based on the similarity between global and local model representations, thereby establishing a novel regularization mechanism. The framework employs a lightweight three-layer architecture (base encoder-projection head-output layer) that uses model-level contrastive learning as a regularization constraint. This approach effectively mitigates client-side training drift caused by Non-IID data distribution while maintaining model performance. The regularization strategy not only alleviates the negative impact of data heterogeneity but also prevents common overfitting issues seen in traditional methods.

    \item \textbf{HarmoFL}: HarmoFL proposes an amplitude normalization strategy. It transforms images into the frequency domain and normalizes the amplitude components to simulate uniform imaging settings, thereby creating a harmonious feature space among local clients. This approach reduces the heterogeneity of feature distributions without violating the preservation of local raw data. Based on the normalized features, HarmoFL designs a weight perturbation strategy to address global drift. This strategy guides each local model to reach a flat optimal point, ensuring uniformly low loss in the neighborhood of the local optimal solution. The perturbation is generated locally without additional communication costs and helps the global model converge to an optimal solution by aggregating multiple local flat optimal solutions.

    \item \textbf{PGFed}: The PGFed framework achieves more efficient personalized learning by explicitly constructing each client's local objective function to directly incorporate the empirical risk from other clients. Specifically, PGFed defines each client's local objective function as the sum of its local empirical risk and an auxiliary risk term—a first-order approximation of other clients' empirical risks. This approach enables PGFed to explicitly leverage information from other clients to guide personalized model training without incurring additional communication overhead.

    \item \textbf{FedDecorr}: FedDecorr addresses dimension collapse by incorporating a regularization term that minimizes the Frobenius norm of the representation covariance matrix during local training. Specifically, during each local iteration, FedDecorr computes the covariance matrix of model representations and adds its Frobenius norm as a regularization term to the local objective function. This approach effectively reduces the risk of dimension collapse while preserving the model's representation capacity.

\end{itemize}

\subsubsection{Aggregation Strategies}
The optimization of FL aggregation strategies focuses on several key aspects, such as dynamic weight allocation, robustness enhancement, and personalized aggregation, in addition to strategies like asynchronous aggregation and security and privacy protection. Its advantages lie in low implementation costs and strong compatibility, requiring only adjustments to the server-side logic without increasing the computational burden on clients. However, this strategy has limited capabilities in handling complex non-independent and identically distributed (non-IID) data and may affect performance due to simplified assumptions. FedBN, FedCD, FedGH, ProxyFL, FCCL, FedAU and PraVFed all fall into this category.

\begin{itemize}
    \item \textbf{FedBN}: FedBN enhances aggregation strategy by excluding the sharing of client-side BN layer parameters (mean and variance), only aggregating non-BN layer parameters to preserve local feature distributions. Each client adapts to its data characteristics through local BN layers, effectively mitigating the negative impact of global averaging on feature shift.

    \item \textbf{FedCD}: The core methods of FedCD consist of two components: local drift detection and adaptation, as well as global model aggregation optimization. Clients monitor the concept drift of their local data through a drift detection module and dynamically adjust their local training strategies upon detecting changes. On the server side, based on the drift feedback information from clients, higher weights are assigned to stable clients during model aggregation, thereby mitigating the negative impact of clients affected by drift on the global model.

    \item \textbf{FedGH}: FedGH stands out due to its innovative support for model heterogeneity, which allows participants to employ diverse model architectures by sharing only feature representations and the global prediction head. This approach not only reduces communication overhead, as only the parameters of the feature representations and the global prediction head need to be transmitted instead of the entire model, but also lowers computational costs associated with calculating local average representations.

    \item \textbf{ProxyFL}: ProxyFL achieves decentralized federated learning by having each participant maintain two models: a private model and a proxy model. The private model is used for training on local data, while the proxy model is used for sharing information with other participants. During training, participants use differential privacy techniques to train their proxy models and send them to their neighboring participants. Upon receiving the proxy models from neighbors, participants update their own proxy models by aggregating these received models, thereby avoiding the direct sharing of private data or model parameters and protecting data privacy.

    \item \textbf{FCCL}: The FCCL method achieves knowledge communication between heterogeneous models by constructing cross-correlation matrices and aligning instance similarity distributions at the feature level, thereby overcoming communication barriers and enhancing the generalization ability of the models. Specifically, this method comprises three core modules: the Federal Cross-Correlation Matrix (FCCM) module, which utilizes public data with irrelevant labels to compute the cross-correlation matrix between model outputs of different participants and optimizes this matrix to enhance correlation among the same dimensions while reducing correlation among different dimensions, thereby learning class-invariant features; the Federal Instance Similarity Learning (FISL) module, which calculates the similarity distribution between feature vectors and aligns this distribution across different participants, realizing feature-level knowledge communication; and the Federal Non-Target Distillation (FNTD) module, which, during the local update phase, decomposes knowledge distillation into target distillation and non-target distillation, preserving cross-domain knowledge through non-target distillation while avoiding optimization conflicts, effectively mitigating the problem of catastrophic forgetting.

    \item \textbf{FedAU}: The core of the FedAU method lies in dynamically adjusting the aggregation weights of model updates for each client based on their participation history. Specifically, building upon FedAvg, FedAU introduces a pluggable aggregation weight calculation module. This module online estimates the optimal aggregation weight for each client in the current round based on their past participation. By doing so, FedAU can adapt to changes in client participation rates without increasing additional communication overhead or memory requirements, thereby improving the performance of federated learning.

    \item \textbf{PraVFed}: PraVFed enables passive parties to perform multiple rounds of local pre-training using labels perturbed with differential privacy, thereby extracting local feature embeddings, and introduces blinding factors to protect these embeddings from revealing raw features. The active party collects all blinded embeddings from the passive parties, computes weights based on the performance of each passive party's pre‑trained model, and performs weighted aggregation to obtain a global embedding. This global embedding is then combined with the active party's own feature embedding to jointly train the global model.

\end{itemize}

\subsubsection{Split Learning}
Split Learning enables collaborative training by vertically partitioning a neural network model into two components: the client and the server. In this setup, the client is responsible for computing the initial layers and uploading the intermediate features, while the server completes the remaining computations and returns the gradients. This approach offers several advantages, including strong privacy preservation, as it only transmits intermediate features rather than raw data, effectively safeguarding data privacy; it also imposes a lower computational load, as the client only needs to run a portion of the model, thereby reducing the computational requirements for the client. However, Split Learning also has some drawbacks. For instance, the design of the split layer is highly sensitive and can directly impact the balance between performance and privacy; additionally, frequent interactions between the client and the server may introduce latency issues. FLOP, SplitFed and SplitAvg all fall into this category.

\begin{itemize}
    \item \textbf{FLOP}: The core idea of FLOP is to keep the parameters of the pre-trained model frozen and only learn and update prompts on the client side. These prompt parameters are uploaded to the server during communication, aggregated, and then redistributed to the clients. To further enhance personalization performance, FLOP combines two types of prompt structures: global prompts and local prompts, which are responsible for learning globally shared information and client-specific features, respectively.

    \item \textbf{SplitFed}: SplitFed innovatively integrates federated learning with split learning for the first time, vertically partitioning the neural network model into client-side (for shallow feature extraction) and server-side (for deep computation) components. This approach allows clients to run only a portion of the model, significantly reducing resource requirements. Meanwhile, SplitFed incorporates the parallel training mechanism of federated learning, enabling multiple clients to interact with the server simultaneously and circumventing the sequential training bottleneck inherent in split learning.

    \item \textbf{SplitAvg}: SplitAVG divides the deep network into client sub-networks and server sub-networks: Local institutions process raw data using client sub-networks to generate intermediate feature maps, which are sent to a central server for concatenation. The server sub-network performs forward propagation based on the concatenated features and propagates gradients back to local clients via backward propagation to update parameters. This method mitigates data distribution discrepancies through feature map concatenation while significantly reducing communication overhead.

\end{itemize}

\subsection{Multi-organ Datasets}

In advancing the field of FL for medical imaging analysis, a core challenge lies in the lack of standardized benchmarks that fully capture the complexity and diversity of real-world applications. To address this gap, the integrated dataset in this benchmark offers remarkable breadth and clinical relevance. Its uniqueness stems from the inclusion of diverse imaging data spanning 12 organs across the human body, comprising a total of 22 independent datasets, whose detailed information is illustrated in Fig.~\ref{datasets}. This design enables, for the first time, a systematic evaluation of FL algorithms under a unified framework, assessing their robustness and generalization capabilities when confronted with varied anatomical sites, disease types, imaging devices, and data distributions. Next, we will provide a detailed introduction to each of these 22 datasets.

\begin{table*}
\centering
\renewcommand\arraystretch{1.2}
\caption{The statistical information of the datasets in MobenFL.}
\label{datasets}
\begin{tabular}{llp{4cm}ccc} 
\hline
          \textbf{Tissue}        & \textbf{Dataset}     & \textbf{Task} & \textbf{Examples} & \textbf{Classes} & \textbf{Imbalance}  \\ 
\hline

\multirow{10}{*}{\textbf{Brain}} & ADNI~\citep{petersen2010alzheimer}               & \footnotesize{Classification of Alzheimer's Disease.}          & 15686             &          3        &            2.32               \\
                                               & ADHD~\citep{adhd2012adhd}          & \footnotesize{Classification of Attention Deficit Hyperactivity Disorder.}           & 767      &          2        &            1.34        \\
                                               & PPMI~\citep{marek2018parkinson}              & \footnotesize{Classification of Parkinson's Disease.}          & 799        &          2        &            2.26      \\
                                               & ABIDE-T1~\citep{di2014autism}       & \footnotesize{Classification of Autism Spectrum Disorder.}          & 1112         &          2        &            1.06       \\ 
                                               & ABIDE-fMRI~\citep{di2014autism}       & \footnotesize{Classification of Autism Spectrum Disorder.}          & 1112         &          2        &            1.06     \\ 
                                               & ASDFace       & \footnotesize{Classification of Autism Spectrum Disorder.}          & 2896        &          2        &            1.00       \\ 
\hline
\multirow{3}{*}{\textbf{Eyes}}                 & Retinopathy~\citep{Tianchi}         & \footnotesize{Classification of Retinal Disease.}          & 35126         &         5         &         16.49              \\
                                               & Papilledema                & \footnotesize{Classification of Causes of Optic Disc Edema.}          & 1369      &         3         &         2.09          \\
                                               & Refuge~\citep{orlando2020refuge}               & \footnotesize{Classification of Glaucoma.}          & 1200          &         2         &         4.97    \\

\hline                         
\multirow{1}{*}{\textbf{Oral}} & OSCC~\citep{rahman2018textural}              & \footnotesize{Classification of Oral Squamous Cell Carcinoma Pathological Images.}          & 1224         &         2         &         2.11     \\

\hline
\multirow{1}{*}{\textbf{Breast}}                & BRACS~\citep{brancati2022bracs}               & Classification of Breast Cancer Subtypes.          & 4539           &       7           &             1.34             \\

\hline
\multirow{1}{*}{\textbf{Lung}}                & ChestXray~\citep{wang2017chestx}               & Classification of chest X-ray image.          & 112120           &        15          &       57.95      \\

\hline
\multirow{3}{*}{\textbf{Skins}}                & HAM10000~\citep{tschandl2018ham10000}               & \footnotesize{Classification of Skin Diseases.}          & 10015    & 7    & 21.74     \\
                                               & ISIC2020~\citep{rotemberg2021patient}               & \footnotesize{Classification of Melanocytic Lesions.}          & 33126   & 2    & 28.36          \\
                                               & PADUFES~\citep{pacheco2020pad}               & \footnotesize{Classification of Skin Diseases.}         & 2298      & 6    & 4.97       \\

\hline
\multirow{1}{*}{\textbf{Blood Cell}}                & BCCD               & \footnotesize{Classification of Blood Cell.}          & 12444       &        4          &         1.00       \\

\hline
\multirow{1}{*}{\textbf{Abdomen}}                & OrganMNIST~\citep{yang2023medmnist}               & \footnotesize{Classification of Abdominal Organs.}           &  58830            &       11           &          2.45             \\

\hline
\multirow{1}{*}{\textbf{Kidney}}                & TissueMNIST~\citep{yang2023medmnist}               & \footnotesize{Classification of Kidney Cortex Cell.}           & 236386            &       8           &          4.39            \\

\hline
\multirow{1}{*}{\textbf{Colon}}                & NCH100K~\citep{kather_jakob_nikolas_2018_1214456}               & \footnotesize{Classification of Colorectal Cancer Histopathology Images.}          & 100000             &        9          &           1.32          \\

\hline
\multirow{1}{*}{\textbf{Colorectal}}                & LIMUC~\citep{polat2022improving}               & \footnotesize{Classification of Endoscopic Images for Colitis Severity.}           & 11276           &        4          &            3.73           \\

\hline
\multirow{2}{*}{\textbf{Knee}}                & KneeXray~\citep{gornale_shivanand_2020_41241861}               & \footnotesize{Classification of knee X-ray  image.}           & 1650             &          5        &             1.82          \\
                                              & Arthrosis               & \footnotesize{Classification of Finger Joints.}           & 28402           &          9        &             1.55   \\

\hline
\end{tabular}
\end{table*}

\begin{itemize}
    \item \textbf{Brain:}
    \begin{enumerate}
        \item \textbf{ADNI}$\footnote{https://adni.loni.usc.edu}$: We first processed the 3D brain MRI images from the ADNI dataset using FSL tools$\footnote{https://fsl.fmrib.ox.ac.uk/fsl}$, performing registration, skull-stripping, and normalization. Subsequently, we selected three slices along the z-axis of the MRI images: the central cross-section, as well as the cross-sections 5 slices above and below the central plane. These three slices were then combined to generate a three-channel image. The dataset includes a total of 15,686 samples across three categories: normal controls, mild cognitive impairment (MCI), and Alzheimer's disease (AD), with a class imbalance ratio of 2.32.
        
        \item \textbf{ADHD}$\footnote{https://fcon\_1000.projects.nitrc.org/indi/adhd200}$: The ADHD dataset is derived from the ADHD200 open-access neuroimaging dataset, which includes resting-state functional magnetic resonance imaging (rs-fMRI), structural MRI, and clinical phenotypic data from 973 participants across 8 international sites. We selected 767 available structural MRI samples from this dataset. The ADHD data processing follows the same approach as ADNI, with a data imbalance ratio of 1.34.
        
        \item \textbf{PPMI}$\footnote{https://www.ppmi-info.org}$: The PPMI dataset includes a multimodal dataset from approximately 50 sites worldwide with 4,000 volunteers. We selected structural MRI data from 799 of these volunteers for evaluation, with data processing methods consistent with those of ADNI. The imbalance ratio is 2.26.

        \item \textbf{ABIDE-T1}$\footnote{https://fcon\_1000.projects.nitrc.org/indi/abide/abide\_I.html}$: ABIDE-T1 is derived from the ABIDE I dataset, which integrates data from 17 international research centers sharing previously collected fMRI, anatomical data, and phenotypic datasets. This initiative ultimately aggregated 1,112 datasets, including 539 from individuals with Autism Spectrum Disorder (ASD) and 573 from typical development controls. We selected the structural MRI data from this collection for evaluation, which exhibits an imbalance ratio of 1.06.

        \item \textbf{ABIDE-fMRI}: ABIDE-fMRI is derived from the fMRI modality within the ABIDE I dataset. We extracted blood oxygen level-dependent (BOLD) signal sequences from the fMRI data, then constructed multichannel functional connectivity images by segmenting the sequences and computing brain functional connectivity matrices.

        \item \textbf{ASDFace}$\footnote{https://www.kaggle.com/discussions/general/123978}$: ASDFace comprises 2,896 images, with an equal number of facial images of children with autism spectrum disorder and non-autistic children. The dataset exhibits a balanced distribution with an imbalance ratio of 1.
    \end{enumerate}

        \item \textbf{Eyes:}
    \begin{enumerate}
        \item \textbf{Retinopathy}\footnote{https://www.kaggle.com/datasets/tanlikesmath/diabetic-retinopathy-resized/data}: Retinopathy is a dataset used for diabetic retinopathy classification, sourced from the APTOS\footnote{https://www.kaggle.com/c/aptos2019-blindness-detection/data} 2019 Blindness Detection challenge. The Retinopathy dataset contains 35,126 JPEG-format fundus images, labeled with 5 severity grades ranging from "No DR" to "Proliferative DR," with a class imbalance ratio of 16.49.
        
        \item \textbf{Papilledema}\footnote{https://osf.io/2w5ce/}: The Papilledema dataset comprises 1,369 PNG format images categorized into three classes: normal, papilledema, and pseudo papilledema. The dataset exhibits an imbalance ratio of 2.09.
        
        \item \textbf{Refuge}\footnote{https://refuge.grand-challenge.org/}: The Refuge dataset was sourced from the Retinal Fundus Glaucoma Challenge held at MICCAI 2018. Refuge publicly released a dataset comprising 1,200 fundus images, which includes expert-annotated segmentations and clinical glaucoma labels. The dataset exhibits an imbalance ratio of 4.97.

    \end{enumerate}

            \item \textbf{Oral:}
    \begin{enumerate}
        \item \textbf{OSCC}\footnote{https://data.mendeley.com/datasets/ftmp4cvtmb/1}: The OSCC dataset is specifically designed for the task of oral squamous cell carcinoma (OSCC) image classification in pathology. It comprises a total of 1,224 pathological images, divided into two sets with different resolutions. The first set includes 89 images of normal oral epithelial tissue and 439 OSCC images at 100x magnification. The second set contains 201 images of normal epithelial tissue and 495 OSCC pathological images at 400x magnification. The imbalance ratio of the dataset is 2.11.
        
    \end{enumerate}
    
        \item \textbf{Breast:}
    \begin{enumerate}
    \item \textbf{BRACS}$\footnote{https://www.bracs.icar.cnr.it/download}$: BRACS dataset is specifically designed for breast cancer subtyping, containing a large number of stained histopathological images. The dataset comprises 547 whole-slide images (WSIs) and 4,539 regions of interest (ROIs) extracted from these slides. Each whole-slide image and its corresponding ROIs were annotated by three board-certified pathologists, classified into different lesion types. Specifically, BRACS includes three main lesion categories—benign, malignant, and atypical—further divided into seven subtypes, with a class imbalance ratio of 1.34.
        
    \end{enumerate}

        \item \textbf{Lung:}
    \begin{enumerate}
    \item \textbf{ChestXray}$\footnote{https://www.kaggle.com/datasets/nih-chest-xrays/data}$: ChestXray dataset consists of 112,120 labeled X-ray images from 30,805 unique patients. This dataset includes 15 disease categories such as emphysema and pneumonia. The labels were generated by the authors using natural language processing to perform text mining of disease classifications from associated radiology reports, with an estimated label accuracy of 90\%. All images in the ChestXray dataset are uniformly sized at 1024×1024 pixels and exhibit extreme class imbalance.
        
    \end{enumerate}

        \item \textbf{Skins:}
    \begin{enumerate}
        \item \textbf{HAM10000}\footnote{https://doi.org/10.7910/DVN/DBW86T}: HAM10000 is a publicly available dermoscopic image dataset containing 10,015 RGB images, covering seven diagnostic categories of pigmented skin lesions including melanoma and dermatofibroma. The data labels are multi-modally validated, with over 50\% of cases confirmed by histopathology. The HAM10000 dataset exhibits class imbalance with an imbalance ratio of 21.74.
        
        \item \textbf{ISIC2020}\footnote{https://challenge2020.isic-archive.com/}: ISIC 2020 is a large-scale dermoscopic image classification dataset released by the International Skin Imaging Collaboration (ISIC). The dataset includes 33,126 images of benign and malignant skin lesions from 2,056 patients. All images were collected from six medical institutions worldwide. The imbalance ratio of the dataset is 28.36.
        
        \item \textbf{PADUFES}\footnote{https://data.mendeley.com/datasets/zr7vgbcyr2/1}: The PADUFES dataset contains 2,298 dermatological images captured by various types of smartphones, collected from 1,373 patients. The dataset covers six categories, three of which are skin cancers and the other three are skin diseases. The imbalance ratio of the dataset is 4.97.

    \end{enumerate}

            \item \textbf{Blood Cell:}
    \begin{enumerate}
    \item \textbf{BCCD}\footnote{https://aistudio.baidu.com/datasetdetail/106627/0}: BCCD dataset is a small-scale dataset for blood cell detection. It consists of 12,500 enhanced blood cell images (JPEG format) with accompanying cell type labels (CSV format). The cell types include eosinophils, lymphocytes, monocytes, and neutrophils, with approximately 3,000 images per category. The BCCD dataset exhibits perfect class balance.
    \end{enumerate}

        \item \textbf{Abdomen:}
    \begin{enumerate}
    \item \textbf{OrganMNIST}\footnote{https://github.com/MedMNIST/MedMNIST}: OrganMNIST originates from the work of Yang et al. in MedMNIST~\citep{yang2023medmnist}. MedMNIST collects various medical imaging data and preprocesses them by resizing the images to dimensions such as 28×28, facilitating algorithm evaluation for researchers. Among these datasets, OrganMNIST is derived from axial slices of 3D computed tomography (CT) images based on the Liver Tumor Segmentation Benchmark (LiTS)~\citep{bilic2023liver}. The organ labels are obtained according to bounding box annotations of 11 abdominal organs from another work~\citep{xu2019efficient}. OrganMNIST contains 58,850 sample slices across 11 abdominal organ classes, with a class imbalance ratio of 2.45.
    \end{enumerate}

        \item \textbf{Kidney:}
    \begin{enumerate}
    \item \textbf{TissueMNIST}$^{8}$: TissueMNIST is derived from the Broad Bioimage Benchmark Collection~\cite{ljosa2012annotated}. The dataset comprises 236,386 human kidney cortex cells, isolated from 3 reference tissue samples and categorized into 8 classes. Each grayscale image has dimensions of 32×32×7 pixels, where 7 represents 7 slices. A 2D maximum projection is generated by taking the maximum pixel value along the axis for each pixel, which is then resized to a 28×28 grayscale image. The class imbalance ratio of TissueMNIST is 4.39.
    \end{enumerate}

        \item \textbf{Colon:}
    \begin{enumerate}
    \item \textbf{NCH100K}$\footnote{https://zenodo.org/records/1214456}$: NCH100K dataset is a specialized pathology dataset for image classification, comprising 100,000 stained histology images of human colorectal cancer and healthy tissues extracted from 86 patients. This dataset includes 9 different types of tissue images, with each tile being a 224x224 pixel image in TIF format and a class imbalance ratio of 1.32.
    \end{enumerate}

        \item \textbf{Colorectal:}
    \begin{enumerate}
    \item \textbf{LIMUC}$\footnote{https://zenodo.org/records/5827695\#.Yi8GJ3pByUk}$: LIMUC dataset is an ulcerative colitis dataset developed by the Gastroenterology Department of Marmara University School of Medicine. This dataset contains 11,276 BMP format images (352×288 pixels) derived from 1,043 colonoscopy procedures performed on 564 patients. All images were classified by two experienced gastroenterologists according to the Mayo Endoscopic Score (MES), with any discordant images independently evaluated by a third expert. The LIMUC dataset encompasses four disease severity categories ranging from mild inflammation to severe inflammation, with a class imbalance ratio of 3.73.
    \end{enumerate}

        \item \textbf{Knee:}
    \begin{enumerate}
    \item \textbf{KneeXray}$\footnote{https://data.mendeley.com/datasets/t9ndx37v5h/1}$: KneeXray dataset comprises 1,650 digital X-ray images of knee joints, with original images stored as 8-bit grayscale. Each knee radiograph was manually annotated by two medical experts according to the Kellgren-Lawrence grading system. The dataset covers five severity categories ranging from normal to severe osteoarthritis, exhibiting a class imbalance ratio of 1.82.

        \item \textbf{Arthrosis}$\footnote{https://www.kaggle.com/datasets/kmader/rsna-bone-age/data}$: The Arthrosis dataset is derived from the 2017 RSNA Pediatric Bone Age Challenge. We selected 28,402 X-ray images along with their corresponding arthrosis categories to form the evaluation dataset. The imbalance ratio of this dataset is 1.55.
    \end{enumerate}
\end{itemize}

\section{Experiments}

\subsection{Experimental setup}
To ensure a fair evaluation of MobenFL's federated learning algorithm, we standardize the training hyperparameters across all experiments: an initial learning rate of 0.0001, batch size of 128, and 5 client nodes. The datasets were split into 80\% training and 20\% testing sets, with training employing an early stopping strategy (patience=10 epochs). All FL implementations used ResNet18~\citep{He_2016_CVPR} architecture with the Adam optimizer. For consistent data preprocessing across all 22 datasets, images underwent horizontal/vertical flipping (50\% probability), random rotation (±10 degrees), and are resized to 224×224 resolution.

\subsubsection{IID setup}
To simulate data heterogeneity in real-world scenarios and validate the robustness of algorithms in complex environments, we utilize the Dirichlet distribution to non-IID data. By constructing a non-IID experimental environment, we can more accurately assess the robustness of algorithms under complex conditions such as skewed data distributions and imbalanced sample sizes, uncover potential issues like client drift, and subsequently drive targeted optimizations (such as personalized federated learning or dynamic weighted aggregation). Next, we will elaborate on the Dirichlet distribution. The Dirichlet distribution~\citep{blei2003latent} is a continuous multivariate probability distribution defined on the probability simplex, commonly used to describe the distribution over multiple categories. Its probability density function is given by:

\begin{equation}
    \text{Dir}(\mathbf{p} \mid \boldsymbol{\alpha}) = \frac{1}{B(\boldsymbol{\alpha})} \prod_{k=1}^K p_k^{\alpha_k - 1},
\end{equation}
where $\mathbf{p} = (p_1, p_2, \dots, p_K)$ is a $K$-dimensional probability vector (i.e., $p_k \geq 0$ and $\sum_{k=1}^K p_k = 1$), representing the class distribution for a particular client. $\boldsymbol{\alpha} = (\alpha_1, \alpha_2, \dots, \alpha_K)$ is the concentration parameter, controlling the sparsity of the distribution:
    \begin{itemize}
        \item If $\alpha_k = 1$, the distribution is uniform across all classes (IID).
        \item If $\alpha_k < 1$, the distribution is sparse, with certain classes dominating (Non-IID).
        \item If $\alpha_k > 1$, the distribution is more concentrated than uniform but less sparse.
    \end{itemize}
$B(\boldsymbol{\alpha})$ is the multivariate Beta function, used for normalization. In our non-IID experiments, the parameter $\alpha$ is set to 0.5.

\subsubsection{Implementation}
All methods are implemented using the Python PyTorch framework. All experiments in this study are conducted on a server equipped cwith 4 RTX 4090 24GB GPUs, running Ubuntu 22.04 as the operating system.

\subsection{Quantitative Single-organ Analysis}\label{exp-1}

\begin{figure*}
    \centering
    \includegraphics[width=0.8\textwidth]{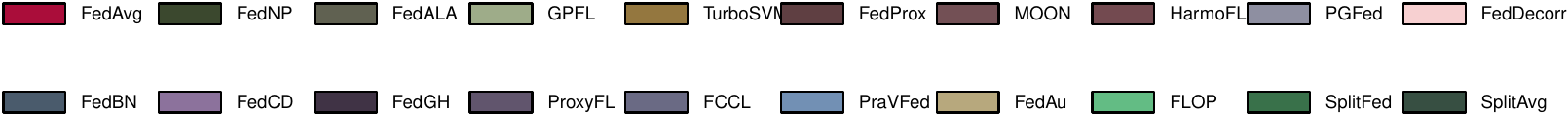}\\
    \includegraphics[width=0.45\textwidth]{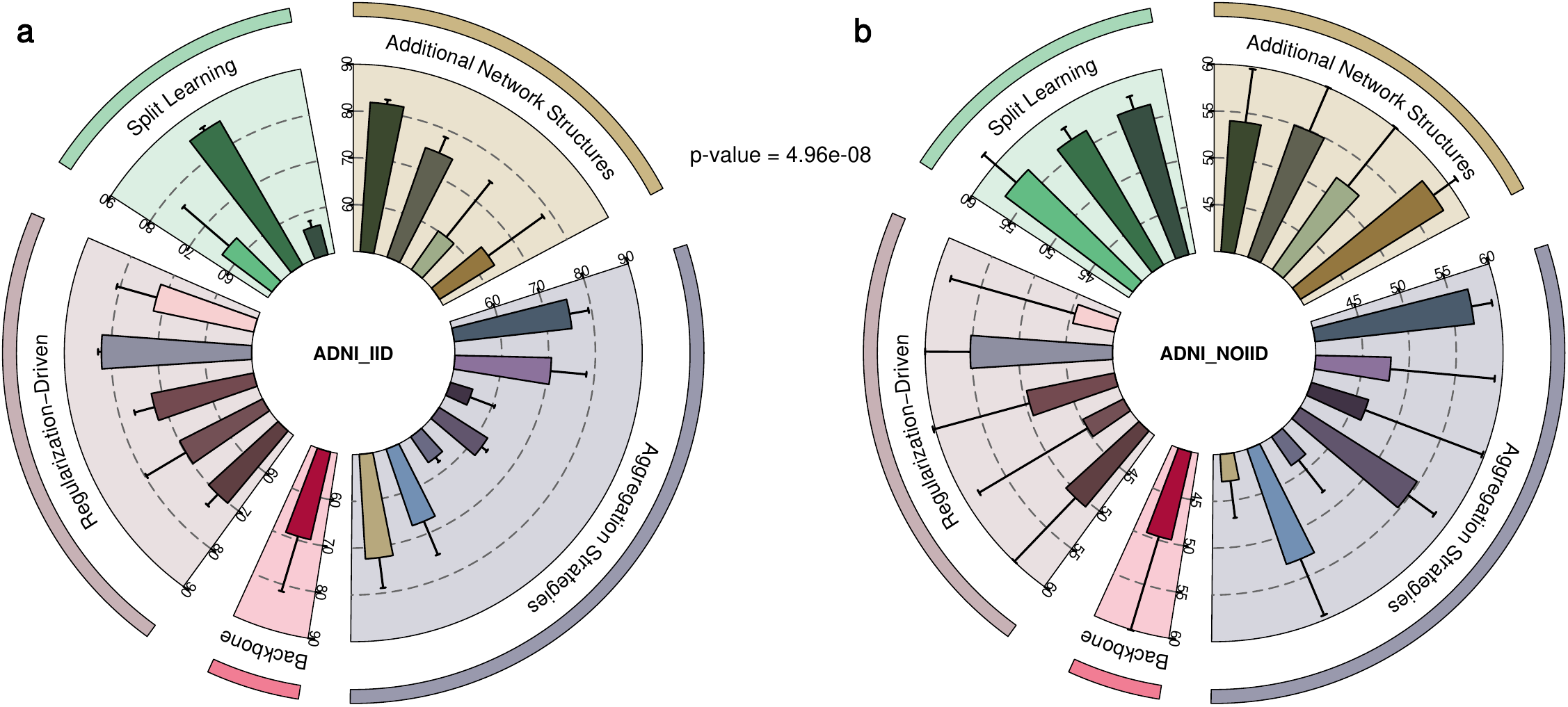}
    \includegraphics[width=0.45\textwidth]{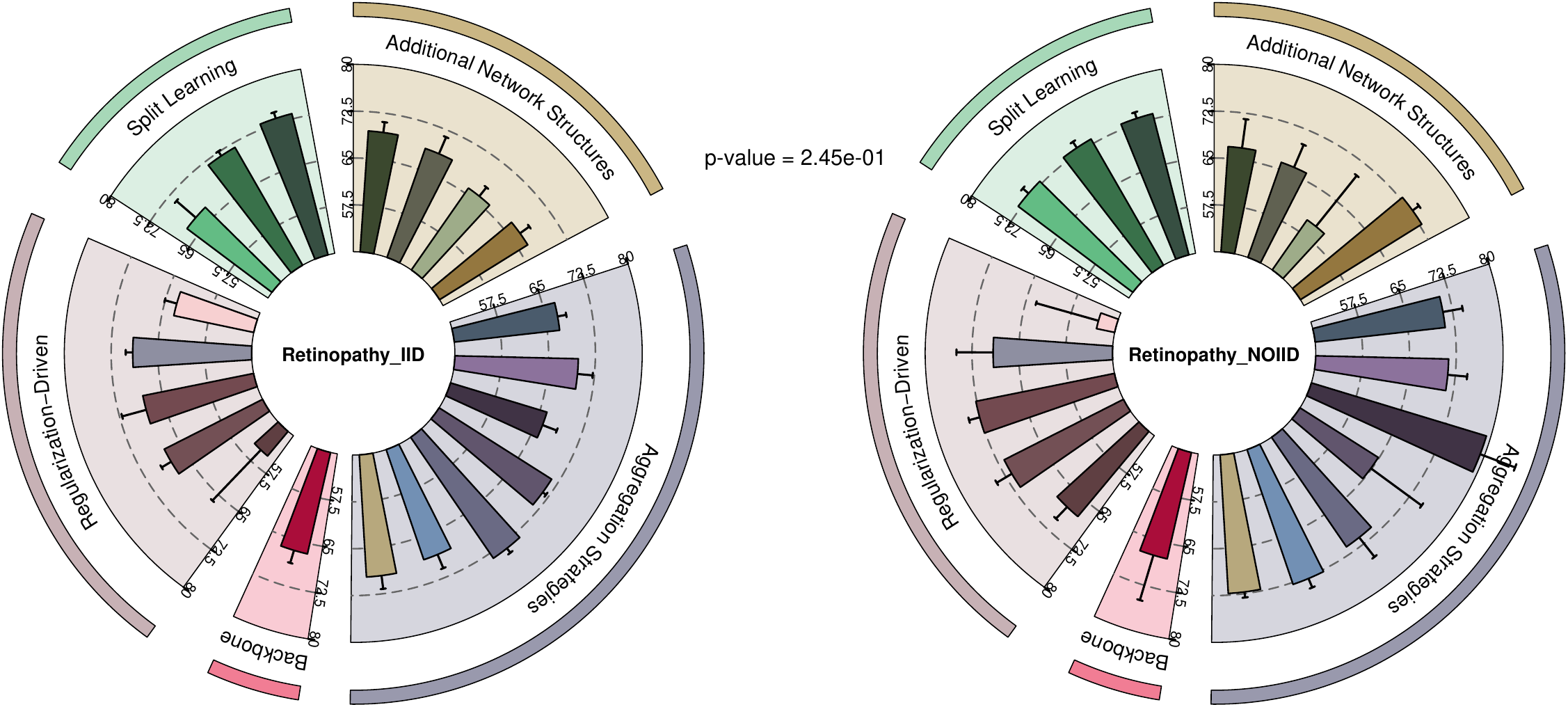} \\
    \includegraphics[width=0.45\textwidth]{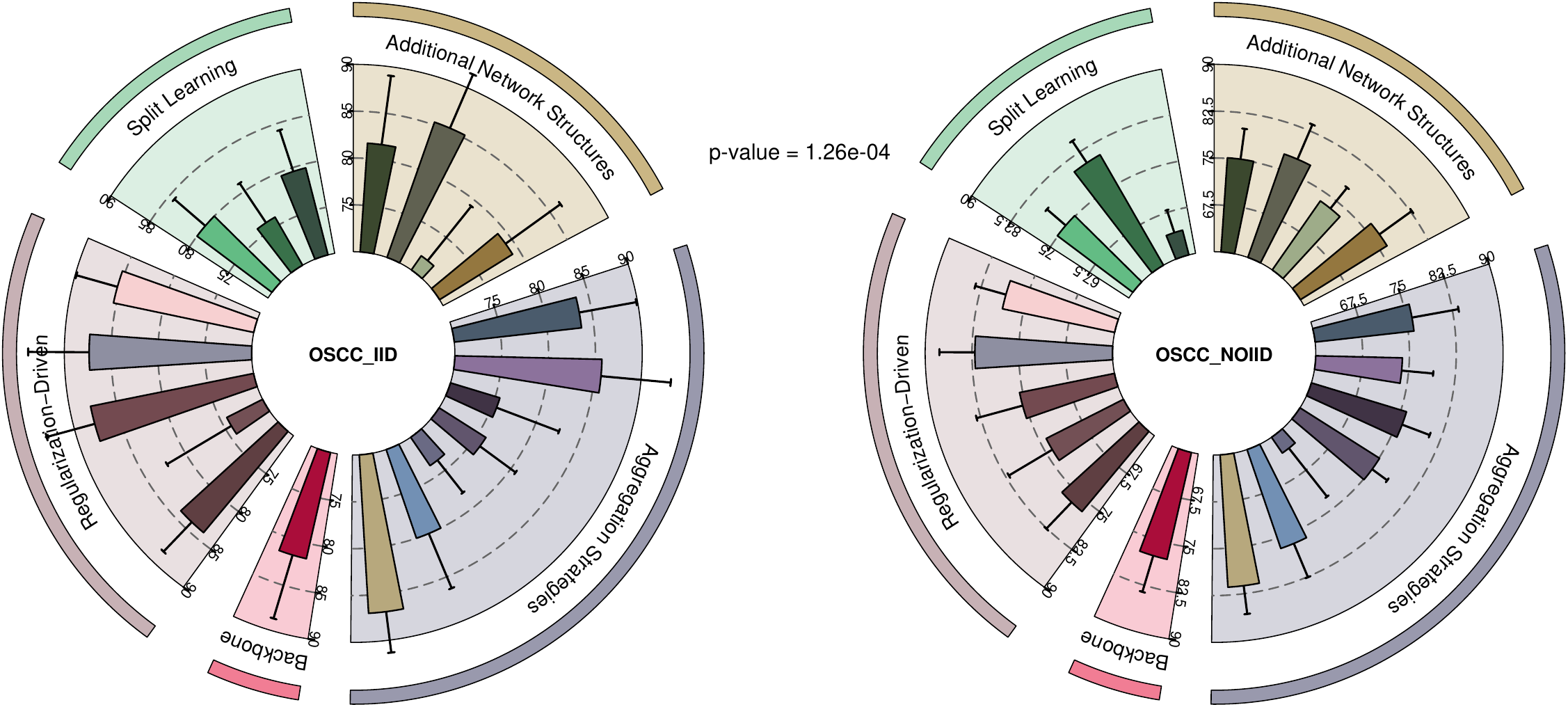}
    \includegraphics[width=0.45\textwidth]{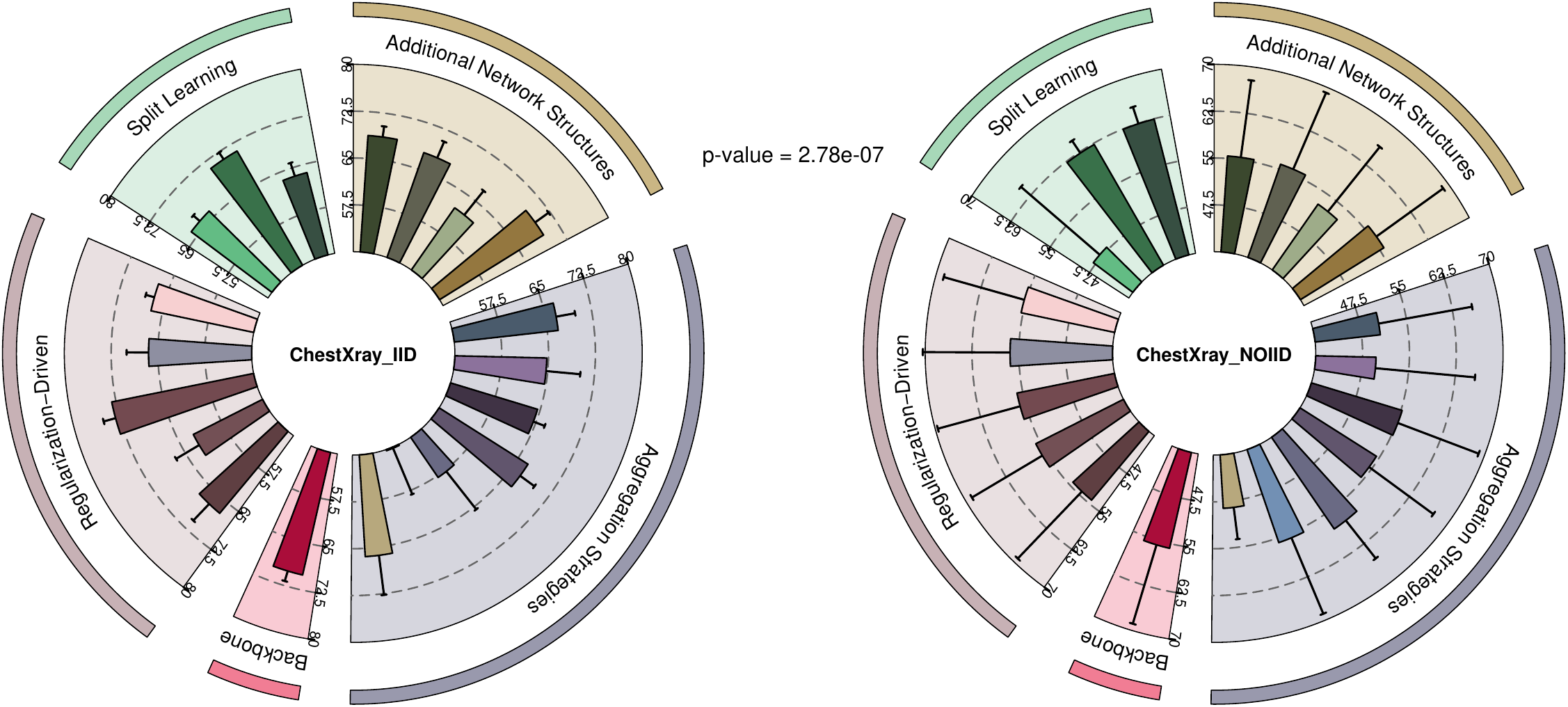} \\
    \includegraphics[width=0.45\textwidth]{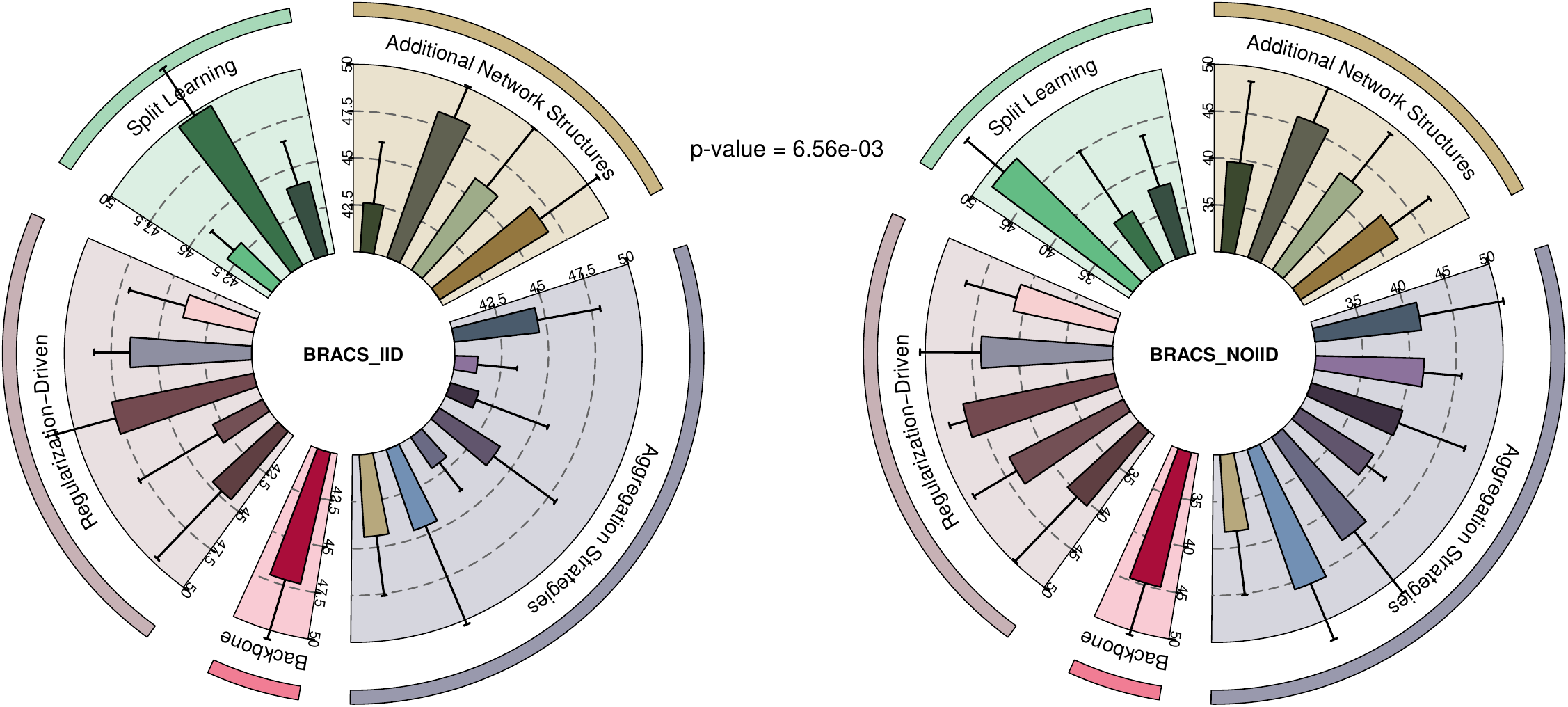} 
    \includegraphics[width=0.45\textwidth]{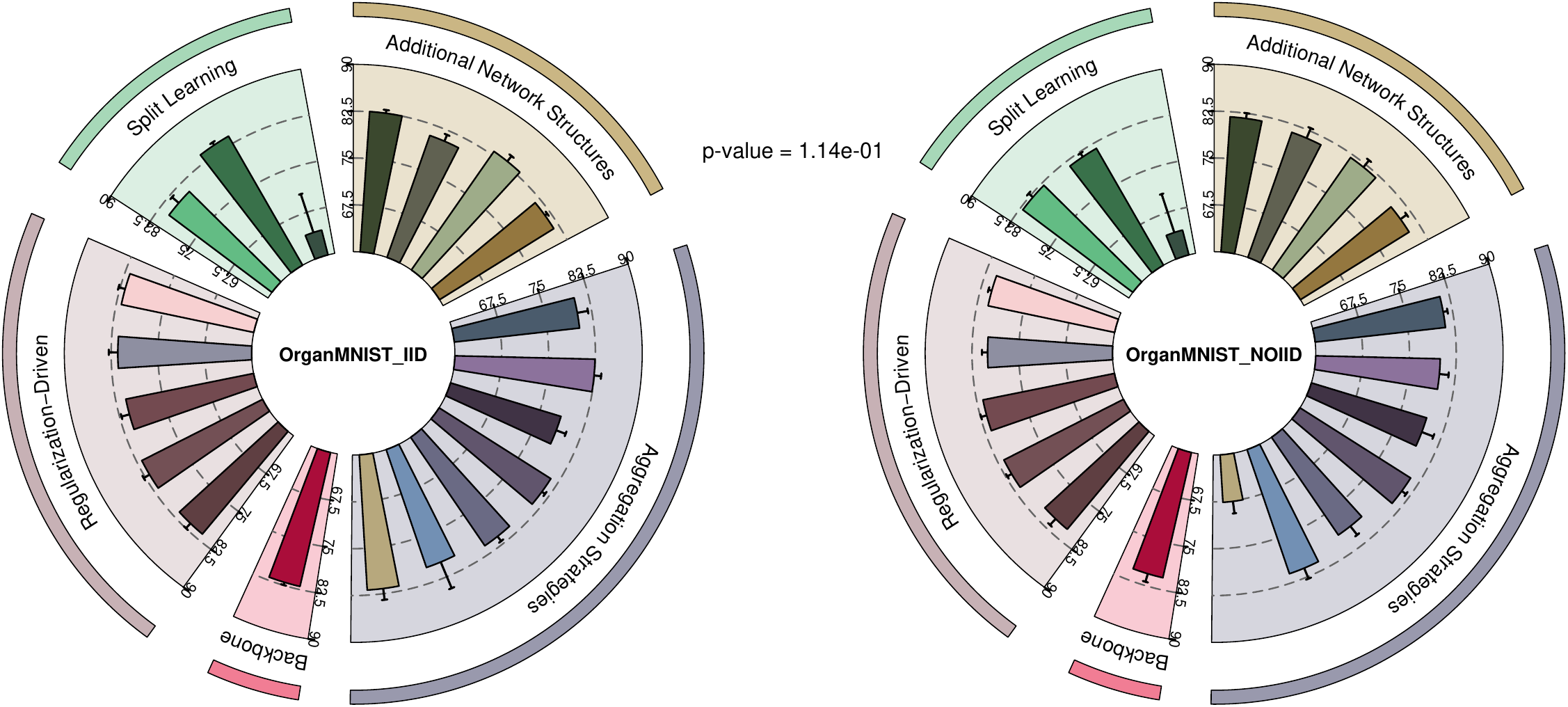}\\
    \includegraphics[width=0.45\textwidth]{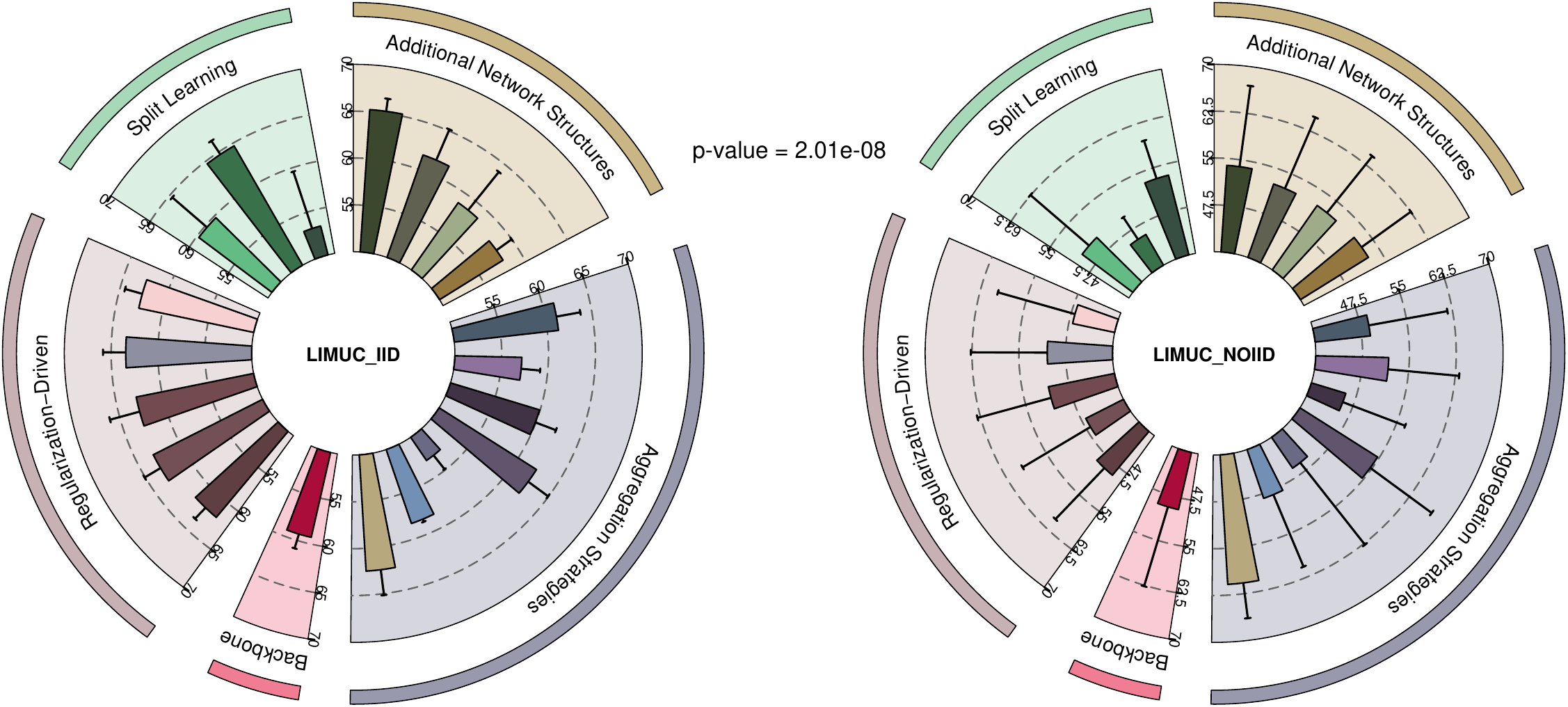} 
    \includegraphics[width=0.45\textwidth]{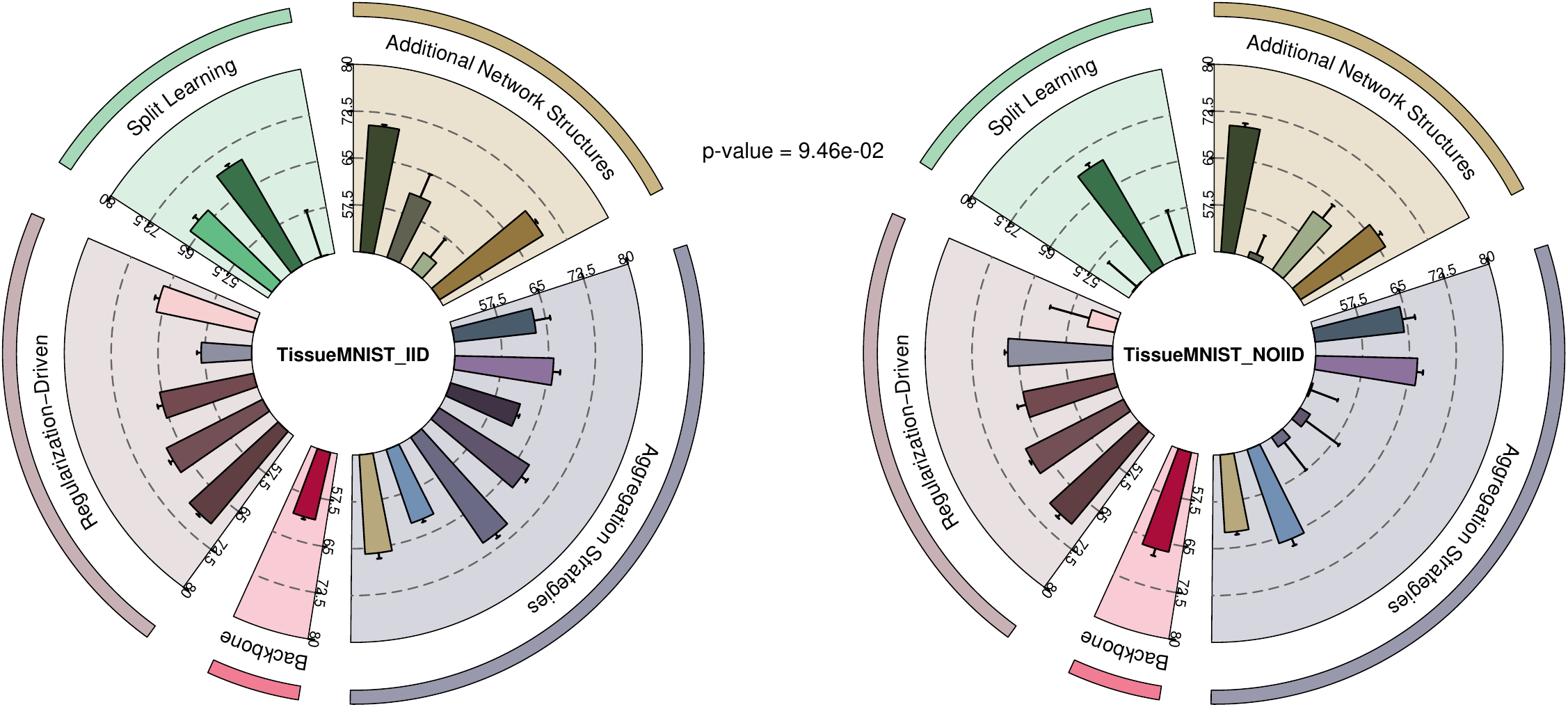} \\
    \includegraphics[width=0.45\textwidth]{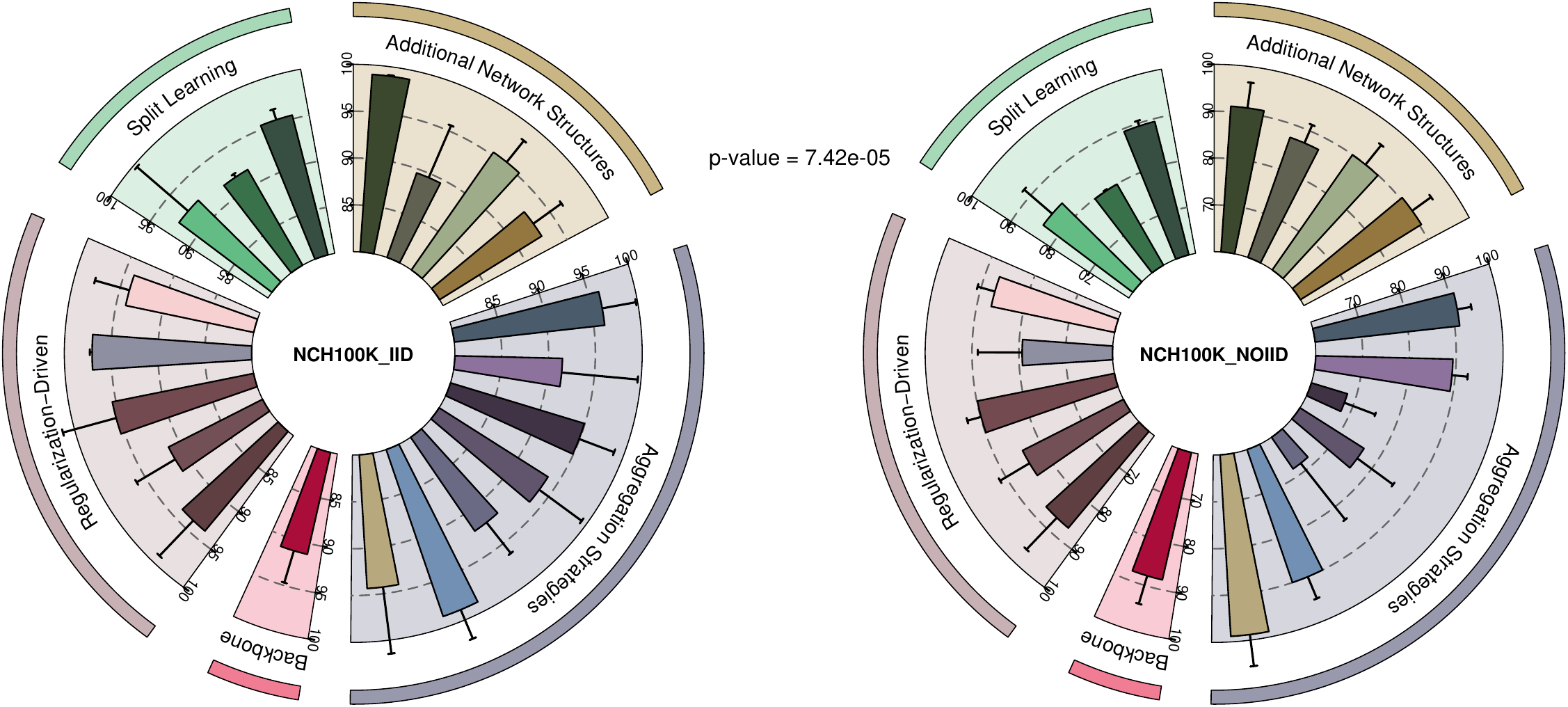}
    \includegraphics[width=0.45\textwidth]{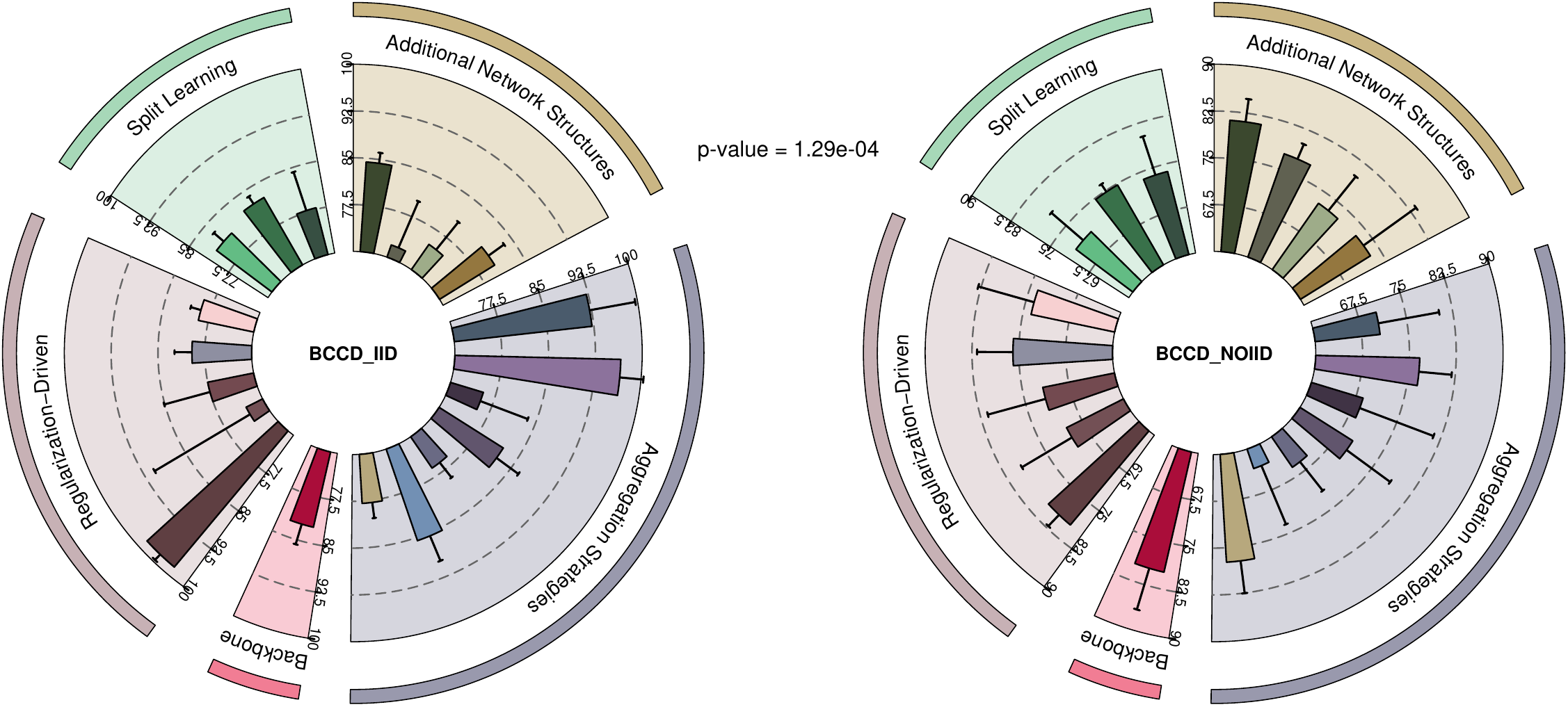} \\
    \includegraphics[width=0.45\textwidth]{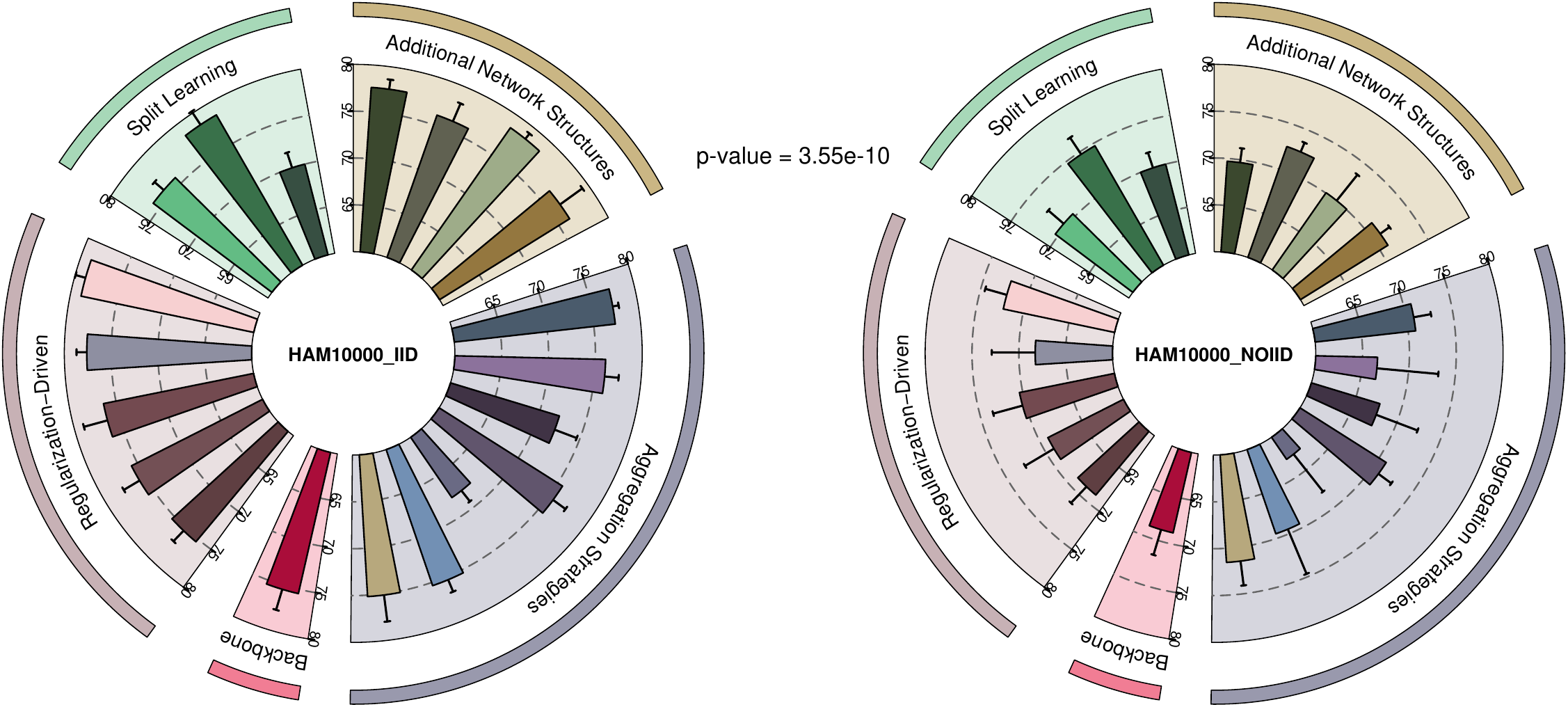} 
    \includegraphics[width=0.45\textwidth]{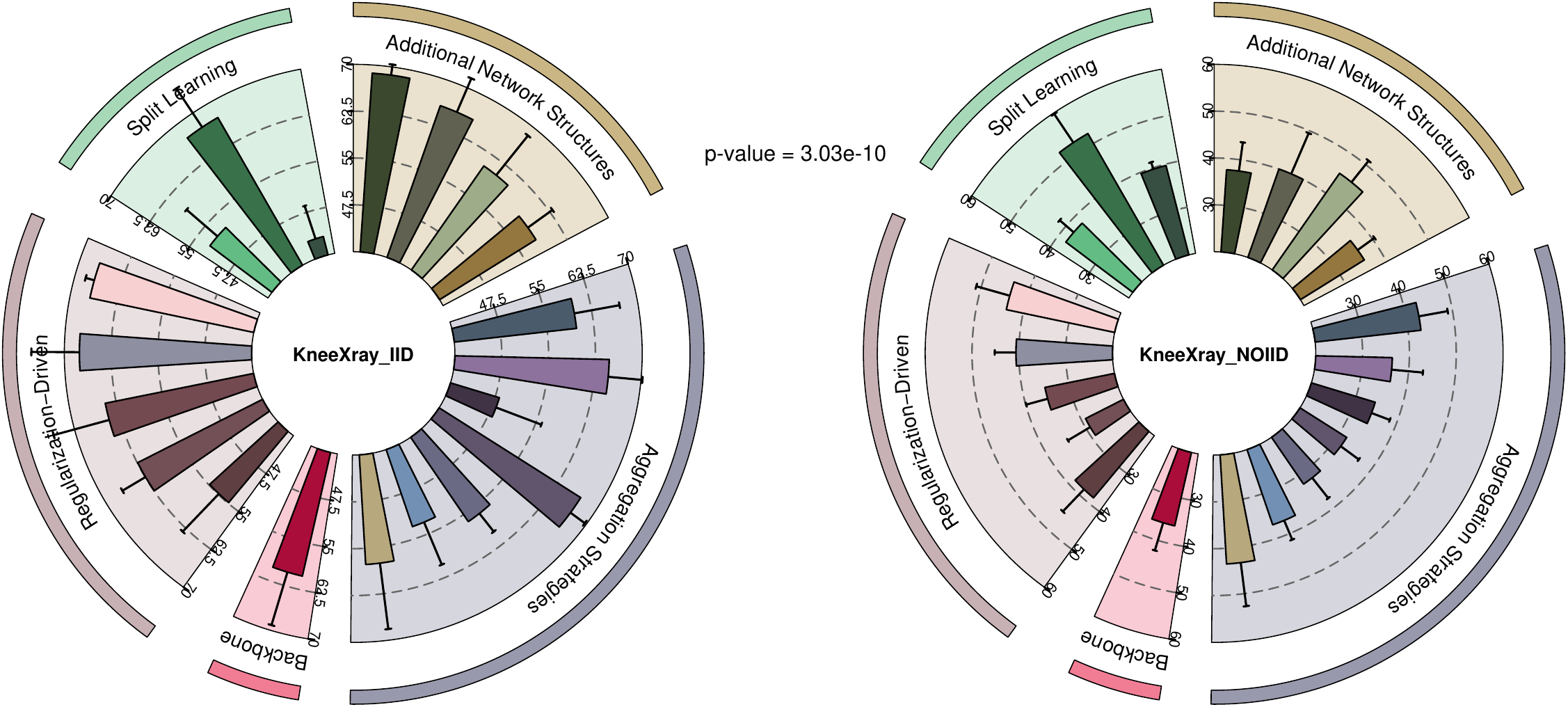}
    \caption{\textbf{Performance comparison of single-organ classification:} (a) Results under the independently and identically distributed (IID) setting; (b) Results under the non-IID setting. The middle panel of each organ comparison illustrates the statistical significance of performance differences between the two settings. All subfigures are arranged vertically.}
    \label{compare_result}
\end{figure*}

To evaluate the performance of FL algorithms in the MobenFL framework across different organ imaging, we select one dedicated medical imaging dataset for each of the 12 organs and conduct a systematic performance comparison and analysis. Fig.~\ref{compare_result} presents the classification performance evaluation of the MobenFL baseline method under both IID (Fig.~\ref{compare_result}.a) and Non-IID (Fig.~\ref{compare_result}.b) data settings.

\textbf{Brain}: For the task of Alzheimer's disease diagnosis in the brain, the overall performance of federated learning methods significantly decreased from an average of 69.04\% in the IID setting to 51.47\% in the Non-IID setting, a drop of 17.57 percentage points. The p-value for this difference is 4.96e-08, indicating that the performance decline is not a random phenomenon but rather a substantial impact of Non-IID data distribution on the performance of federated learning algorithms. Among different types of federated learning methods, the stability of performance varied significantly: Split Learning methods demonstrated the strongest adaptability, with an average performance drop of 11.71 percentage points, while Regularization-Driven methods were the most affected, with an average performance decline of 24.30 percentage points.
A detailed analysis of the performance of individual methods revealed that methods that performed excellently in the IID environment, such as SplitFed (85.45\%), PGFed (82.07\%), and FedNP (81.95\%), experienced significant performance degradation in the Non-IID environment, dropping to 56.56\%, 55.18\%, and 54.00\%, respectively, with declines exceeding 26 percentage points. This indicates the sensitivity of these methods to changes in data distribution. In contrast, TurboSVM maintained relatively stable performance in both IID and Non-IID environments, with 63.80\% and 57.99\%, respectively. FedBN decreased from 75.25\% in the IID setting to 57.07\% in the Non-IID setting, and although it experienced a decline, it still performed relatively well in the Non-IID environment. On the other hand, FedAu (from 72.38\% to 42.92\%), FedNP (from 81.95\% to 54.00\%), and FedDecorr (from 71.90\% to 44.56\%) exhibited extremely high sensitivity to changes in data distribution, with performance declines exceeding 27 percentage points.

\textbf{Eyes}: For the task of diabetic retinopathy diagnosis in eye images, the average performance of federated learning methods was 67.82\% in the IID setting and 69.83\% in the Non-IID setting. The p-value for this difference is 2.45e-01, indicating that the performance change did not reach statistical significance, suggesting that the Non-IID data distribution had limited impact on the performance of federated learning algorithms for this diagnostic task.
Among different types of federated learning methods, performance changes showed variation: Split Learning methods demonstrated stable performance in the Non-IID environment, with average performance remaining essentially unchanged; Regularization-Driven methods showed an average improvement of 4.71 percentage points in the Non-IID environment; while Aggregation Strategy and Additional Network Structure categories improved by 0.68 and 0.92 percentage points, respectively, with relatively small fluctuations.
A detailed analysis of individual method performance revealed that methods which performed well in the IID environment, such as FCCL (73.47\%), SplitAvg (73.34\%), and ProxyFL (71.35\%), showed varying degrees of change in the Non-IID environment. Among these, FCCL decreased to 71.25\%, ProxyFL dropped to 64.85\%, while SplitAvg maintained its performance at 73.34\%. Notably, FedGH significantly improved from 66.14\% to 79.29\% in the Non-IID environment, an increase of 13.15 percentage points, demonstrating excellent adaptability to non-independent and identically distributed data. Regularization-Driven methods such as TurboSVM (increasing from 66.63\% to 72.88\%), HarmoFL (rising from 68.10\% to 72.65\%), and MOON (improving from 67.66\% to 71.43\%) all achieved performance gains exceeding 3 percentage points in the Non-IID environment.
In contrast, FedDecorr (decreasing from 63.09\% to 52.97\%) and ProxyFL (dropping from 71.35\% to 64.85\%) demonstrated sensitivity to changes in data distribution, with relatively significant performance declines.

\textbf{Oral cavity}: For the oral squamous cell carcinoma diagnosis task, the overall performance of federated learning methods decreased from an average of 80.28\% in the IID setting to 75.63\% in the Non-IID setting, representing a decline of 4.65 percentage points. The p-value for this difference is 1.26e-04, indicating a statistically significant performance drop. Among different types of federated learning methods, performance stability showed notable variations: Split Learning methods were the most affected, with an average performance decrease of 8.16 percentage points; Aggregation Strategy and Additional Network Structure categories declined by 4.67 and 4.26 percentage points, respectively; while Regularization-Driven methods remained relatively stable, with only a 2.30 percentage point average performance drop.
A detailed analysis of individual method performance revealed that methods which performed excellently in the IID environment, such as HarmoFL (87.76\%), PGFed (87.35\%), and FedAu (86.94\%), all experienced varying degrees of performance degradation in the Non-IID environment, dropping to 75.51\%, 82.04\%, and 81.22\%, respectively. Among these, PGFed demonstrated relatively strong robustness, declining by only 5.31 percentage points. In contrast, FCCL (decreasing from 73.67\% to 63.06\%) and SplitAvg (dropping from 79.59\% to 64.08\%) showed the most significant performance declines, with reductions of 10.61 and 15.51 percentage points, respectively, indicating extremely high sensitivity to changes in data distribution. Notably, SplitFed improved from 75.92\% to 80.41\% in the Non-IID environment, making it one of the few methods that achieved performance improvement under non-independent and identically distributed conditions.

\textbf{Lung}: For the lung disease diagnosis task, the overall performance of federated learning methods significantly decreased from an average of 64.47\% in the IID setting to 54.21\% in the Non-IID setting, representing a decline of 10.26 percentage points. The p-value for this difference is 2.78e-07, indicating a statistically significant performance drop. Among different types of federated learning methods, performance stability showed notable variations: Split Learning methods demonstrated the strongest adaptability, with an average performance decrease of only 2.76 percentage points; Additional Network Structure methods declined by 11.13 percentage points; while Regularization-Driven and Aggregation Strategy methods were the most significantly affected, with average performance drops of 11.33 and 13.06 percentage points, respectively.
A detailed analysis of individual method performance revealed that methods which performed excellently in the IID environment, such as HarmoFL (73.18\%), SplitFed (71.09\%), and FedAvg (70.12\%), all experienced significant performance degradation in the Non-IID environment, dropping to 55.93\%, 62.29\%, and 55.74\%, respectively. Among these, SplitFed demonstrated relatively strong robustness, declining by only 8.80 percentage points. In contrast, FedAu (decreasing from 66.32\% to 48.58\%), FLOP (dropping from 66.16\% to 47.91\%), and FedCD (falling from 64.73\% to 49.66\%) showed the most significant performance declines, with reductions all exceeding 15 percentage points, indicating extremely high sensitivity to changes in data distribution.

\textbf{Breast}: For the breast cancer subtype diagnosis task, the overall performance of federated learning methods decreased from an average of 44.36\% in the IID setting to 42.16\% in the Non-IID setting, representing a decline of 2.2 percentage points. The p-value for this difference is 6.56e-03, indicating a statistically significant performance drop. Among different types of federated learning methods, Split Learning methods were the most affected, with an average performance decrease of 6.67 percentage points; Aggregation Strategy and Additional Network Structure categories declined by 2.58 and 2.74 percentage points, respectively; while Regularization-Driven methods remained relatively stable, with only a 1.87 percentage point average performance drop.
A detailed analysis of individual method performance revealed that methods which performed well in the IID environment, such as SplitFed (49.82\%), FedALA (48.25\%), and HarmoFL (47.72\%), showed varying degrees of change in the Non-IID environment. Among these, FedALA and HarmoFL demonstrated relatively strong stability, maintaining performances of 45.96\% and 46.49\%, respectively; while SplitFed showed a significant decline, dropping from 49.82\% to 36.67\%, a decrease of 13.15 percentage points. Notably, FLOP improved from 42.98\% to 48.77\% in the Non-IID environment, making it the only method that achieved significant performance improvement. In contrast, FedAu (decreasing from 44.39\% to 38.25\%), FedNP (dropping from 42.63\% to 39.65\%), and SplitAvg (falling from 44.04\% to 37.89\%) demonstrated sensitivity to changes in data distribution, with relatively noticeable performance declines.

\textbf{Abdomen}: For the abdominal organ classification task, the overall performance slightly decreased from an average of 80.59\% in the IID setting to 79.58\% in the Non-IID setting, representing a decline of 1.01 percentage points. The p-value for this difference is 1.14e-01, which does not reach statistical significance (p > 0.05), indicating that the Non-IID data distribution had limited impact on the performance of federated learning algorithms for this classification task.
Among different types of federated learning methods, Split Learning methods demonstrated the strongest stability, with an average performance decrease of only 0.29 percentage points; Additional Network Structure methods showed an average improvement of 0.48 percentage points in the Non-IID environment; while Regularization-Driven and Aggregation Strategy categories declined by 0.86 and 1.53 percentage points, respectively, with relatively small fluctuations.
A detailed analysis of individual method performance revealed that methods which performed excellently in the IID environment, such as SplitFed (83.83\%), GPFL (83.14\%), and FedNP (82.52\%), showed varying degrees of change in the Non-IID environment. Among these, GPFL and FedNP demonstrated good stability, maintaining performances of 81.94\% and 81.68\%, respectively; while SplitFed experienced a certain decline, dropping from 83.83\% to 81.57\%. Notably, FedALA improved from 80.70\% to 81.23\% in the Non-IID environment, and MOON slightly decreased from 81.69\% to 81.47\%, both maintaining relatively high performance levels.
In contrast, FedAu significantly declined from 81.68\% to 67.61\% in the Non-IID environment, a decrease of 14.07 percentage points, demonstrating particular sensitivity to changes in data distribution.
 
\textbf{Colorectal}: For the colitis severity grading task, the overall performance significantly decreased from an average of 60.37\% in the IID setting to 50.89\% in the Non-IID setting, representing a decline of 9.48 percentage points. The p-value for this difference is 2.01e-08, indicating a statistically significant performance drop.
Among different types of federated learning methods, additional network structure methods demonstrated the strongest adaptability, with an average performance decrease of 8.35 percentage points; Regularization-Driven and Aggregation Strategy methods were the most significantly affected, with average performance declines of 12.42 and 9.70 percentage points, respectively; while Split Learning methods showed an average decrease of 10.14 percentage points.
A detailed analysis of individual method performance revealed that methods which performed excellently in the IID environment, such as FedNP (65.22\%), SplitFed (64.69\%), and PGFed (63.44\%), all experienced significant performance degradation in the Non-IID environment, dropping to 53.89\%, 45.67\%, and 50.45\%, respectively. Among these, FedNP demonstrated relatively strong robustness, declining by 11.33 percentage points.
In contrast, MOON (decreasing from 63.09\% to 46.77\%), SplitFed (dropping from 64.69\% to 45.67\%), and FedDecorr (falling from 62.55\% to 46.88\%) showed the most significant performance declines, with reductions all exceeding 15 percentage points, indicating extremely high sensitivity to changes in data distribution.
Notably, FedAu was the only method that achieved performance improvement in the Non-IID environment, increasing from 62.43\% to 60.77\%, demonstrating unique adaptive characteristics.

\textbf{Kidney}: For the renal cortex cell classification task, the overall performance decreased from an average of 64.71\% in the IID setting to 61.43\% in the Non-IID setting, representing a decline of 3.28 percentage points. The p-value for this difference is 9.46e-02, which does not reach statistical significance (p > 0.05), indicating that the Non-IID data distribution had limited impact on the performance of federated learning algorithms for this classification task.
Among different types of federated learning methods, Regularization-Driven methods demonstrated the strongest stability, with an average performance decrease of only 0.63 percentage points; Additional Network Structure methods declined by an average of 3.56 percentage points; while Aggregation Strategy and Split Learning categories were more noticeably affected, with average performance drops of 8.14 and 7.76 percentage points, respectively.
A detailed analysis of individual method performance revealed that methods which performed excellently in the IID environment, such as FCCL (70.51\%), FedNP (70.33\%), and SplitFed (69.53\%), showed varying degrees of change in the Non-IID environment. Among these, FedNP and SplitFed demonstrated good stability, maintaining performances of 70.30\% and 69.46\%, respectively; while FCCL experienced a significant decline, dropping from 70.51\% to 51.87\%.
Notably, FedProx slightly improved from 69.25\% to 69.32\% in the Non-IID environment, and MOON increased from 67.26\% to 67.56\%, both maintaining relatively high performance levels.
In contrast, FedALA (decreasing from 60.92\% to 50.98\%), ProxyFL (dropping from 67.09\% to 51.95\%), and FLOP (falling from 66.37\% to 50.12\%) demonstrated sensitivity to changes in data distribution, with relatively noticeable performance declines.

\textbf{Colon}: For the colonic pathological cell classification task, the overall performance significantly decreased from an average of 93.86\% in the IID setting to 85.78\% in the Non-IID setting, representing a decline of 8.08 percentage points. The p-value for this difference is 7.42e-05, indicating a statistically significant performance drop.
Among different types of federated learning methods, Additional Network Structure methods demonstrated the strongest adaptability, with an average performance decrease of 6.65 percentage points; Split Learning methods declined by an average of 8.20 percentage points; while Aggregation Strategy and Regularization-Driven categories were the most significantly affected, with average performance drops of 9.56 and 9.89 percentage points, respectively.
A detailed analysis of individual method performance revealed that methods which performed excellently in the IID environment, such as FedNP (98.99\%), PraVFed (98.82\%), and PGFed (97.03\%), all experienced significant performance degradation in the Non-IID environment, dropping to 91.16\%, 90.00\%, and 79.29\%, respectively. Among these, FedNP demonstrated relatively strong robustness, declining by 7.83 percentage points.
In contrast, FedGH (decreasing from 94.93\% to 68.04\%), PGFed (dropping from 97.03\% to 79.29\%), and FCCL (falling from 92.13\% to 68.16\%) showed the most significant performance declines, with reductions all exceeding 23 percentage points, indicating extremely high sensitivity to changes in data distribution.
Notably, FedAu was one of the few methods that achieved performance improvement under non-independent and identically distributed conditions, increasing from 94.29\% to 98.81\%.

\textbf{Blood}: For the blood cell classification task, the overall performance of federated learning methods significantly decreased from an average of 81.54\% in the IID setting to 73.69\% in the Non-IID setting, representing a drop of 7.85 percentage points. The p-value for this difference is 1.29e-04, indicating that the performance decline is statistically significant. Among different types of federated learning methods, the stability of performance varied noticeably: Split Learning methods demonstrated the strongest adaptability, with an average performance drop of 7.03 percentage points; Regularization-Driven methods were the most affected, with an average performance decline of 9.30 percentage points; while Aggregation Strategy and Additional Network Structure categories declined by 8.66 and 6.19 percentage points, respectively.
A detailed analysis of individual method performance revealed that methods which performed excellently in the IID environment, such as FedProx (98.43\%), FedCD (96.58\%), and FedBN (92.19\%), all experienced significant performance degradation in the Non-IID environment, dropping to 79.64\%, 76.66\%, and 70.46\%, respectively. These declines amounted to 18.79, 19.92, and 21.73 percentage points, demonstrating these methods sensitivity to changes in data distribution. In contrast, FedAvg maintained relatively stable performance in both IID and Non-IID environments, with 82.37\% and 79.76\%, respectively; FedNP decreased from 84.39\% in the IID setting to 81.09\% in the Non-IID setting, with only a 3.30 percentage point drop, demonstrating strong robustness. Meanwhile, FedBN (declining 21.73 percentage points), FedCD (declining 19.92 percentage points), and FedProx (declining 18.79 percentage points) exhibited extremely high sensitivity to data distribution changes. Although these methods performed excellently in the IID environment, they showed the most significant performance degradation in the Non-IID setting.

\textbf{Skin}: For the dermatological disease classification task, the overall performance significantly decreased from an average of 75.74\% in the IID setting to 69.64\% in the Non-IID setting, representing a decline of 6.10 percentage points. The p-value for this difference is 3.55e-10, indicating a statistically significant performance drop.
Among different types of federated learning methods, Split Learning methods demonstrated the strongest adaptability, with an average performance decrease of 4.20 percentage points; Additional Network Structure methods declined by an average of 6.15 percentage points; while Regularization-Driven and Aggregation Strategy categories were the most significantly affected, with average performance drops of 6.87 and 7.02 percentage points, respectively.
A detailed analysis of individual method performance revealed that methods which performed excellently in the IID environment, such as GPFL (78.50\%), FedDecorr (78.85\%), and SplitFed (78.50\%), all experienced noticeable performance degradation in the Non-IID environment, dropping to 70.05\%, 72.15\%, and 74.70\%, respectively. Among these, SplitFed demonstrated relatively strong robustness, declining by only 3.80 percentage points.
In contrast, FedCD (decreasing from 76.10\% to 66.60\%), FCCL (dropping from 67.80\% to 62.85\%), and PGFed (falling from 77.60\% to 68.25\%) showed more significant performance declines, with reductions all exceeding 9 percentage points, indicating sensitivity to changes in data distribution.
Notably, FedALA decreased from 76.10\% to 72.55\% in the Non-IID environment, with a relatively small decline, demonstrating a certain degree of stability.

\textbf{Knee}: For the knee fracture classification task, the overall performance significantly decreased from an average of 59.43\% in the IID setting to 38.63\% in the Non-IID setting, representing a substantial decline of 20.80 percentage points. The p-value for this difference is 3.03e-10, indicating a statistically significant performance drop.
Among different types of federated learning methods, Split Learning methods demonstrated the strongest adaptability, with an average performance decrease of 9.56 percentage points; Additional Network Structure methods declined by an average of 21.00 percentage points; while Regularization-Driven and Aggregation Strategy categories were the most significantly affected, with average performance drops of 24.94 and 22.87 percentage points, respectively.
A detailed analysis of individual method performance revealed that methods which performed excellently in the IID environment, such as FedNP (68.68\%), PGFed (67.58\%), and SplitFed (67.27\%), all experienced severe performance degradation in the Non-IID environment, dropping to 37.58\%, 40.61\%, and 52.42\%, respectively. Among these, SplitFed demonstrated the strongest robustness, declining by only 14.85 percentage points.
In contrast, MOON (decreasing from 62.44\% to 29.09\%), FedNP (dropping from 68.68\% to 37.58\%), and HarmoFL (falling from 64.24\% to 35.15\%) showed the most significant performance declines, with reductions all exceeding 29 percentage points, indicating extremely high sensitivity to changes in data distribution.

\subsection{Quantitative Evaluation under Clinical Scenarios}\label{exp-2}

Clinical scenarios are highly diverse, with significant variations in disease presentation, imaging equipment, and data sources. Therefore, it is crucial to systematically evaluate the generalization capability and robustness of federated learning algorithms across different clinical situations. To this end, this study designs the following three quantitative evaluation experiments: \textit{Cross-organ Performance Evaluation}, which examines the algorithm's ability to distinguish lesions in different organs within the same imaging modality; \textit{Cross-disease Performance Evaluation}, which assesses the algorithm's performance in differentiating among various diseases affecting the same organ; and \textit{Cross-device Performance Evaluation}, which verifies the stability and adaptability of the algorithm when dealing with the same disease but from different devices or data sources.

\subsubsection{Cross-organ Performance Evaluation}
\begin{figure*}
	\centering
	\includegraphics[width=\linewidth,scale=1.0]{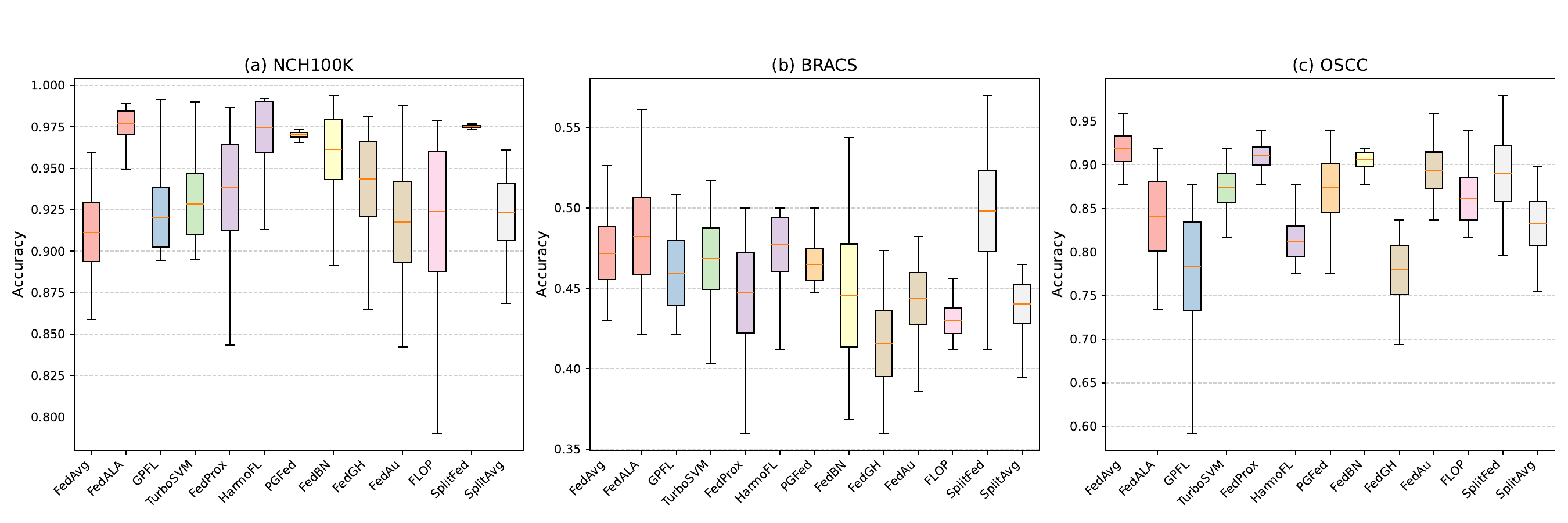}
	\caption{\textbf{Detection results from different pathological Datasets}. Each box plot summarizes the performance distribution on the test sets of 13 federated learning algorithms. The yellow line in the middle of the box represents the mean performance of the algorithm, the length of the box represents the standard deviation of the performance, and the upper and lower extensions of the whiskers represent the best and worst performances, respectively. The three datasets are: \textbf{NCH100K}: Hematoxylin \& eosin (H\&E) stained histological images of human colorectal cancer (CRC) and normal tissue.; \textbf{BRACS}: Breast carcinoma subtyping; \textbf{OSCC}: Oral squamous cell carcinoma.}
	\label{SMD_Path}
\end{figure*}

\begin{figure*}
	\centering
	\includegraphics[width=\linewidth,scale=1.0]{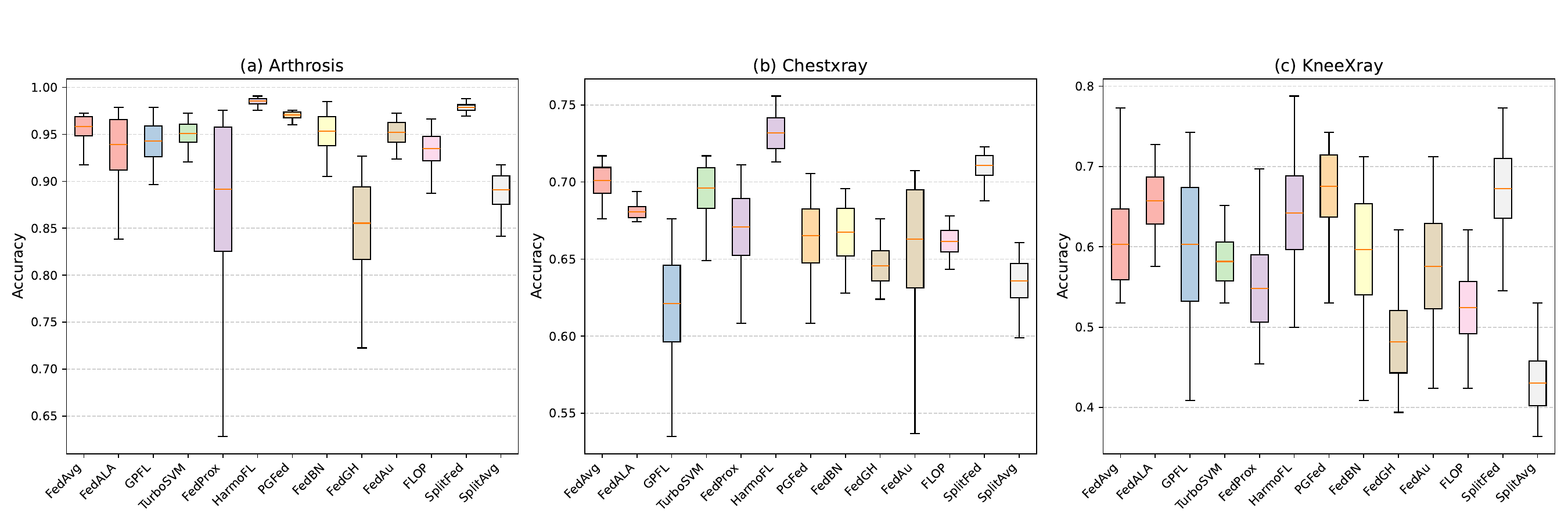}
	\caption{\textbf{Detection results from different X-ray datasets}. Each box plot summarizes the performance distribution on the test sets of 13 federated learning algorithms. The yellow line in the middle of the box represents the mean performance of the algorithm, the length of the box represents the standard deviation of the performance, and the upper and lower extensions of the whiskers represent the best and worst performances, respectively. The three datasets are \textbf{Arthrosis}: X-ray Hand Small Joint Classification; \textbf{ChestXray}: Chest X-ray examination; and \textbf{kneeXray}: Digital Knee X-ray Images.}
	\label{SMD_Xray}
\end{figure*}

To investigate the generalizable diagnostic capability of federated learning algorithms across multiple diseases with divergent pathological mechanisms and anatomical locations under a unified imaging modality, we conducted experimental evaluations on pathological and X-ray images, with the results presented in Fig.~\ref{SMD_Path} and Fig.~\ref{SMD_Xray}, respectively.

On the OSCC dataset in Fig.~\ref{SMD_Path}, FedAvg, FedProx, and FedBN stand out prominently, with accuracy rates all exceeding 0.9 and relatively small standard deviations. This indicates that these methods exhibit excellent robustness and consistency in scenarios where the data distribution is relatively uniform and the classes are fairly balanced. FedALA and TurboSVM also perform decently. However, GPFL, FedGH, and HarmoFL have large standard deviations, suggesting poor stability and possible sensitivity to the local data distributions of certain clients. On the BRACS dataset, the overall accuracy rates of all methods decline significantly (concentrated in the range of 0.4 - 0.5), and most methods have large standard deviations. This demonstrates that the task is more challenging, likely due to significant class imbalance or distribution differences across clients. SplitFed performs the best, with an accuracy rate close to 0.5, followed by FedALA and HarmoFL, while FedGH, FedAu, and FLOP perform poorly. This indicates that in complex multi-class tasks, partial aggregation or split learning mechanisms (such as SplitFed) may be better equipped to handle heterogeneous data. On the NCH100K dataset, almost all methods achieve accuracy rates above 0.90, with AUC values approaching 0.995. This suggests that the classification task on this dataset is relatively simple, or the data quality is high with significant inter-class distinguishability. FedALA, HarmoFL, PGFed, SplitFed, and FedBN perform the best, with accuracy rates exceeding 0.97 and extremely small standard deviations, demonstrating excellent consistency and stability. FedAvg, as a baseline method, also shows decent performance but still lags behind the aforementioned methods, indicating that introducing adaptive mechanisms or personalized strategies can lead to significant improvements on large datasets.

Overall, FedAvg serves as a robust baseline, performing reliably in most cases. FedProx and FedBN excel at handling heterogeneous data. SplitFed has a distinct advantage in complex tasks (such as on the BRACS dataset). Meanwhile, FedALA, HarmoFL, and PGFed perform best in large-scale or highly consistent datasets.


On the Arthrosis dataset in Fig.~\ref{SMD_Xray}, the overall accuracy (ACC) and AUC values of all methods are relatively high, indicating that the task is relatively simple or the data quality is good. HarmoFL performs the best, achieving an ACC of 0.9854 and an AUC of 0.9991. Moreover, its performance across different clients is fairly balanced, demonstrating its excellent cross-client consistency. In contrast, FedGH and SplitAvg exhibit relatively poor performance, with ACC values of 0.8555 and 0.8909, respectively. Particularly, FedGH has an accuracy as low as 0.7225 on one client, which may be attributed to local data distribution bias or deficiencies in the model aggregation strategy. On the ChestXray dataset, the overall performance is moderate. HarmoFL once again takes the lead, with an ACC of 0.7318 and an AUC of 0.8457, and its performance across clients remains relatively stable. FedAvg and SplitFed follow closely, with ACC values of 0.7012 and 0.7109, respectively, showing robust performance. In contrast, GPFL and SplitAvg perform poorly, with ACC values of only 0.6213 and 0.636, respectively. The lowest accuracy on a client is as low as 0.5348 for GPFL, suggesting that such methods may lack effective feature alignment or local optimization mechanisms when dealing with complex chest X-ray images. On the KneeXray dataset, the performance of all methods is generally low and highly variable, reflecting the greater challenge of this task. PGFed and SplitFed perform relatively well, with ACC values of 0.6758 and 0.6727, respectively, and AUC values exceeding 0.9. However, there are significant differences in performance across clients (e.g., PGFed shows client performance ranging from 0.5303 to 0.7424). FedGH and SplitAvg perform the worst, with ACC values of only 0.4818 and 0.4303, respectively, and the lowest accuracy on a client dropping as low as 0.3939. This indicates that these methods have insufficient generalization ability when faced with highly variable knee X-ray data.

Overall, HarmoFL demonstrates stable and outstanding performance across multiple datasets, particularly excelling in high-precision tasks. SplitFed also shows decent performance on complex data, albeit with greater variability. In contrast, methods such as FedGH and SplitAvg perform poorly across all datasets, likely due to their aggregation strategies or local training approaches being unable to effectively handle data heterogeneity.

\subsubsection{Cross-disease Performance Evaluation}
\begin{figure*}
	\centering
	\includegraphics[width=\linewidth,scale=1.0]{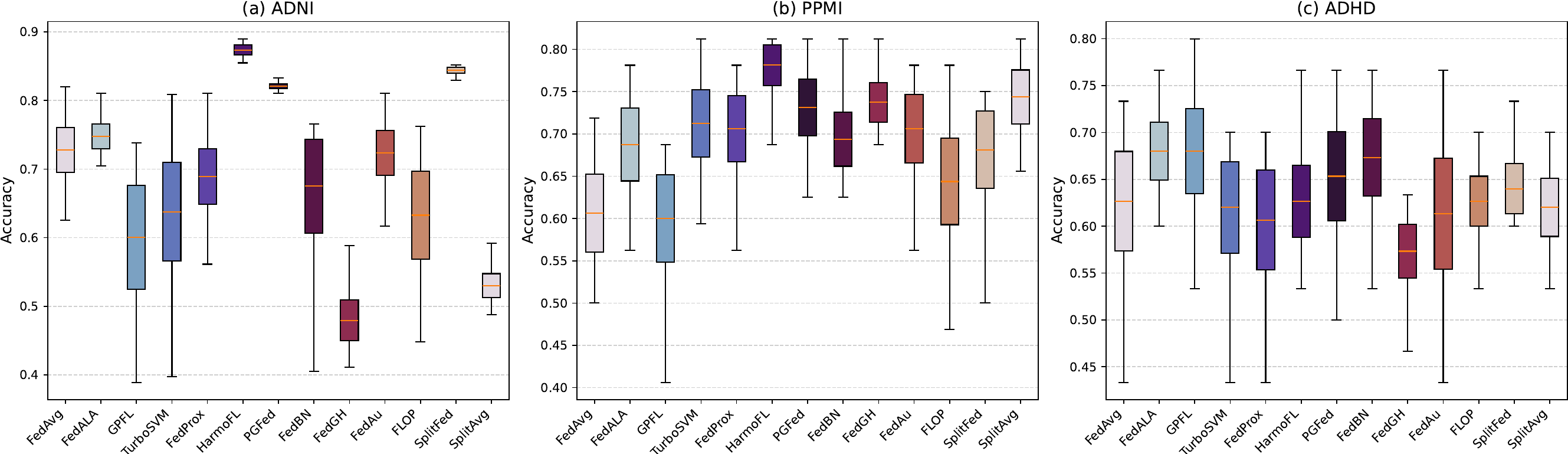}
	\caption{\textbf{Detection results for different brain diseases}. Each box plot summarizes the performance distribution on the test sets of 13 federated learning algorithms. The yellow line in the middle of the box represents the mean performance of the algorithm, the length of the box represents the standard deviation of the performance, and the upper and lower extensions of the whiskers represent the best and worst performances, respectively. The three datasets are: \textbf{ADNI}: Alzheimer's Disease Neuroimaging Initiative; \textbf{ADHD}: Attention Deficit Hyperactivity Disorder; \textbf{PPMI}: Parkinson’s Progression Markers Initiative.}
	\label{STD_Brain}
\end{figure*}

\begin{figure*} 
	\centering
	\includegraphics[width=\linewidth,scale=1.0]{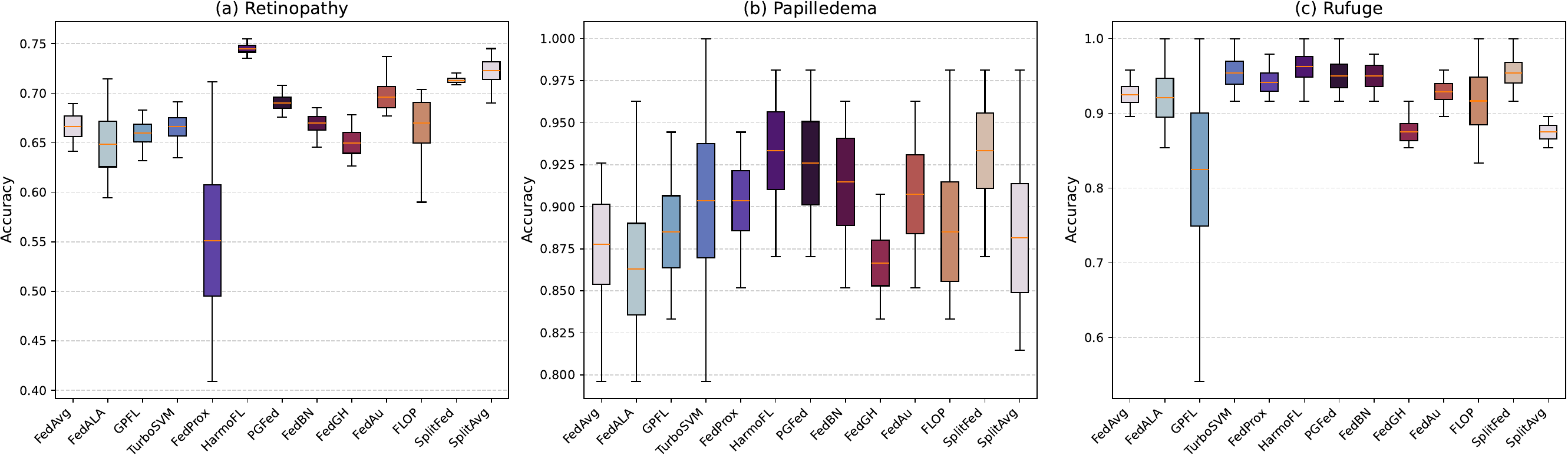}
	\caption{\textbf{Detection results for different eye diseases}. Each box plot summarizes the performance distribution on the test sets of 13 federated learning algorithms. The yellow line in the middle of the box represents the mean performance of the algorithm, the length of the box represents the standard deviation of the performance, and the upper and lower extensions of the whiskers represent the best and worst performances, respectively. The three datasets are \textbf{Retinopathy}; \textbf{Papilledema}; and \textbf{Refuge}: Retinal Fundus Glaucoma Challenge Edition.}
	\label{STD_Eye}
\end{figure*}

To investigate the adaptability of federated learning methods in diagnosing different diseases within the same organ, we select the brain and the eye as representative organs and conducted disease-specific evaluation experiments for multiple conditions, with the results presented in Fig.~\ref{STD_Brain} and Fig.~\ref{STD_Eye}, respectively.

Overall, HarmoFL and PGFed have demonstrated outstanding performance and stability on the ADNI and PPMI datasets in Fig.~\ref{STD_Brain}. Particularly on the ADNI dataset, HarmoFL achieved an accuracy of 0.873 and an AUC of 0.959, with extremely small standard deviations, indicating its consistently excellent performance across different clients. PGFed and SplitFed also performed prominently on ADNI, with accuracies exceeding 0.82, highlighting the advantages of these methods in handling neuroimaging data. However, in the ADHD diagnosis task, all FL (Federated Learning) algorithms encountered significant challenges, with the best accuracy reaching only 0.68 (FedALA) and the highest AUC being 0.729, reflecting the inherent complexity of this disease diagnosis or the specific difficulties posed by the data characteristics. FedGH exhibited the poorest performance across all three datasets, with an accuracy as low as 0.495 and an AUC of only 0.700 on ADNI. Moreover, its performance fluctuated dramatically across clients (e.g., ranging from 0.389 to 0.588 on ADNI), suggesting that its aggregation strategy or model architecture struggles to accommodate the distribution disparities in medical data.

From the perspective of stability, different FL algorithms exhibited markedly distinct characteristics across various disease datasets. On ADNI, HarmoFL, PGFed, and SplitFed all demonstrated small standard deviations, indicating stable performance. However, on the PPMI and ADHD datasets, almost all algorithms exhibited relatively large standard deviations. Notably, FedAu on ADHD had an accuracy standard deviation of 11.85\%, indicating significant performance fluctuations, which may stem from the high degree of data heterogeneity or the method's sensitivity to specific data distributions. Client-level analysis further reveals the core challenges in federated learning. Each model exhibited significant performance variations across the five clients. Taking FedAvg as an example, on the ADNI dataset, client accuracies ranged from 0.6251 to 0.8197, with such performance fluctuations being prevalent across almost all methods. Nevertheless, HarmoFL demonstrated relatively uniform client performance across the three datasets (e.g., all client accuracies on ADNI exceeded 0.85), showcasing its strong generalization ability and client adaptability. In contrast, methods such as GPFL and TurboSVM experienced sharp performance declines on certain clients (e.g., GPFL had an accuracy of only 0.3891 on one client in ADNI), highlighting the vulnerability of these methods to changes in data distribution.

Overall, model performance on ophthalmic datasets generally surpasses that on brain disease datasets in Fig.~\ref{STD_Eye}. Particularly in the Refuge and Papilledema tasks, multiple models achieved accuracies exceeding 0.90, with some even reaching above 0.96, and AUC values predominantly higher than 0.95. This demonstrates the immense potential of federated learning in ophthalmic image diagnosis. However, in the Retinopathy task, model performance was notably lower, with the best accuracy reaching only 0.744 (HarmoFL), reflecting significant variations in the diagnostic difficulty across different ophthalmic diseases. HarmoFL once again showcased its exceptional performance, maintaining a leading or near-leading position across all three ophthalmic tasks. Especially in the Retinopathy task, it significantly outperformed other methods with an accuracy of 0.744 and an AUC of 0.783, coupled with extremely small standard deviations, indicating outstanding stability. In the Refuge task, HarmoFL achieved an accuracy of 0.962 and an AUC of 0.986, approaching near-perfect performance. SplitFed and TurboSVM also performed excellently across multiple tasks, achieving accuracies of 0.954 and 0.954, respectively, in the Refuge task, demonstrating the adaptability of Split Learning methods and TurboSVM to ophthalmic data.

Notably, in the Retinopathy task, FedProx exhibited abnormal performance, with an accuracy as low as 0.551 and a substantial standard deviation. Client performance ranged from 0.409 to 0.711, indicating that this method is highly sensitive to such data distributions and may even fail on certain clients. GPFL also encountered similar issues in the Refuge task, achieving an accuracy of only 0.825 but with a high standard deviation of 0.151. While one client had a performance as low as 0.541, others reached 1.0. Such extreme fluctuations highlight the consistency challenges that federated learning methods may face when processing certain ophthalmic data.

From the perspective of client performance analysis, in the Refuge and Papilledema tasks, client performance across various algorithms was generally high and relatively evenly distributed. For instance, HarmoFL maintained an accuracy above 0.916 across all clients in the Refuge task, and TurboSVM even achieved perfect performance (1.0) on one client. This suggests that these ophthalmic diagnostic tasks may possess relatively consistent data feature distributions, or the models can better learn shared features across clients. However, in the Retinopathy task, client performance exhibited significant fluctuations, with FedProx showing a performance difference exceeding 0.3 across different clients, reflecting the substantial heterogeneity that may exist in such data.

Overall, in the tasks of diagnosing brain and ophthalmic diseases, HarmoFL demonstrated the most outstanding and stable performance. It achieved the highest or near-highest accuracy and AUC in the vast majority of tasks, such as ADNI, PPMI, as well as the ophthalmic Refuge and Papilledema tasks, significantly outperforming other methods. PGFed, SplitFed, and TurboSVM also exhibited excellent performance, particularly showcasing high precision and strong robustness in ophthalmic tasks. In contrast, FedGH and GPFL performed the worst across all tasks, being extremely sensitive to data heterogeneity and experiencing severe performance fluctuations. FedProx displayed an unstable state, with catastrophic failures occurring in some tasks. Traditional baseline methods, FedAvg and FedBN, generally ranked at a medium level in most cases.
From the perspective of task difficulty, the diagnosis of brain diseases, especially ADHD, presented a greater overall challenge compared to ophthalmic tasks. The latter achieved extremely high accuracies of over 0.9 in diagnosing glaucoma and papilledema. Meanwhile, the retinopathy task exhibited a difficulty level similar to that of brain-related tasks.

\subsubsection{Cross-device Performance Evaluation}

\begin{figure*}
	\centering
	\includegraphics[width=\linewidth,scale=1.0]{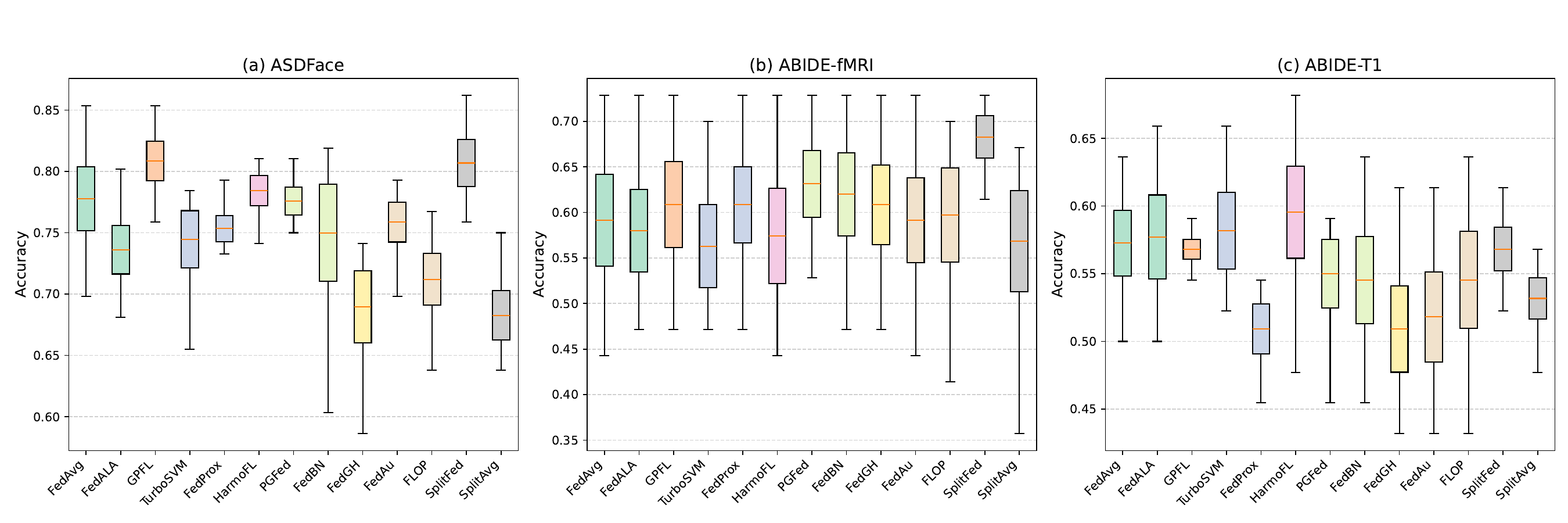}
	\caption{\textbf{Detection Results from Different ASD Datasets}. Each box plot summarizes the performance distribution on the test sets of 13 federated learning algorithms. The yellow line in the middle of the box represents the mean performance of the algorithm, the length of the box represents the standard deviation of the performance, and the upper and lower extensions of the whiskers represent the best and worst performances, respectively. The three datasets are: \textbf{ASDFace}: Autism facial recognition data; \textbf{ABIDE-fMRI}: Functional magnetic resonance imaging data in Autism Brain Imaging Data Exchange(ABIDE); \textbf{ABIDE-T1}: T1 structural imaging data in Autism Brain Imaging Data Exchange(ABIDE).}
	\label{SDD_ASD}
\end{figure*}

\begin{figure*} 
	\centering
	\includegraphics[width=\linewidth,scale=1.0]{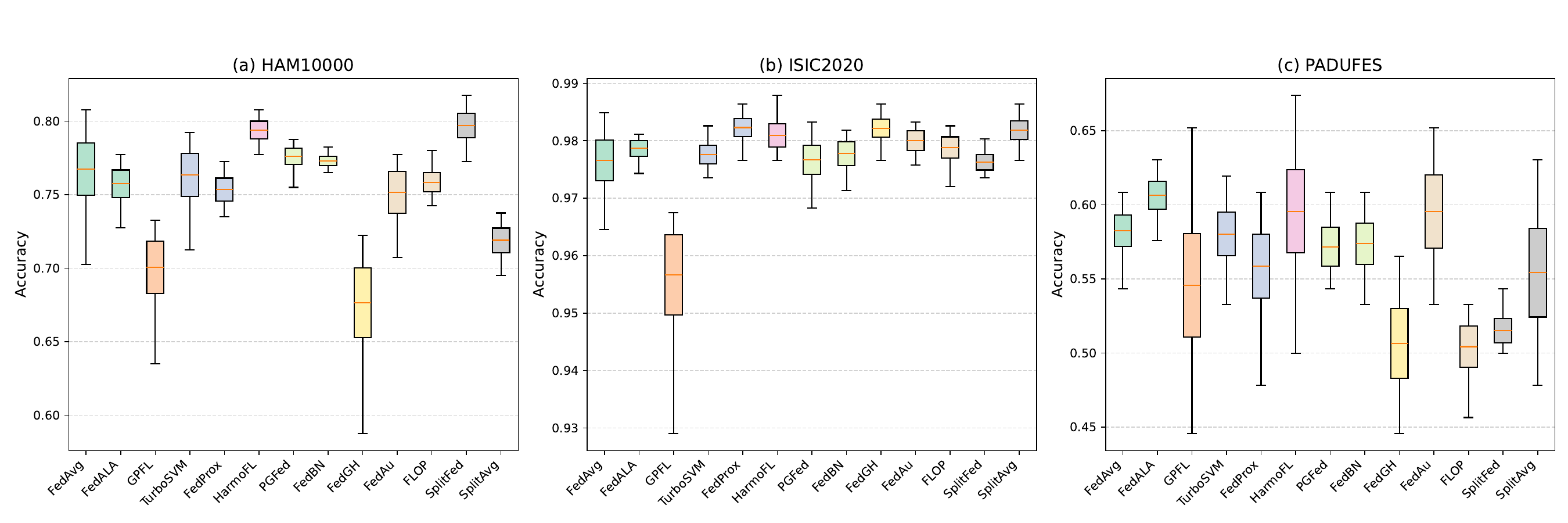}
	\caption{\textbf{Detection results for different skin diseases}. Each box plot summarizes the performance distribution on the test sets of 13 federated learning algorithms. The yellow line in the middle of the box represents the mean performance of the algorithm, the length of the box represents the standard deviation of the performance, and the upper and lower extensions of the whiskers represent the best and worst performances, respectively. The three datasets are \textbf{HAM10000}: Human Against Machine with 10000 training images; \textbf{ISIC2020}: International Skin Imaging Collaboration (ISIC) 2020 Challenge Dataset; \textbf{PADUFES}: A skin lesion dataset composed of patient data and clinical images collected from smartphones.}
	\label{SDD_Skin}
\end{figure*}

Due to the fact that different imaging devices in clinical practice tend to generate heterogeneous data, evaluating the performance of federated learning algorithms in such scenarios is crucial. We conducted cross-device performance evaluations using autism and dermatological disease data to assess the generalization capability and robustness of the algorithms in the presence of differences in data modalities and sources, with the results shown in Fig.~\ref{SDD_ASD} and Fig.~\ref{SDD_Skin}, respectively.

Through an analysis of the performance results of various FL algorithms on three autism spectrum disorder (ASD) datasets (ASDFace, ABIDE-fMRI, and ABIDE-T1) in Fig.~\ref{SDD_ASD}, it is evident that different data modalities have a significant impact on classification performance. Overall, the classification performance on facial data (ASDFace) is the best, with average accuracy and Area Under the Curve (AUC) values generally higher than those of the two brain imaging datasets. This aligns with intuitive understanding, as facial features exhibit high discriminative power in autism diagnosis, particularly in capturing differences through visual perception and behavioral characteristics. For instance, on the ASDFace dataset, GPFL and SplitFed achieved accuracies of 0.8086 and 0.8069, respectively, with corresponding AUC values of 0.8735 and 0.8859, demonstrating strong classification capabilities. In contrast, FedGH and SplitAvg performed relatively poorly on this dataset, with accuracies of only 0.6897 and 0.6828, respectively, likely due to their insufficient adaptability to heterogeneous client data.

Among brain imaging data, ABIDE-fMRI (functional magnetic resonance imaging) demonstrated superior overall performance compared to ABIDE-T1 (structural magnetic resonance imaging). For example, on the ABIDE-fMRI dataset, SplitFed achieved an accuracy of 0.6829 and an AUC of 0.6895, significantly outperforming most other methods. FedAvg, FedProx, and PGFed also exhibited stable performance, with accuracies around 0.6. This result supports the effectiveness of functional connectivity features in autism diagnosis, as fMRI can capture functional abnormalities in brain networks that are closely related to the neural mechanisms of autism. In contrast, T1 structural images primarily reflect morphological features of brain regions, with relatively subtle changes in autism, resulting in generally lower classification performance. On the ABIDE-T1 dataset, the best performance was achieved by HarmoFL, with an accuracy of 0.5955 and an AUC of 0.6454, which is still significantly lower than the results obtained from fMRI and facial data.

From an algorithmic perspective, SplitFed demonstrated outstanding performance across multiple datasets, indicating that its split learning architecture offers certain advantages in handling cross-center heterogeneous data. Methods such as FedBN and FedProx also exhibited stable performance in some scenarios, particularly when there were significant differences in client data distributions (e.g., notable fluctuations in values in the Client column). These methods mitigated challenges posed by non-IID data through local normalization or regularization techniques. In contrast, FedGH, FLOP, and SplitAvg performed relatively weakly across multiple datasets, likely due to their insufficient adaptability to data distribution shifts or client differences.

In summary, data modality has a decisive impact on autism classification performance: facial data is the easiest to distinguish, followed by fMRI, which offers greater neuroscientific interpretability, while T1 structural images present the greatest classification difficulty due to their subtle morphological changes.


We can observe the capabilities of different algorithms in handling data heterogeneity. On the ISIC2020 task in Fig~.\ref{SDD_Skin}, which is considered "simple," traditional algorithms such as FedAvg and FedProx have demonstrated exceptional performance, even outperforming many newly proposed complex algorithms. FedProx (with an accuracy of 0.9823) effectively mitigates the client drift problem by introducing a proximal term to constrain local updates, and its AUC (0.8493) is significantly higher than other methods, proving the reliability of its output confidence. FedGH (with an accuracy of 0.9822) achieves top-tier performance by generating high-quality features to aid training. This indicates that in scenarios with relatively balanced data distributions, enhancing training stability and feature quality are crucial.

However, the true strengths and weaknesses of algorithms become apparent on the more challenging HAM10000 and PADUFES datasets. On HAM10000, HarmoFL (with an accuracy of 0.7940) and SplitFed (with an accuracy of 0.7970) stand out. HarmoFL focuses on addressing style discrepancies across clients, and its AUC value (0.9462) demonstrates that it has learned highly generalized features, effectively reducing domain shifts caused by different devices. SplitFed, as a representative of split learning, divides the model, potentially better protecting privacy and reducing the computational burden on clients while achieving commendable performance on this dataset.

On the most difficult PADUFES dataset, most algorithms struggle, but FedALA (with an accuracy of 0.6065) and HarmoFL (with an accuracy of 0.5957) still manage to achieve relatively good results. FedALA may be more adept at handling the extremely non-independent and identically distributed (non-IID) data distributions in PADUFES by adaptively aggregating local models. Notably, some algorithms that performed well on simple datasets, such as FedGH and FLOP, experience a sharp drop in accuracy to around 0.50 on PADUFES, becoming the worst-performing methods. This strongly suggests that their mechanisms may be unable to adapt to highly heterogeneous, low-quality data environments and may even amplify noise or bias in the data.

Finally, the impact of data heterogeneity can be clearly observed from the accuracy of individual clients. On ISIC2020, all clients achieve accuracies above 0.96, with closely clustered values. In contrast, the client metrics on PADUFES are highly dispersed. For example, under FedALA, client accuracies range from 0.5760 to 0.6304, and under GPFL, one client's accuracy drops as low as 0.4456. This visually reveals the significant differences in data distribution and quality across different clients (which may represent different hospitals or devices), which is precisely the core challenge that federated learning aims to address.

\subsection{Algorithmic Efficiency Analysis}\label{exp-3}

\begin{figure}
	\centering
	\includegraphics[width=\linewidth,scale=1.0]{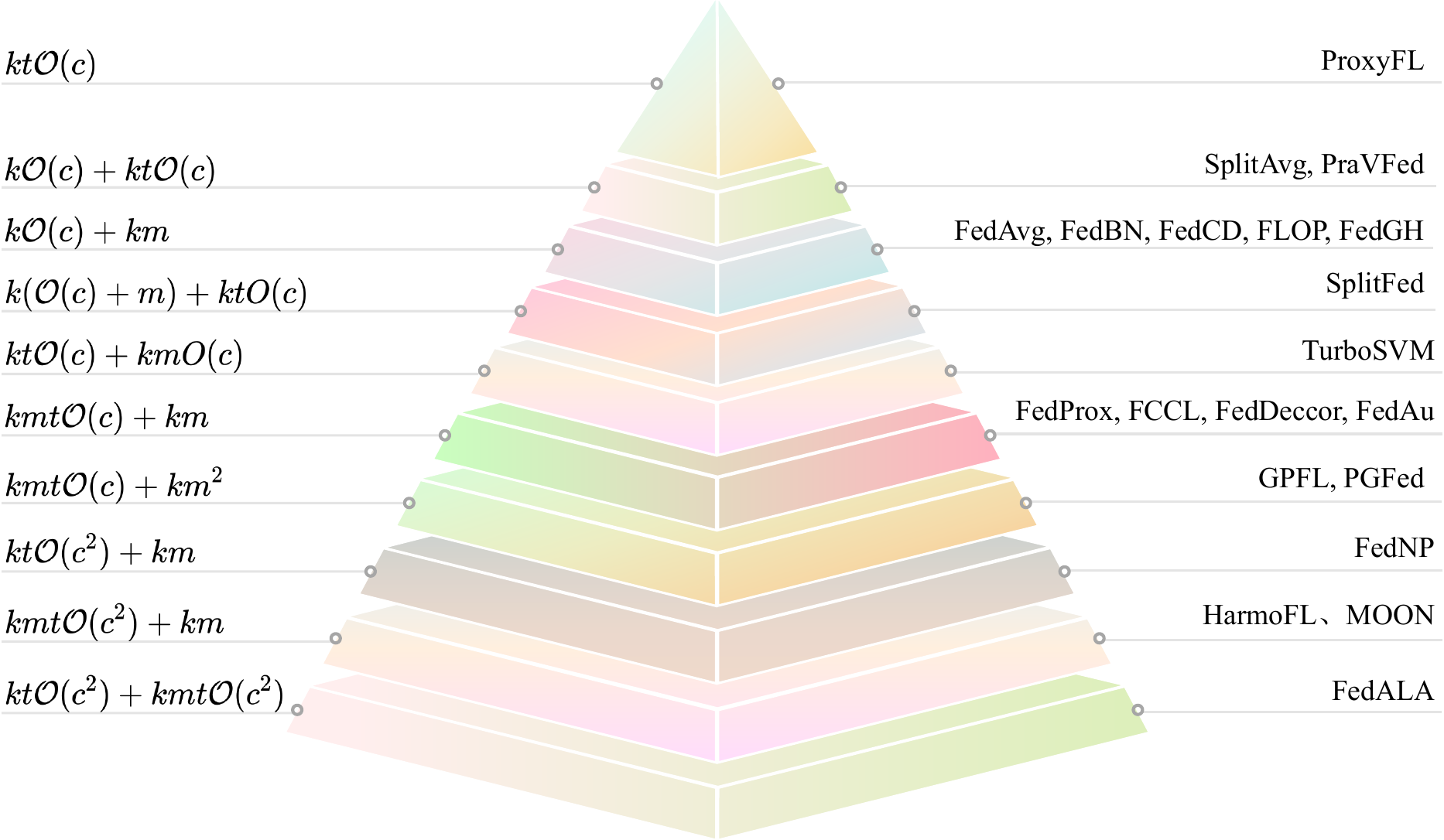}
	\caption{The theoretical time complexity of federated learning; a lower position on the time complexity axis indicates a longer theoretical time consumption.}
	\label{complexity}
\end{figure}

\subsubsection{Analysis of Theoretical Complexity}

\begin{figure}
	\centering
	\includegraphics[width=\linewidth,scale=1.0]{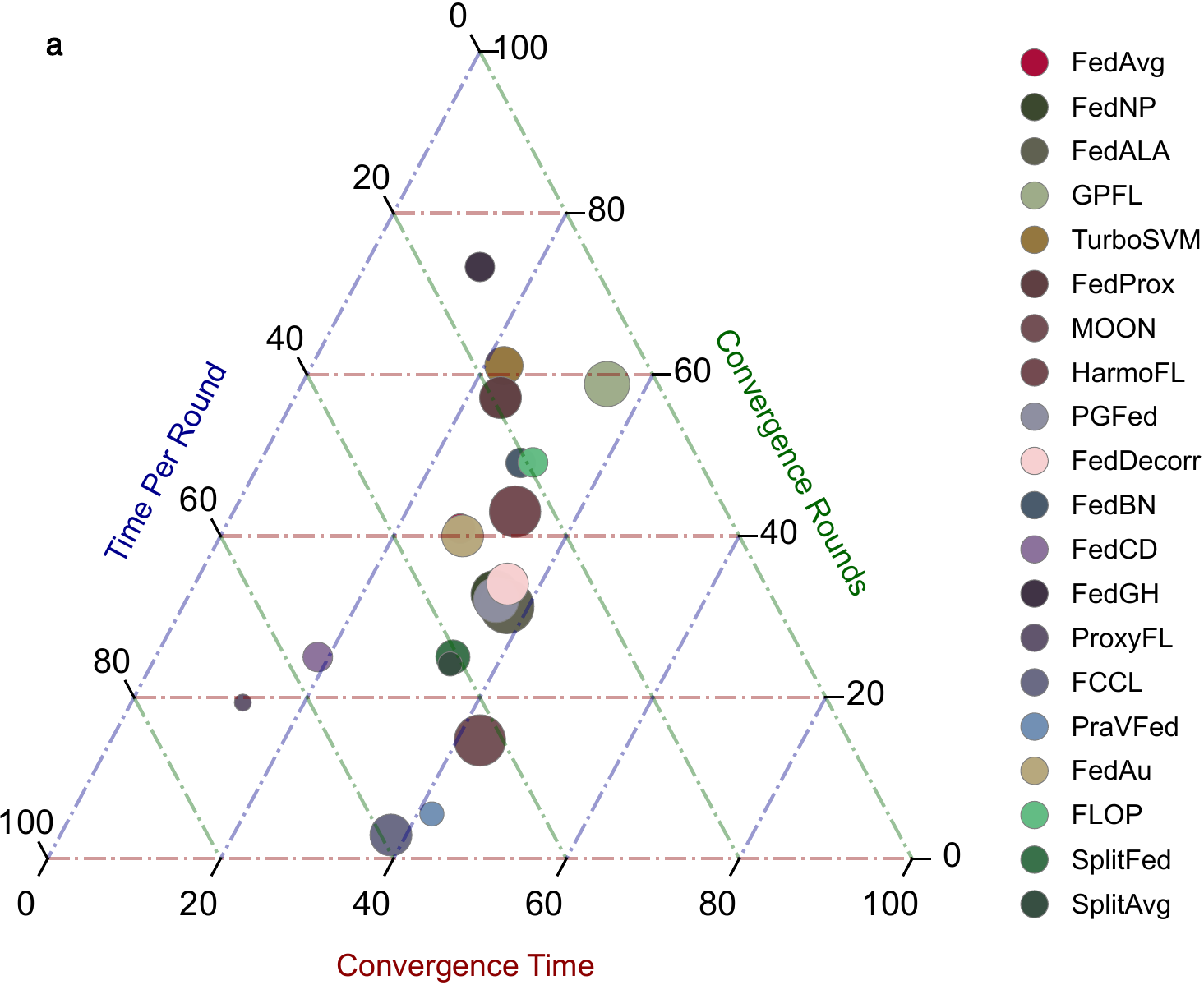}
    \includegraphics[width=\linewidth,scale=1.0]{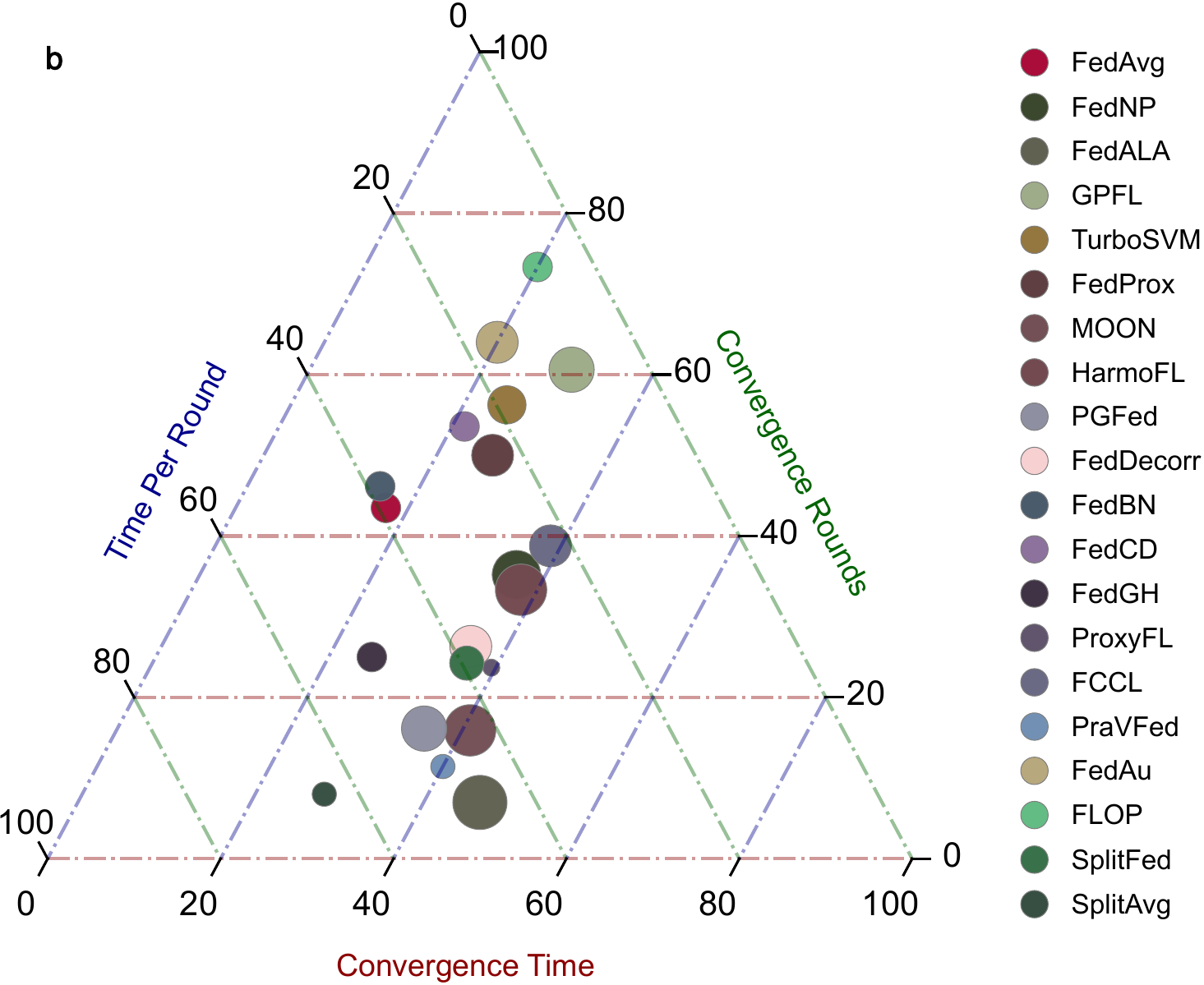}
	\caption{Ternary analysis of the efficiency of federated learning algorithms on the HAM10000 and OrganMNIST datasets. The three vertices of the ternary diagram represent three evaluation dimensions: the number of convergence rounds, the convergence time, and the average per-round runtime of the algorithm. The position of each point in the diagram is determined by the normalized values of these three metrics: the closer a point is to a vertex, the higher the ranking of the algorithm on that metric, while the closer a point is to an edge, the lower the ranking on the corresponding metric. Subfigure \textbf{a} represents the HAM10000 dataset, and subfigure \textbf{b} represents the OrganMNIST dataset.}
	\label{complexity_HO}
\end{figure}

In the practical deployment of federated learning, particularly in medical image analysis, conducting a computational complexity analysis of the algorithms is crucial. Theoretical performance gains will lose their practical value in resource-constrained clinical environments if they come at the cost of prohibitive computational or communication overhead. Operational efficiency directly determines model training cycles, energy consumption, and computational costs, which are critical factors affecting its integration into real-world clinical workflows. Therefore, we analyzed the theoretical complexity and actual convergence time of the FL algorithms in MobenFL, with the results shown in Fig.~\ref{complexity} and Fig.~\ref{complexity_HO}.

 In the pyramid structure of Fig.~\ref{complexity}, a lower level indicates higher theoretical complexity. Here, $k$ denotes the number of clients participating in training, $t$ represents the number of local iterations on each client, $m$ refers to the number of model parameters, and $\mathcal{O}(c)$ denotes the computational complexity required for forward and backward propagation per sample.
The time complexity of federated learning algorithms primarily depends on two components: client-side training and server-side model parameter aggregation. Therefore, we use a "+" to separate these two parts, as shown in Fig.~\ref{complexity}.
Algorithms with time complexity similar to FedAvg include FedBN, FedCD, FLOP, and FedGH. These methods mainly introduce specific optimizations in the model parameter aggregation phase without incurring additional computational overhead. In contrast, FedProx, FCCL, FedDecorr, and FedAu incorporate regularization constraints related to model parameters during training.
Notably, ProxyFL, as a decentralized algorithm, does not involve server-side parameter aggregation. FedNP, HarmoFL, and MOON introduce additional model training steps during client-side training, significantly increasing time complexity. Conversely, FedALA, GPFL, and PGFed incorporate extra model training in the server-side aggregation phase.
SplitAvg and SplitFed, derived from split learning, require forward propagation of the model on both the client and server sides, contributing an $\mathcal{O}(c)$ time complexity for each part. Additionally, SplitFed further requires an extra round of model parameter aggregation among clients.

\subsubsection{Evaluation of Actual Runtime Efficiency}

\begin{table}[htbp]
  \centering
  \renewcommand\arraystretch{1.3}
  \caption{Comparison of Actual Convergence Time and Convergence Rounds of FL Algorithms on HAM10000 and OrganMNIST Datasets.}
  \label{Runtime}
  \begin{tabular}{lcccc}
    \toprule
    Datasets & \multicolumn{2}{c}{HAM10000} & \multicolumn{2}{c}{OrganMNIST} \\
    \cmidrule(lr){2-3} \cmidrule(lr){4-5}
    Methods & Time (h) & Rounds & Time (h) & Rounds \\
    \midrule
    FedAvg     & 0.96 & 147 & 2.90 & 110 \\
    FedNP      & 3.61 & 214 & 4.32 & 126 \\
    FedALA     & 2.14 & 184 & 7.03 & 88  \\
    GPFL       & 1.28 & 254 & 3.86 & 181 \\
    TurboSVM   & 0.85 & 160 & 3.60 & 163 \\
    FedProx    & 0.95 & 165 & 3.74 & 154 \\
    MOON       & 1.67 & 133 & 6.51 & 98  \\
    HarmoFL    & 1.60 & 214 & 5.32 & 133 \\
    PGFed      & 2.87 & 211 & 3.91 & 91  \\
    FedDecorr  & 1.82 & 206 & 6.42 & 126 \\
    FedBN      & 1.05 & 167 & 2.05 & 95  \\
    FedCD      & 0.84 & 121 & 3.17 & 139 \\
    FedGH      & 0.65 & 133 & 1.32 & 63  \\
    ProxyFL    & 0.72 & 96  & 4.60 & 108 \\
    FCCL       & 1.48 & 50  & 6.67 & 180 \\
    PraVFed    & 1.63 & 77  & 4.48 & 90  \\
    FedAu      & 1.04 & 157 & 3.10 & 144 \\
    FLOP       & 1.13 & 179 & 2.32 & 115 \\
    SplitFed   & 1.24 & 142 & 4.27 & 101 \\
    SplitAvg   & 1.16 & 135 & 3.47 & 70  \\
    \bottomrule
  \end{tabular}
\end{table}

Next, we will analyze the combination of theoretical complexity levels and the actual convergence time on the HAM10000 and OrganMNIST datasets, with the results shown in Fig.~\ref{complexity_HO} and Table \ref{Runtime} . Overall, there is a general positive correlation between theoretical complexity and actual convergence time, meaning that methods with higher theoretical complexity tend to require longer total training times.

ProxyFL, located at the top of the theoretical complexity hierarchy, serves as a representative example. On the HAM10000 dataset, it indeed demonstrated the advantage of low complexity, with a total convergence time of only 0.72 hours, ranking as the second fastest. However, on the OrganMNIST dataset, its convergence time surged to 4.6 hours, making it one of the slower methods. In contrast, FedAvg, which is also considered low complexity, exhibited robust and fast convergence on both datasets, establishing itself as a highly efficient benchmark. FedGH performed even more impressively, achieving the shortest total convergence time on both datasets, with its low-complexity design yielding significant practical benefits.

Among the methods with moderately low complexity, FedProx and TurboSVM demonstrated good balance. Their convergence times on both datasets were in the middle range, neither becoming exceptionally slow due to increased complexity nor potentially sacrificing performance gains from regularization. FedBN and FedCD also fall into this category, with convergence speeds comparable to or even faster than FedAvg, indicating that their targeted handling of client data distributions is both efficient and effective. FLOP's performance closely resembled FedAvg, showing stable efficiency. However, FedDecorr and MOON exhibited certain overheads, particularly on OrganMNIST, where convergence times were significantly longer, likely due to the additional contrastive learning introduced to enhance model performance.

When examining methods that employ split learning frameworks, such as SplitAvg and SplitFed, it is evident that they did not demonstrate significant advantages in efficiency. Their convergence times were generally higher than FedAvg, revealing that the communication bottlenecks introduced by model splitting may offset some of the computational benefits. The performances of PraVFed and FCCL further corroborate this point, especially FCCL, which had the highest per-round time consumption on HAM10000 among all methods. Although its total number of rounds was small, the overall time remained considerable, indicating substantial computational or communication overhead within each round.

Moving into the higher tiers of theoretical complexity, the situation becomes more intricate. Both GPFL and PGFed involve personalization or graph structure processing, and their convergence times fall within the medium range. GPFL performed acceptably on HAM10000 but required many rounds on OrganMNIST, resulting in a relatively long total time. PGFed's convergence times on both datasets were close to 3-4 hours, placing it in the slower category, as its complexity indeed incurred considerable time costs. HarmoFL's performance was similar to FedDecorr, with slower convergence on OrganMNIST.

Finally, among the methods with the highest theoretical complexity, we observed the greatest variability. FedNP ranked among the slowest in total convergence time on both datasets, particularly on HAM10000, where it was the slowest, with its high complexity directly translating into prolonged waiting times. FedALA, on the other hand, presented the most extreme case: on HAM10000, its convergence time was acceptable (2.14 hours), far faster than FedNP; however, on OrganMNIST, it became the slowest of all methods (7.03 hours), with the highest per-round time consumption. This strongly suggests that high-complexity algorithms like FedALA are highly sensitive to dataset characteristics, potentially performing well in one domain while encountering severe efficiency bottlenecks in another.

In summary, theoretical complexity is a useful but not entirely accurate indicator for predicting the convergence time of federated learning algorithms. Methods like FedGH achieve an ideal combination of low complexity and high speed, while FedAvg continues to serve as a reliable benchmark due to its robust efficiency. Many methods (e.g., FedProx, TurboSVM, FedBN) strike a good balance between complexity and efficiency. However, some methods, particularly high-complexity ones like FedNP and FedALA, as well as FCCL with its substantial per-round overhead, indeed require users to pay significant time costs for potential accuracy improvements. Moreover, these costs can fluctuate dramatically across different datasets. Therefore, in practical applications, algorithm selection must be based on specific data characteristics and a comprehensive trade-off among training time, computational resources, and model performance.

\subsection{Robustness under Privacy-Preserving Constraints}\label{exp-4}

In clinical federated learning, the adoption of privacy-preserving algorithms is a compliance prerequisite, yet it inevitably diminishes model performance. Therefore, evaluating the robustness of algorithms becomes crucial~\citep{10891624}. Robustness not only measures the extent of performance degradation but, more importantly, examines the algorithm's stability under various perturbations. A truly robust algorithm must maintain reliable operation under both privacy constraints and real-world disturbances, which is essential for its safe and practical deployment.

\subsubsection{Evaluation with Differential Privacy}
\begin{table*}[htbp] 
  \centering
  \renewcommand\arraystretch{1.3}
  \caption{Performance Comparison on HAM10000 and OrganMNIST Datasets under Differential Privacy Constraints.} 
  \label{results_dp}
  \begin{tabular}{lcccccc}
    \toprule
     Datasets & \multicolumn{3}{c}{HAM10000} & \multicolumn{3}{c}{OrganMNIST} \\
    \cmidrule(lr){2-4} \cmidrule(lr){5-7}
    DP settings & None & $\epsilon=0.01$ & $\epsilon=0.1$ & None & $\epsilon=0.01$ & $\epsilon=0.1$ \\
    \midrule
    FedAvg  & 77.55±{\scriptsize0.83} & 77.55±{\scriptsize 1.71} & 66.75±{\scriptsize 1.39} & 82.00±{\scriptsize 0.45} & 80.29±{\scriptsize 1.02} & 63.23±{\scriptsize 1.27} \\
    FedBN       & 77.30±{\scriptsize 0.64} & 76.70±{\scriptsize 1.12} & 54.15±{\scriptsize 9.86} & 80.19±{\scriptsize 1.72} & 80.22±{\scriptsize 0.78} & 55.70±{\scriptsize 0.65} \\
    FedProx   & 75.35±{\scriptsize 1.56} & 72.35±{\scriptsize 1.65} & 64.60±{\scriptsize 2.95} & 81.42±{\scriptsize 1.05} & 72.66±{\scriptsize 1.11} & 57.57±{\scriptsize 1.39} \\
    HarmoFL     & \textbf{79.45±{\scriptsize 1.48}} & \underline{78.15±{\scriptsize 2.11}} & 65.60±{\scriptsize 1.63} & 80.88±{\scriptsize 1.31} & \underline{82.25±{\scriptsize 0.52}} & 64.75±{\scriptsize 1.27} \\
    FedGH     & 72.15±{\scriptsize 2.51} & 72.40±{\scriptsize 2.35} & 64.00±{\scriptsize 5.36} & 78.49±{\scriptsize 1.60} & 76.46±{\scriptsize 1.27} & 54.43±{\scriptsize 2.18} \\
    GPFL        & 78.50±{\scriptsize 0.72} & 68.60±{\scriptsize 8.37} & 52.95±{\scriptsize 6.08} & \underline{83.14±{\scriptsize 1.09}} & 77.90±{\scriptsize 3.08} & 51.94±{\scriptsize 5.65} \\
    PGFed       & 77.60±{\scriptsize 1.10} & 76.85±{\scriptsize 1.49} & \underline{67.10±{\scriptsize 1.28}} & 81.43±{\scriptsize 1.45} & 81.87±{\scriptsize 0.96} & \underline{65.91±{\scriptsize 1.16}} \\
    TurboSVM    & 76.35±{\scriptsize 2.94} & 76.70±{\scriptsize 1.91} & 66.95±{\scriptsize 1.49} & 81.55±{\scriptsize 0.44} & 80.33±{\scriptsize 0.74} & 63.54±{\scriptsize 1.17} \\
    FedDecorr   & \underline{78.85±{\scriptsize 1.10}} & 77.00±{\scriptsize 0.94} & 64.60±{\scriptsize 1.93} & 81.68±{\scriptsize 1.08} & 81.52±{\scriptsize 1.22} & 58.78±{\scriptsize 0.93} \\
    FedAu     & 75.15±{\scriptsize 2.85} & 75.75±{\scriptsize 2.15} & 65.20±{\scriptsize 1.09} & 81.68±{\scriptsize 1.81} & 80.34±{\scriptsize 0.79} & 63.61±{\scriptsize 1.15} \\
    SplitFed     & 78.50±{\scriptsize 1.70} & \textbf{78.25±{\scriptsize 1.42}} & \textbf{79.01±{\scriptsize 0.52}} & \textbf{83.83±{\scriptsize 0.49}} & \textbf{82.47±{\scriptsize 0.64}} & \textbf{83.00±{\scriptsize 0.47}} \\
    \bottomrule
  \end{tabular}
\end{table*}

To evaluate the security and practicality of deploying federated learning algorithms in real-world medical scenarios, we systematically examined the performance of various FL algorithms under Differential Privacy~\citep{abadi2016deep} constraints. In this study, we simulate differential privacy by adding normally distributed noise to the model parameters transmitted from the client to the server, with the formula as follows:

\begin{equation}
    w' = w + \epsilon \cdot \mathcal{N}(0, \sigma^{2}), 
\end{equation}
where $\mathcal{N}(0, \sigma^{2})$ is the normal distribution noise, and $\epsilon$ is the scaling factor used to control the magnitude of the noise. Considering the training and evaluation costs for all algorithms and datasets, we select only two datasets, HAM10000 and OrganMNIST, along with 11 algorithms including FedDecorr and FedAu. 

Table~\ref{results_dp} clearly demonstrates the performance of various methods on two medical datasets, HAM10000 and OrganMNIST, after introducing differential privacy constraints into the federated learning framework. Overall, the most critical conclusion is that the introduction of differential privacy almost always leads to a decline in model performance, and a stronger privacy protection intensity typically results in more significant performance degradation. This aligns with our basic expectations, as the noise injected by the DP mechanism interferes with the updates of model parameters or gradients, thereby reducing the learning efficiency and final accuracy of the model.

However, against this general trend, different methods exhibit distinctly different levels of robustness. In the non-DP setting, the performance among the methods is highly competitive. HarmoFL, FedDecorr, GPFL, and SplitFed hold slight advantages on the two datasets respectively, but the leading margins are small. This indicates that in an ideal, privacy-unconstrained environment, many advanced methods can effectively handle the non-IID data problem and achieve a similar performance ceiling. However, when DP constraints are introduced, the robustness of the methods becomes clearly distinguishable. It is particularly noteworthy that the SplitFed method demonstrates exceptional performance across almost all settings, especially under the strictest privacy constraint, where its performance is truly disruptive. On HAM10000, while the performance of all other methods drops significantly to below 0.67, SplitFed not only avoids a decline but achieves a remarkable accuracy of 0.7901, even significantly surpassing its own performance without DP. On OrganMNIST, its accuracy of 0.83 similarly far exceeds that of other methods and remains comparable to its own performance without DP.

In stark contrast to the robustness of SplitFed, some methods that perform excellently in non-DP environments, such as FedBN and GPFL, experience an "avalanche-like" performance drop under strong DP constraints. This is particularly evident on the HAM10000 dataset, where their performance decreases by over 20 percentage points. This suggests that the components these methods rely on may be highly sensitive to noise in gradients or parameters, as the DP noise disrupts the carefully maintained client-specific statistics or the stability of their generative processes.
In comparison, while classical methods like FedAvg and the straightforward TurboSVM do not achieve top performance without DP, they demonstrate "mediocre robustness" in DP environments. Their performance degradation is relatively gradual, and under the strictest privacy constraint ($\epsilon$=0.1), their results surpass those of most more complex methods. This highlights the inherent stability advantage that simpler approaches may possess when facing strong interference.

\subsubsection{Analysis across Privacy Budgets}
\begin{figure*}
	\centering
	\includegraphics[width=\linewidth,scale=1.0]{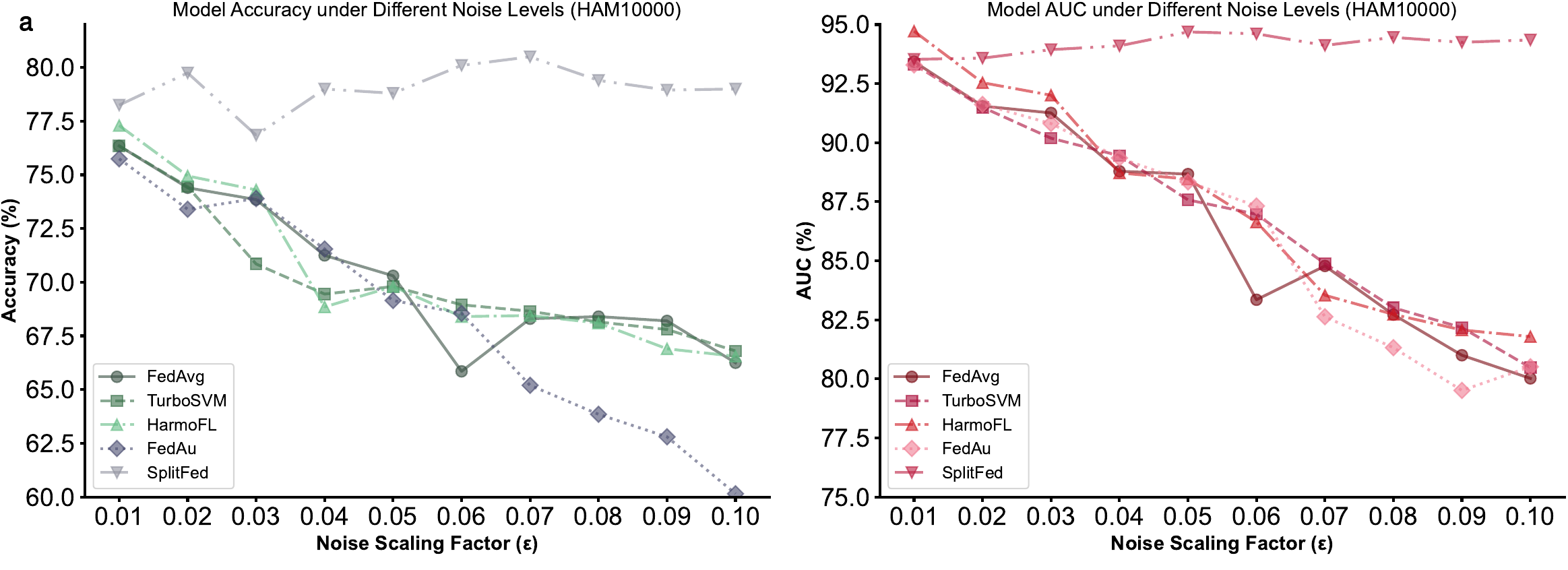}
    \includegraphics[width=\linewidth,scale=1.0]{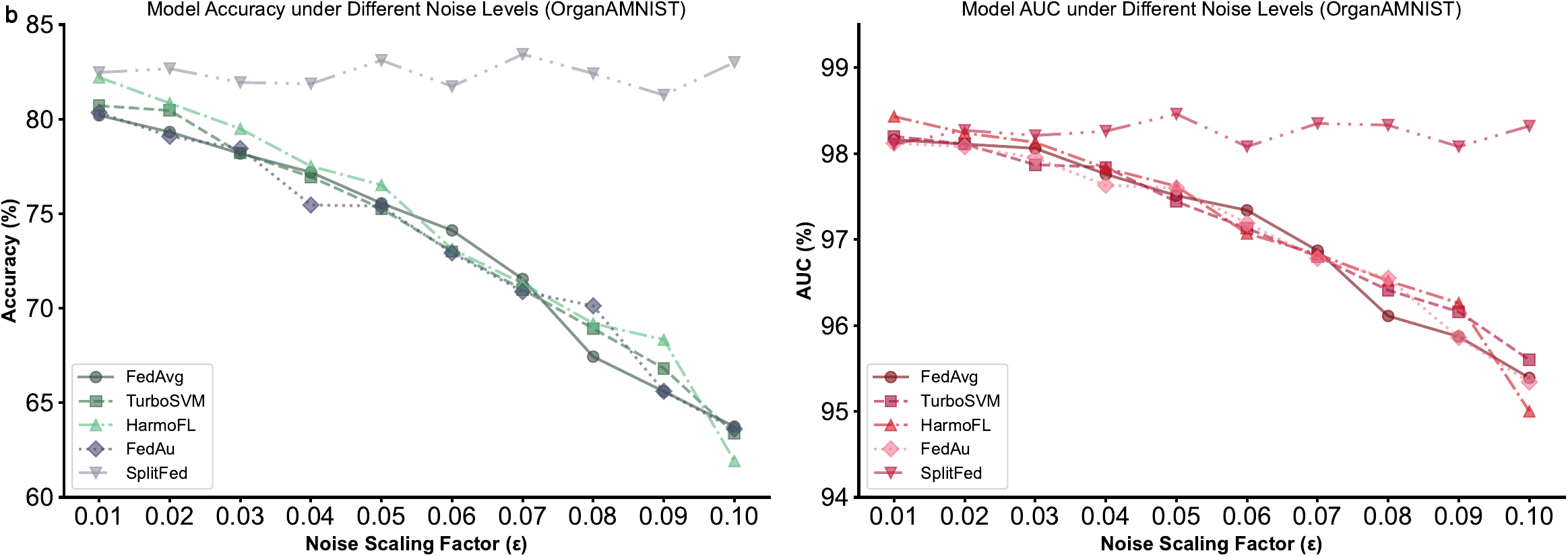}
	\caption{\textbf{a}: Performance  of accuracy and AUC on the HAM10000 dataset under different noise scaling factors. \textbf{b}: Performance evaluation of accuracy and AUC on the OrganMNIST dataset under different noise scaling factors.}
	\label{dp_level}
\end{figure*}

Based on preliminary observations under different differential privacy constraints, we further analyzed the robustness of various federated learning methods by continuously varying the noise scaling factor, with results shown in Fig.~\ref{dp_level}.a and Fig.~\ref{dp_level}.b.
As the noise scaling factor increases (from 0.01 to 0.1), the accuracy and AUC values of all methods generally exhibit a declining trend, but the extent and pattern of decline vary across algorithms. 

On the HAM10000 dataset in Fig.~\ref{dp_level}.a, the accuracy of FedAvg, TurboSVM, HarmoFL, and FedAu decreases from approximately 0.76-0.77 to 0.66-0.65, while their AUC drops from 0.93-0.94 to 0.80-0.81, indicating a significant negative impact of noise on performance. In contrast, SplitFed demonstrates exceptional robustness, maintaining stable accuracy and AUC levels on HAM10000 at around 0.78-0.80 and 0.93-0.95, respectively, with almost no impact from increasing noise.
A similar trend is evident on the OrganMNIST dataset: traditional methods such as FedAvg and HarmoFL see their accuracy decline from 0.80-0.82 to 0.63-0.62, and their AUC drops from around 0.98 to approximately 0.95. Meanwhile, SplitFed maintains stable accuracy and AUC at around 0.81-0.83 and 0.98, respectively. This contrast highlights the sensitivity of most methods to noise, while SplitFed’s architecture may inherently provide a noise isolation mechanism.

Different categories of federated learning methods exhibit significant variations in their responses to noise. The Split Learning approach (SplitFed) maintained the most stable performance across both datasets, benefiting from its model-splitting architecture which likely confines noise to local clients and reduces global model perturbation.
Regularization-Driven methods (HarmoFL) performed well under low noise levels (e.g., an initial accuracy of 0.773 on HAM10000), but their performance deteriorated markedly as noise intensified (final accuracy 0.6655), indicating that their regularization mechanisms cannot adequately compensate for gradient bias under strong noise interference.
Aggregation Strategies methods (FedAu) demonstrated the highest sensitivity to noise, particularly on HAM10000 where accuracy plummeted from 0.7575 to 0.6520. This suggests their aggregation strategies are prone to failure in noisy environments.
Additional Network Structures methods (TurboSVM) exhibited moderate robustness with a relatively gradual performance decline (e.g., HAM10000 accuracy decreasing from 0.7635 to 0.6680), though their final performance remained inferior to SplitFed.
As a baseline, FedAvg displayed a characteristic linear degradation pattern, underscoring the necessity for advanced methods to incorporate robustness-focused design.

\subsection{Benchmark Hyperparameter Configuration}\label{subsec2.6}

\begin{figure}[htbp]  
	\centering
	\includegraphics[width=\linewidth,scale=1.0]{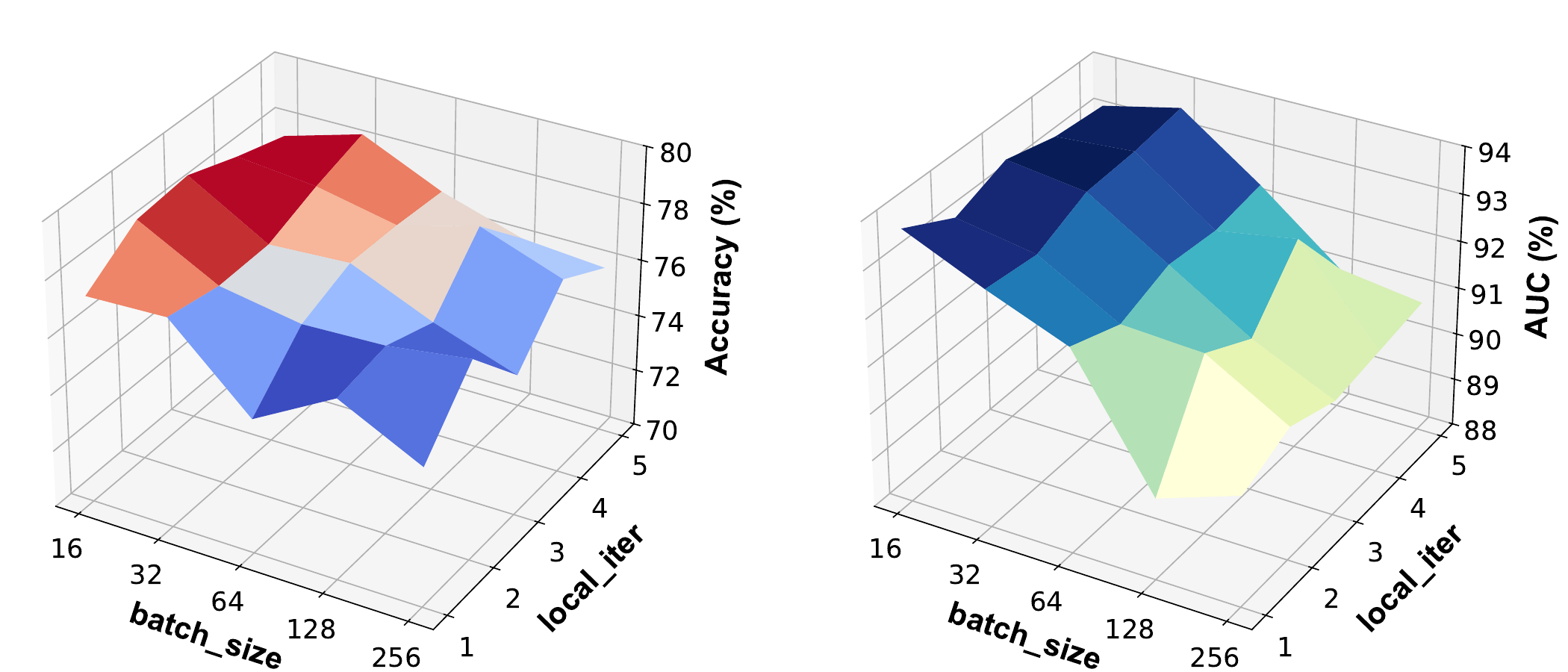}
	\caption{Performance evaluation of accuracy and AUC on the HAM10000 dataset under different noise scaling factors.}
	\label{batch_HAM10000}
\end{figure}

\begin{figure}[htbp]  
	\centering
	\includegraphics[width=\linewidth,scale=1.0]{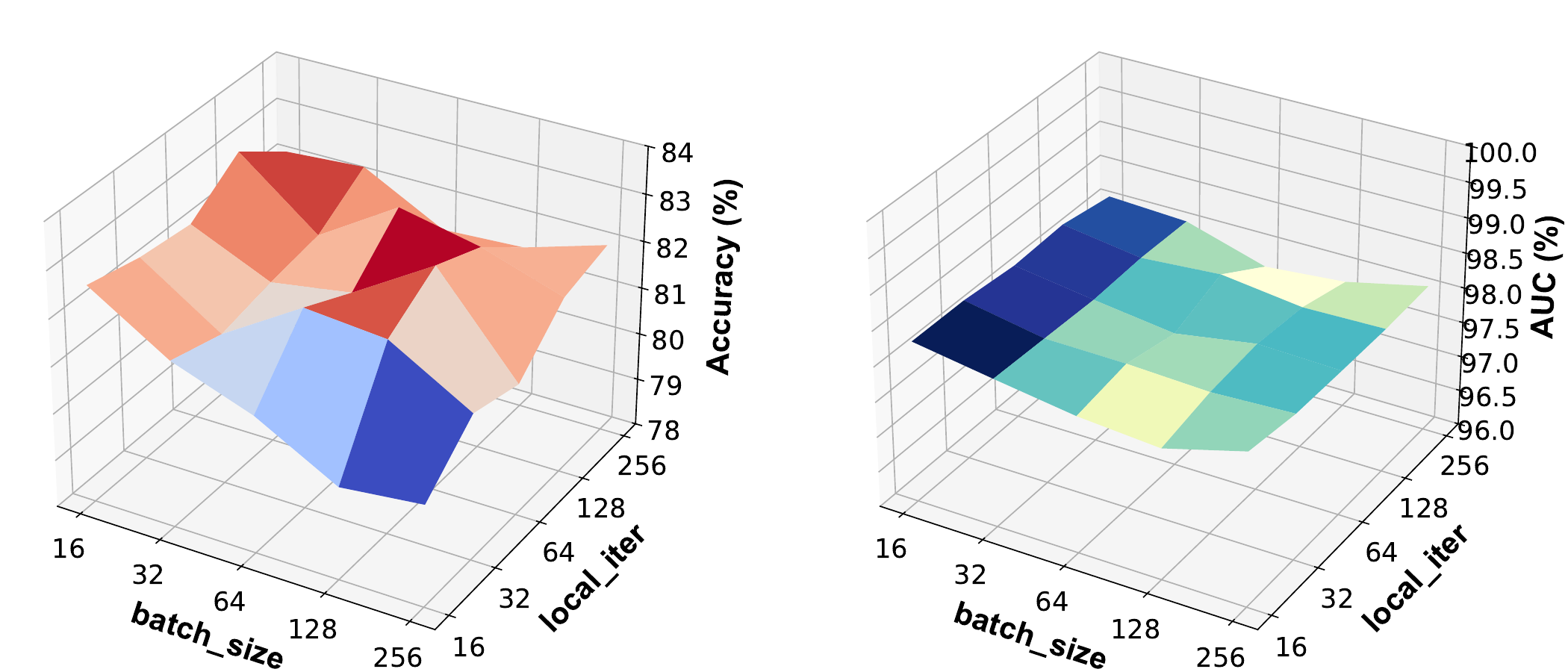}
	\caption{Performance evaluation of accuracy and AUC on the OrganMNIST dataset under different noise scaling factors.}
	\label{batch_OrganMNIST}
\end{figure}

The configuration of hyperparameters is crucial to the operation of FL algorithms. To establish a unified benchmark for hyperparameter configuration and identify optimal parameters, we conduct a sensitivity analysis focusing on two important hyperparameters: \textbf{batch size} and \textbf{local iterations}.
This study evaluate the impact of adjusting the number of local iterations and batch size in the FedAvg algorithm on the OrganMNIST and HAM10000 datasets, the results are shown in Fig.~\ref{batch_HAM10000} and Fig.~\ref{batch_OrganMNIST}. On the HAM10000 dermoscopic image dataset, the impact of hyperparameters on model performance exhibits highly sensitive and well-defined patterns. Batch size emerges as the dominant factor influencing performance: as the batch size increases from 16 to 256, both model accuracy and AUC values show a systematic decline, regardless of the setting for local iterations. For instance, under the condition of local\_iter=3, the best performance is achieved with a batch size of 16 (Acc=78.75\%, AUC=93.58\%), whereas performance significantly deteriorates with a batch size of 256 (Acc=75.00\%, AUC=90.46\%). The mechanism for this sensitivity lies in the HAM10000 dataset's inherent characteristics: the dermatological features exhibit significant morphological diversity and class imbalance. Smaller batch sizes generate gradient estimates with higher variance, which in turn allows the model to explore a richer feature space and avoids converging to suboptimal local minima due to overly averaged gradient directions in larger batches. Meanwhile, the local iteration count displays a distinct inverted U-shaped influence curve: performance peaks at 3 iterations, while too few iterations (1 iteration) result in insufficient client learning, and too many iterations (5 iterations) lead to severe client drift. This occurs as the client model overfits to local non-IID data, undermining the convergence stability of federated aggregation, a phenomenon that is particularly pronounced with larger batch sizes (e.g., when batch\_size=256, increasing local\_iter from 3 to 5 results in a 0.02 drop in accuracy).


On the OrganMNIST organ CT image dataset in Fig.~\ref{batch_OrganMNIST}, the influence of hyperparameters reveals a different pattern of sensitivity. Overall performance demonstrates higher baseline levels and greater robustness, yet the optimal configuration trends still point in the same direction.
The impact of batch size exhibits a marginal diminishing characteristic: a batch size of 16 achieved optimal performance in 7 out of 12 experimental groups, but when the batch size increased to 32, the performance degradation was significantly less pronounced compared to the HAM10000 dataset. This phenomenon may stem from the dataset's relatively standardized image structure and the high consistency of organ anatomical features, which enhance the model's tolerance for variance in gradient estimation.
The influence of local iterations shows a plateau effect: performance remains high within the range of 2 to 4 iterations (e.g., with a batch size of 16, the accuracy difference between 3 and 4 iterations is less than 0.007), whereas a clear performance gap appears at 1 iteration (average accuracy decreases by 0.015). Notably, a specific performance peak (Acc=82.78\%) was observed with a batch size of 128 and 3 local iterations. This suggests that, under specific parameter combinations, large-batch training and moderate iteration may produce a positive synergy—a phenomenon not observed in the more sensitive HAM10000 dataset.

Based on empirical patterns across datasets, we propose a three-level configuration scheme:
The core benchmark configuration is (batch\_size=16, local\_iter=3). This combination achieves the global optimum on HAM10000 and falls within the optimal range (top 25\%) on OrganMNIST, demonstrating the best universality.
The performance‑prioritized configuration is (batch\_size=16, local\_iter=4), which is suitable for scenarios with sufficient computational resources and pursuit of peak performance.
The efficiency‑optimized configuration is (batch\_size=32, local\_iter=2) reduces per‑round communication and computation costs by approximately 33\% while keeping performance degradation controllable (average accuracy drop 1.5\%).
Configuration selection should follow the principle of data complexity: for complex imbalanced datasets similar to HAM10000, the core benchmark configuration must be strictly adopted; for well‑structured datasets like OrganMNIST, flexible selection among the three‑level configurations is allowed.

\section{Discussion}

\begin{figure}
	\centering
	\includegraphics[width=\linewidth,scale=1.0]{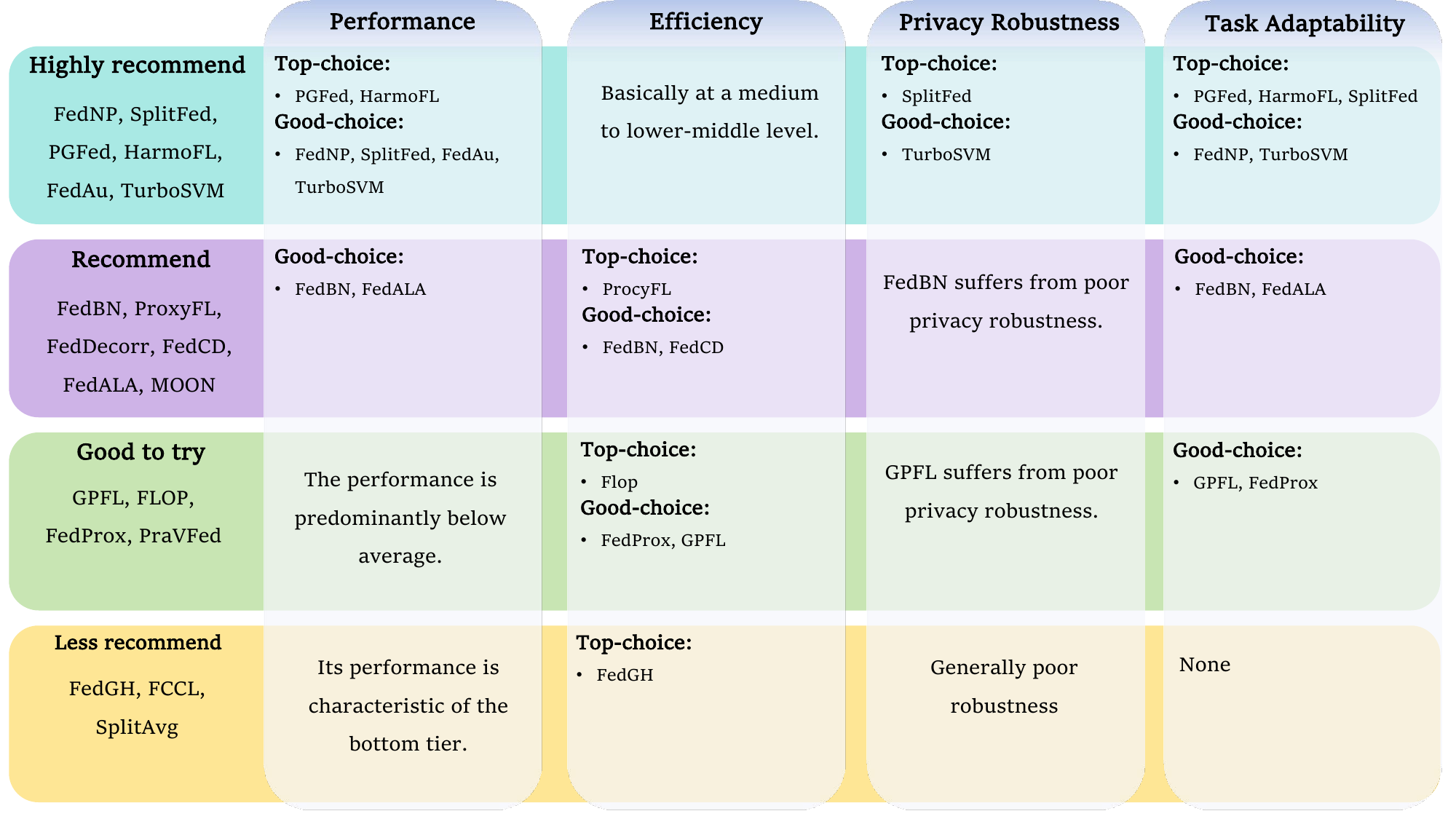}
	\caption{Recommended algorithms for applying federated learning methods in medical image classification.}
	\label{disscusion}
\end{figure}

Federated learning presents a promising paradigm for medical imaging by enabling collaborative model training without data sharing. However, the practical deployment of FL hinges on navigating the complex interplay among model performance, computational efficiency, privacy guarantees, and generalization ability. Based on our benchmark results, which rigorously evaluated these aspects, we discuss the nuanced landscape of algorithm selection. The specific scenarios each FL algorithm is suited for are illustrated in Fig.~\ref{disscusion}.

Under the independent and identically distributed (IID) setting, algorithms such as FedNP, SplitFed, PGFed, and HarmoFL, which demonstrate excellent performance, also perform very well under the non-IID setting (as shown in Sec.~\ref{exp-1}), with the exception of PGFed. On the other hand, FedAu and TurboSVM show strong performance under non-IID conditions while exhibiting suboptimal performance under IID settings. Additionally, both PGFed and HarmoFL perform outstandingly in evaluations of same-tissue-different-disease and same-modality-different-disease scenarios, achieving optimal or suboptimal results in multiple tasks (as shown in Sec.~\ref{exp-2}).
In terms of convergence speed (as shown in Sec.~\ref{exp-3}), FedNP, SplitFed, PGFed, and HarmoFL are among the slower algorithms, while FedAu and TurboSVM fall into the medium category. The above analysis indicates that it is difficult to balance algorithmic efficiency and performance, as no single algorithm can train a model that is both fast and accurate.
It is worth noting that under the influence of differential privacy noise factors, SplitFed's performance does not degrade significantly, and TurboSVM experiences a smaller performance decline compared to other methods (as shown in Sec.~\ref{exp-4}).

FedBN, ProxyFL, FedDecorr, FedCD, FedALA, and MOON demonstrate satisfactory performance in evaluations; however, their improvements over the baseline algorithm FedAvg are relatively limited (as shown in Sec.~\ref{exp-1}). Moreover, under the non-IID setting, these methods generally underperform compared to FedAvg. On the other hand, FedBN and FedALA also deliver competitive results in evaluations involving same-modality-different-disease and same-disease-different-dataset scenarios (as shown in Sec.~\ref{exp-2}).
In terms of algorithmic efficiency, FedBN, ProxyFL, FedDecorr, and FedCD achieve relatively fast training speeds, whereas FedALA and MOON exhibit slower convergence (as shown in Sec.~\ref{exp-3}).
In contrast, algorithms such as GPFL, FLOP, FedProx, and PraVFed yield less satisfactory performance in evaluations, showing no notable advantages in either IID or non-IID settings. That said, GPFL and FedProx still display some competitive performance in specific scenarios such as same-tissue-different-disease and same-modality-different-disease tasks. Regarding computational efficiency, with the exception of PraVFed, which is relatively slow, GPFL, FLOP, and FedProx all fall into the category of faster algorithms (as shown in Sec.~\ref{exp-3}).

FedGH, FCCL, and SplitAvg lag significantly behind other algorithms in performance evaluations, both in main comparative experiments and other tasks. Among them, SplitAvg achieves moderate efficiency, while FedGH falls into the faster category. It is worth noting that FCCL exhibits relatively slow efficiency, which may be attributed to its scenario-specific design and high algorithmic complexity, thereby resulting in relatively poor reproducibility.

Overall, in practical scenarios of federated learning applied to medical imaging analysis and deployment, if model performance is the primary focus, researchers may consider selecting algorithms such as FedNP, SplitFed, PGFed, and HarmoFL. In cases where data distributions across different centers exhibit significant heterogeneity, FedAu and TurboSVM can be prioritized. For scenarios where training efficiency is the main concern rather than high performance, FedBN, ProxyFL, FedDecorr, and FedCD are recommended choices. Taking into account the balance between model performance and efficiency, FedAvg also remains a competitive option.

\section{Conclusion}

In this work, we present the most comprehensive federated medical imaging benchmark to date, MobenFL. It encompasses 22 medical imaging datasets covering 12 organs across the human body and integrates 20 different types of federated learning algorithms. MobenFL comprehensively evaluates federated learning algorithms in terms of performance, efficiency, privacy robustness, and various clinical scenarios. Based on the evaluation results, we recommend suitable federated algorithms for different clinical application scenarios, aiming to bridge the gap in the practical implementation of federated learning in the medical field. We hope that this work will assist medical researchers in conducting more in-depth studies in this domain more rapidly and efficiently.

\section*{Acknowledge}

This work was supported in part by the National Natural Science Foundation of China under Grant 62172444, in part by the Natural Science Foundation of Hunan Province under Grant 2022JJ30753, in part by the Central South University Innovation-Driven Research Programme under Grant 2023CXQD018, and in part by the High Performance Computing Center of Central South University. 

\bibliographystyle{elsarticle-harv.bst}

\bibliography{cas-refs}





\end{document}